\documentclass[preprint,12pt]{elsarticle}
\usepackage{times}
\usepackage{epsfig}
\usepackage{graphicx}
\usepackage{amsmath}
\usepackage{amssymb}
\usepackage{multirow} 
\usepackage{booktabs} 
\usepackage{adjustbox}
\usepackage[draft,inline,nomargin,marginclue]{fixme}
\usepackage{url}
\usepackage{caption}
\usepackage{subcaption}
\usepackage{lipsum}
\usepackage{stmaryrd}
\usepackage{placeins}
\usepackage{flushend}

\usepackage[colorlinks=false, hidelinks, breaklinks=true,bookmarks=false]{hyperref}

\usepackage{cleveref}

\usepackage{fontawesome}
\usepackage{framed}
\usepackage[dvipsnames]{xcolor}
\definecolor{formalshadelight}{RGB}{242,242,242}
\definecolor{formalshadedark}{RGB}{50,50,50}
\newenvironment{highlight}{%
  \MakeFramed{\advance\hsize-\width\FrameRestore}%
  \noindent\begin{minipage}{\linewidth}\noindent\hspace{-4.55pt}%
}
{%
  \vspace{2pt}\vspace{-2pt}\end{minipage}\endMakeFramed%
}

\begin{document}

\begin{frontmatter}

\title{The Hidden Cost of an Image: Quantifying the Energy Consumption of AI Image Generation}

\author[1]{Giulia Bertazzini}
\ead{giulia.bertazzini@unifi.it}
\author[1]{Chiara Albisani}
\ead{chiara.albisani@unifi.it}
\author[1]{Daniele Baracchi}
\ead{daniele.baracchi@unifi.it}
\author[1]{Dasara Shullani}
\ead{dasara.shullani@unifi.it}
\author[1]{Roberto Verdecchia}
\ead{roberto.verdecchia@unifi.it}
\affiliation[1]{organization={Department of Information Engineering},
            addressline={University of Florence}, 
            city={Florence},
            postcode={50139}, 
            country={Italy}}

\begin{abstract}
With the growing adoption of AI image generation, in conjunction with the ever-increasing environmental resources demanded by AI, we are urged to answer a fundamental question: What is the environmental impact hidden behind each image we generate? In this research, we present a comprehensive empirical experiment designed to assess the energy consumption of AI image generation. Our experiment compares 17 state-of-the-art image generation models by considering multiple factors that could affect their energy consumption, such as model quantization, image resolution, and prompt length. Additionally, we consider established image quality metrics to study potential trade-offs between energy consumption and generated image quality. 
Results show that image generation models vary drastically in terms of the energy they consume, with up to a 46x difference. Image resolution affects energy consumption inconsistently, ranging from a 1.3x to 4.7x increase when doubling resolution. U-Net-based models tend to consume less than Transformer-based one. Model quantization instead results to deteriorate the energy efficiency of most models, while prompt length and content have no statistically significant impact. Improving image quality does not always come at the cost of a higher energy consumption, with some of the models producing the highest quality images also being among the most energy efficient ones.

\end{abstract}


\end{frontmatter}

\section{Introduction}
Text-to-image models have seen rapid adoption in the most recent years and are, with high probability, here to stay. Albeit the numerous benefits image generation models offer, a looming question remains to date unanswered: \textit{What is the environmental impact of image generation models}? The ever-increasing environmental impact of Artificial Intelligence (AI) is becoming an established concern in academic literature that can no longer be neglected~\cite{verdecchia2023systematic}. Although past studies covered a wide range of topics related to the environmental sustainability of AI, to the best of our knowledge, no research effort so far considered the energy required by image generation models. 

In this research, we aim to gain a comprehensive understanding of the energy consumption of image generation models by presenting a comprehensive empirical experiment comprising 17 distinct state-of-the-art diffusion models. The experiment is conducted considering various aspects that may influence model energy efficiency, such as model quantization, image resolution, and prompt length. To gain systematic and sound insights, our research method comprises multiple experimental re-runs and statistical analyses, leading to the execution of more than 9k distinct experimental measurements. In addition to model energy consumption, we complement our results by evaluating also the quality of the generated images, achieved \textit{via} the use of consolidated metrics specifically designed to measure the quality of generated images.

On one hand, our research targets researchers interested in understanding and improving the environmental sustainability of image generation, by providing a comprehensive analysis on the factors influencing energy consumption of this process. On the other hand, our study is intended to reach a wider audience, by providing concrete evidence grounded in empirical data of the to date overlooked environmental cost hidden behind image generation.

To support scrutiny of our results, aligned with open science principles, we make a comprehensive supplementary package of our study 
available in our companion material.

\subsection{Related Work}
Recent research has highlighted the substantial carbon footprint and high energy consumption associated with training large deep learning models. However, relatively few studies focus on their efficiency during inference, despite its critical role in real-world deployment.

Gowda et al.~\cite{gowda2023watt} analyze the energy consumption of various deep learning models, providing a detailed trade-off between accuracy and efficiency. Their study quantifies the energy consumption for both training and testing in terms of backbone architecture, dataset, GFLOPs, and parameter count, by disregarding however image generation tasks.

A large-scale assessment on the environmental impact of machine learning is conducted by Castano et al.~\cite{castano2023exploring}. They evaluated the carbon footprint of 1,417 models hosted on Hugging Face, covering diverse domains such as multimodal models, audio, and computer vision.
Notably, they find no statistically significant differences in carbon footprint between pre-trained and fine-tuned models.

As other pioneering study in this space, Budennyy et al.~\cite{budennyy2022eco2ai} assesses the $CO_2$ emissions of text-to-image generative models. They focus only on two models, Malevich  and Kandinsky~\cite{razzhigaev2023kandinsky}, evaluating their consumption based on training epochs, loss, GPU/CPU usage, and batch size. 

Among the models considered in this work, only PixArt-$\alpha$ \cite{chen2023pixart} reported $CO_2$ emissions. We therefore assert that, while commonly computational cost of training is considered -- typically in terms of GPU hours or Neural Function Evaluations (NFEs) -- the environmental impact of models is most commonly overlooked.

Synthetic image generation models are usually evaluated using both quantitative metrics, such as FID (Fréchet Inception Distance)~\cite{fid}, Precision-Recall~\cite{precision_recall_distributions}, IS (Inception Score)~\cite{salimans2016improved}, and CLIPScore~\cite{hessel2021clipscore}, as well as qualitative user studies. FID and Precision-Recall require a reference dataset, representing the real data distribution against which distances are measured. In contrast, IS operates without the need of a reference dataset. All these discussed metrics rely on features extracted by a pre-trained InceptionV3~\cite{Szegedy_2016_CVPR} network. Unlike them, CLIPScore does not 
assess image quality directly but instead measures how well the generated images align with the given prompts. 

Some research has explored the impact of image resolution on generation quality. PixArt-$\sigma$~\cite{chen2024pixart} finds that increasing resolution improves both FID and CLIP scores, testing images at 512 and 1024 pixels with aspect ratios between 1 and 9. Similarly, Lumina~\cite{gao2024lumina} provides qualitative comparisons of images upscaled from 512 to 2048 pixels, benchmarking against PixArt-$\alpha$~\cite{chen2023pixart}. Stable Diffusion 1.5~\cite{rombach2022high} examines 4$\times$ upscaling with FID and IS metrics.

Beyond resolution, studies investigate how architectural choices influence generative performance. Factors such as number of diffusion steps, guidance scale, and network type (e.g., UNet or LoRA) are frequently considered. Stable Diffusion XL Turbo~\cite{sauer2025adversarial} explores variations in loss functions, discriminator types, conditioning methods (image- or text-based), and initialization strategies. Stable Diffusion 3~\cite{esser2024scaling} studies the impact of model depth and guidance weight on FID, while Flash Stable Diffusion~\cite{chadebec2024flash} and Flash Stable Diffusion XL~\cite{chadebec2024flash} analyze the effects of timestamp sampling.

To the best of our knowledge, no prior work has systematically examined the energy consumption of image generation by analyzing the interplay between model architecture, quantization, resolution, and prompt length.
In this study, we provide new insights into the the environmental impact of diffusion models during inference, while also investigating the relationship between a model's environmental footprint and the quality of the generated images.

\section{Preliminaries}
In this section, we provide an overview of the key concepts related to image generation with diffusion models, that are at the basis of this study. 
Specifically, we focus on two main classes of diffusion models: \textbf{U-Net-based models}, that leverage U-Net backbone, and \textbf{Transformer-based models}, that replace the U-Net with a Vision Transformer.

\subsection{U-Net-based Diffusion Models}
Diffusion models have emerged as a leading paradigm in image generation.
Originally introduced by \cite{pmlr-v37-sohl-dickstein15}, Denoising Diffusion Probabilistic Models (DDPMs)~\cite{ho2020denoising} involve a forward diffusion process and a reverse denoising process. 
These early models are defined as \textit{unconditional}, as they generate images without external input.

In the \textit{forward diffusion process}, Gaussian noise $\epsilon \sim \mathcal{N}(0,I)$ is progressively added to an input image $x_0 \sim q(x_0)$, sampled from the data distribution, over $T$ timesteps, following a noise schedule. 
By timestep $T$, the image is transformed into pure noise.
This process is modeled as a Markov chain, defined by $q(x_1,...,x_T):=\prod_{t=1}^{T}q(x_t|x_{t-1})$ where $ q(x_t|x_{t-1}) = \mathcal{N}(x_t; \sqrt{1-\beta_t}x_{t-1}, \beta_tI)$.
In this formulation, $t$ and $\beta_t$ denote the timestep and noise schedule, respectively. 

On the other side, the \textit{reverse denoising process} consists of removing the noise and reconstructing the original image $x_0$, starting from the completely noisy version, $x_T \sim \mathcal{N}(0, I)$.
This is done by iteratively denoising each timestep, using a U-Net based neural network predicting the mean and the variance of the denoised image at each timestep: $p_\theta(x_{t-1} | x_t) = \mathcal{N}(x_{t-1}; \mu_\theta(x_t, t), \Sigma_\theta(x_t, t))$, where $\mu_\theta(x_t, t)$ and $\Sigma_\theta(x_t, t)$ represent the mean and variance of the Gaussian distribution.

To improve the efficiency of diffusion model training on limited computational resources without compromising quality and flexibility, \cite{rombach2022high} introduced latent diffusion models (LDMs).
LDMs use a Variational Autoencoder (VAE) to encode the high-resolution input image into a low-dimensional latent representation, allowing the U-Net to perform the diffusion process with reduced computational complexity.
Additionally, a CLIP-based text encoder \cite{radford2021learning} converts input text prompts into embeddings that condition the U-Net and effectively guide the diffusion process. 
These kind of models are categorized as \textit{conditional}.

\subsection{Transformer-based Diffusion Models}
Transformer-based diffusion models \cite{peebles2023scalable} replace the commonly-used U-Net backbone
with a transformer \cite{vaswani2017attention} that operates on latent patches.
Starting from Gaussian noise, a transformer network reverses the diffusion process to generate the target image, leveraging diffusion timesteps to generate distinct features at each stage.

These models support class conditioning, enabling to generate images that correspond to specific class labels. This feature allows the network to produce images from predefined categories by conditioning the transformer on both the class label and the timestep embedding, making it highly flexible for controlled image generation.

The architecture of these models follows the Vision Transformers (ViT) \cite{alexey2020image}, combining multi-head self-attention (MHSA) and a multi-layer perceptron (MLP) blocks, described as $X \leftarrow X + \alpha \text{MHSA}(\gamma X + \beta)$, $X \leftarrow X + \alpha'\text{MLP}(\gamma' X + \beta')$, where $X \in \mathbb{R}^{N\times C}$ represents image tokens, with $N$ as the number of tokens and $C$ as the channel dimension.
The parameters ${\alpha, \gamma, \beta, \alpha', \gamma', \beta'}$ come from adaptive layer normalization (adaLN) integrating class condition embedding $E_{\text{cls}}$ and timestep embedding $E_t$.

\section{Methodology}
This section documents the methodology used to study the energy consumption of different diffusion models during the image generation process.
To ensure in-depth analysis, a wide set of diffusion models was selected. 
The study investigated how changes of key factors, such as models, quantization, image resolution, and prompt length, impact energy consumption. 
In addition to the energy consumption analysis, we also conducted a quality assessment of the generated images to explore the relationship between model energy consumption and output quality.
The details of our experimental setup are presented in the following sections.

\subsection{Experimental Variables}\label{sec:experimental_variables}
To evaluate the energy consumption of different diffusion models during the image generation process, we selected a diverse set of diffusion models, including both U-Net-based and Transformer-based ones.
These categories were based on their different backbone architecture, resulting in distinct image generation processes and therefore potentially different energy consumption patterns. 

In designing our experiments, we identified key variables to assess their influence on energy consumption while ensuring the fairest comparison possible between models.
To analyze model behavior across multiple realistic usage scenarios, we selected \textbf{model quantization}, \textbf{image resolution}, and \textbf{prompt length} as the primary experimental variables.
Conversely, we excluded the number of diffusion steps, as each diffusion model is typically designed to operate with a specific step count, which can range from as few as 2 steps to 50 or more. Running a model designed for 2 steps with 50 steps, or \textit{vice versa}, would not only degrade image quality but also introduce significant biases in the comparison, as models are not intended to function under such conditions.

\subsubsection{Diffusion Models Selection}
For this study, we selected a total of \textbf{17 diffusion models}, based on their underlying architectures: U-Net-based and Transformer-based models. 
By including models from both groups, we aimed to capture a comprehensive range of diffusion-based generation techniques.

In selecting the models, we prioritized state-of-the-art models over older ones, with a focus on the models that are more widely used. 
We also included a few slightly outdated models that remain popular and continue to be highly downloaded by end users.
In this way, we ensured a representation of both cutting-edge technologies and well-established ones for our experimentation. 
Furthermore, we selected models that are free for academic use and accessible offline, i.e., excluding subscription-based solutions such as DALL$\cdot$E \cite{ramesh2022hierarchical} and Midjourney \cite{midjourney}, which are not publicly accessible and hence we could not experiment on.
All chosen models are available on Hugging Face, allowing for their standardized use and minimizing implementation variability. 
The models selected for our experimentation are: 

\textbf{U-Net-based}: Stable Diffusion 1.5 (SD\_1.5)~\cite{rombach2022high}, Stable Diffusion XL (SDXL)~\cite{podell2023sdxl}, Stable Diffusion XL Turbo (SDXL\_Turbo)~\cite{sauer2025adversarial}, Stable Diffusion XL Lightning (SDXL\_Lightning)~\cite{lin2024sdxl}, Hyper Stable Diffusion (Hyper\_SD)~\cite{ren2024hyper}, Segmind Stable Diffusion 1B (SSD\_1B)~\cite{gupta2024progressive}, Latent Consistency Model Segmind Stable Diffusion 1B (LCM\_SSD\_1B)~\cite{luo2023latent}, Latent Consistency Model Stable Diffusion XL (LCM\_SDXL)~\cite{luo2023latent}, Flash Stable Diffusion (Flash\_SD)~\cite{chadebec2024flash}, Flash Stable Diffusion XL~(Flash\_SDXL)~\cite{chadebec2024flash}.

\textbf{Transformer-based}: PixArt-$\alpha$ (PixArt\_Alpha)~\cite{chen2023pixart}, PixArt-$\sigma$ (PixArt\_Sigma)~\cite{chen2024pixart}, Flash PixArt (Flash\_PixArt)~\cite{chadebec2024flash}, Stable Diffusion 3 (SD\_3)~\cite{esser2024scaling}, Flash Stable Diffusion 3 (Flash\_SD3)~\cite{chadebec2024flash}, Lumina-Next-SFT (Lumina)~\cite{gao2024lumina}, Flux.1 schnell (Flux\_1)~\cite{flux2023}.

Many of the models considered in this study are derivatives of a base model that has been distilled or otherwise modified to enhance image quality and improve efficiency. For instance, Stable Diffusion XL serves as the foundation for multiple variants, including Turbo, Lightning, LCM, and Flash. While these versions all originate from the same base model, they integrate distinct optimizations and refinements that influence the image generation process, potentially leading to energy consumption variations.

\subsubsection{Model Quantization}

To address the substantial memory demands of diffusion models, quantization emerged as an effective technique for reducing memory usage while preserving high model performance. Given the widespread use of quantization, we study the energy consumption of each investigated model \textbf{in both quantized and unquantized forms}.
For this purpose, we utilized Quanto\footnote{\url{https://github.com/huggingface/optimum-quanto}}, a quantization toolkit built on PyTorch provided through Hugging Face Optimum, a suite of tools for hardware optimization. Among the available quantization options, we selected \texttt{int8} quantization, as it is a widely used approach that significantly reduces memory usage while maintaining acceptable image quality.

\subsubsection{Image Resolution}
Image resolution is a critical factor that influences both computational complexity and visual quality of generated images. 
Higher resolutions generally produce higher visual quality, while being characterized by increased computational requirements, and hence higher memory consumption, processing times, and potentially energy consumption. 

In our study, we considered two widely used square resolutions, \textbf{512$\times$512} and \textbf{1024$\times$1024}, along with a non-square resolution, namely \textbf{768$\times$1024}.
This approach allowed us to explore the potential impact of aspect ratio variations on generation performance and visual quality of the images.

\subsubsection{Prompt Length}\label{sec:prompt_length}
Understanding whether longer prompts introduce unexpected computational overhead or efficiency variations in the image generation process is important for optimizing prompt engineering strategies for different applications. 
Subtle variations in processing time, memory usage, or energy consumption could arise when handling more complex or descriptive input prompts. 

To evaluate the effect of prompt length on energy consumption, we considered three different lengths: short, medium, and long. 
For \textbf{short-length prompts}, we randomly selected 30 classes from CIFAR-100~\cite{krizhevsky2009learning} dataset and we used them without any adaptation as short prompts.  Therefore, short prompts consist of one or two words as presented in the original dataset.
The CIFAR-100 prompts were then expanded into \textbf{long-length prompts} by using the Phi-3-Mini-4K-Instruct~\cite{abdin2024phi} large language model (LLM), a lightweight and state-of-the-art open model, specifically designed for general-purpose AI systems.
We used 13 examples from Stable Diffusion's best-performing photographic prompts~\cite{Wind2023} as a reference, instructing the LLM to add additional details to the input short prompt. %
Therefore, long prompts consists of an average of 33 words.
Finally, long prompts were manually truncated to obtain \textbf{medium-length prompts}, consisting of an average of 14 words.

\subsection{Experiment Execution}

We carried out a preliminary test to evaluate the influence of prompt semantic content on energy consumption. We selected as prompts the 100 distinct classes of the CIFAR-100 dataset, leveraging its broad class variety. Each model processed the prompts 10 times in randomized order to mitigate ordering biases.
Then, to ensure consistency, all models were tested using their default settings, avoiding potential parameter-related influences.

The results of the preliminary test indicated no significant correlation between the semantic content of the prompts and energy consumption, confirming that semantic variability is not a critical factor for this study.
Further details on the preliminary test are provided in are provided in the supplementary material of this work. 

Building upon the findings of the preliminary test, we conducted the main experiments of this study, which focused on measuring energy consumption for each model while varying the independent variables outlined in \Cref{sec:experimental_variables}.
To this end, we randomly selected 30 prompts from the CIFAR-100 dataset classes to form the short-length prompts.
Then we expanded and truncated these prompts to obtain the long-length and medium-length prompt, following the approach detailed in \Cref{sec:prompt_length}.
Similar to the preliminary experiment, to prevent biases associated to prompt ordering and boundary effects, the prompts were randomly shuffled before each experiment.

Our setup involved generating images using 17 different diffusion model under their default settings while varying two quantization parameters, three image resolution levels, and three prompt lengths.
As a results, each prompt underwent $17 \times 2 \times3 \times 3 = 540$ experimental runs, leading to a total of $540 \times 30 = 9180$ experiments across all prompts. 

Energy consumption was measured \textit{via} CodeCarbon~\cite{benoit_courty_2024_11171501}, a Python library allowing to measure the energy consumed during the generation process of a image in kilowatt-hours (kWh) through the formula $E=E_{CPU}+E_{GPU}+E_{RAM}$, i.e., the sum of CPU, GPU, and RAM energy consumption.

All experiments were performed on a dedicated workstation with an AMD Ryzen 9 7950X processor, 128 GB of RAM, and an NVIDIA GeForce RTX 4090.

\subsection{Image Quality Assessment}\label{sec:qass}

Since no definitive reference set exists for text-to-image diffusion models, a common approach~\cite{chen2023pixart, chadebec2024flash} is to use the validation set of the dataset from which class labels were used as prompts. 
In our setting, instead of using the CIFAR-100 validation images, which have low resolution (32$\times$32), we employed the ImageNet-1k validation set. However, since the class labels in ImageNet-1k do not fully overlap with our selected prompt classes, we applied a sampling strategy to guarantee that the same subjects are uniformly represented. Specifically, we utilized our prompt classes as queries to identify corresponding classes in ImageNet-1k. First, we selected all ImageNet-1k classes which had at least one match with a prompt. If a unique match was found, we retrieved all 50 images for that class. Otherwise, where multiple matches existed (e.g., prompt ``clock" matched multiple classes such as ``wall clock", ``analog clock'' and ``digital clock''), we performed uniform random sampling across the matched classes to maintain balance and diversity in the final dataset.

\textit{Via} our filtering strategy we obtained a final dataset for quality assessment consisting of 1,100 images from the ImageNet-1k validation set and 396 images for each diffusion model considered.
FID computation was performed using IQA-PyTorch library~\cite{pyiqa}, while Precision and Recall were computed using the implementation provided by \cite{coverage_density}.

\section{Results}\label{sec:results}
In this section, we present the results of our experiment by considering the various independent variables studied, i.e., models, quantization, image resolution, and prompt length. As concluding remark, we additionally explore the interplay between energy consumption and image quality.

\subsection{Median Model Energy Consumption}
\begin{figure}
    \centering
    \includegraphics[width=\linewidth]{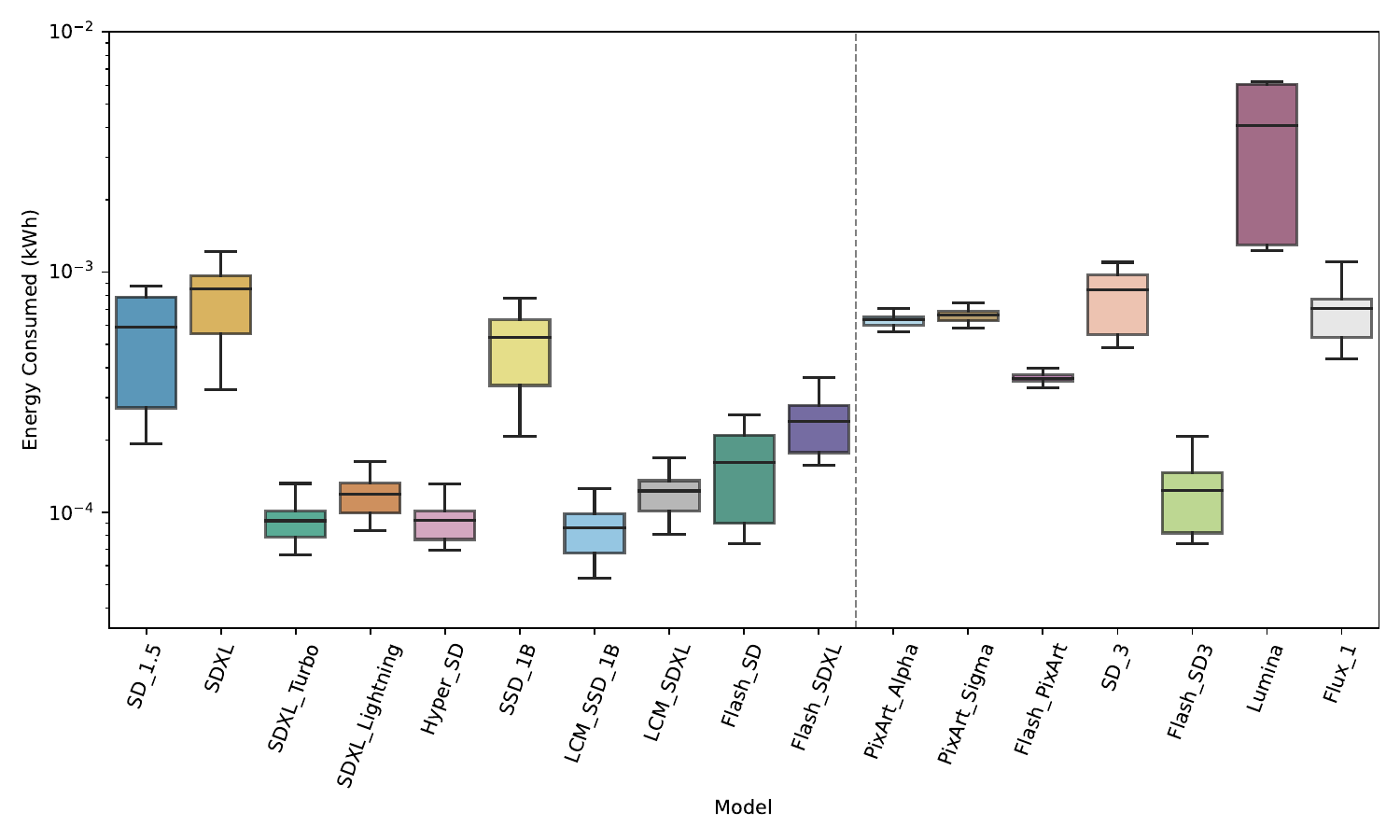}
    \caption{Boxplot of energy consumption across the investigated models. The dotted grey lines separate U-Net (left) from Transformer-based models (right).}
    \label{fig:all_models}
\end{figure}

An overview of the overall energy consumption of the models (i.e., the median energy consumption of each model across all independent variables) is depicted in Figure~\ref{fig:all_models}. As we can observe from Figure~\ref{fig:all_models}, the considered models display a notable difference in terms of median energy consumption, with Lumina being the models consuming the highest energy value ($4.08\times10^{-3}$ kWh) and LCM\_SSD\_1B the lowest one ($8.6\times10^{-5}$ kWh). While the difference of median energy consumption is less drastic for other models (e.g., PixArt\_Alpha and PixArt\_Sigma differ only for a negligible $4\times10^{-5}$ factor), we conclude that choosing a model over another one can lead to drastic energy savings up to 46x. Overall, we can observe that generally U-Net-based models tend to consume less than Transformer-based ones, reflecting the greater computational complexity inherent to transformer architectures.  As additional observation, we note that the variability in terms of energy consumption differs between models, with some displaying only a limited energy consumption fluctuation across settings (e.g., the Pixart-based models), while others a much higher one (e.g., Lumina, SD\_1.5, and Flash\_SD).

\begin{highlight}
\noindent\faLeaf~\textbf{Median Model Energy Consumption:} Energy consumption varies drastically across models, showcasing up to a 46x increment. U-Net-based models tend to consume less than Transformer-based ones, while consumption fluctuations vary from model to model.
\end{highlight}

\subsection{Model Quantization}

\begin{figure*}[t]
    \centering
    \includegraphics[width=\textwidth]{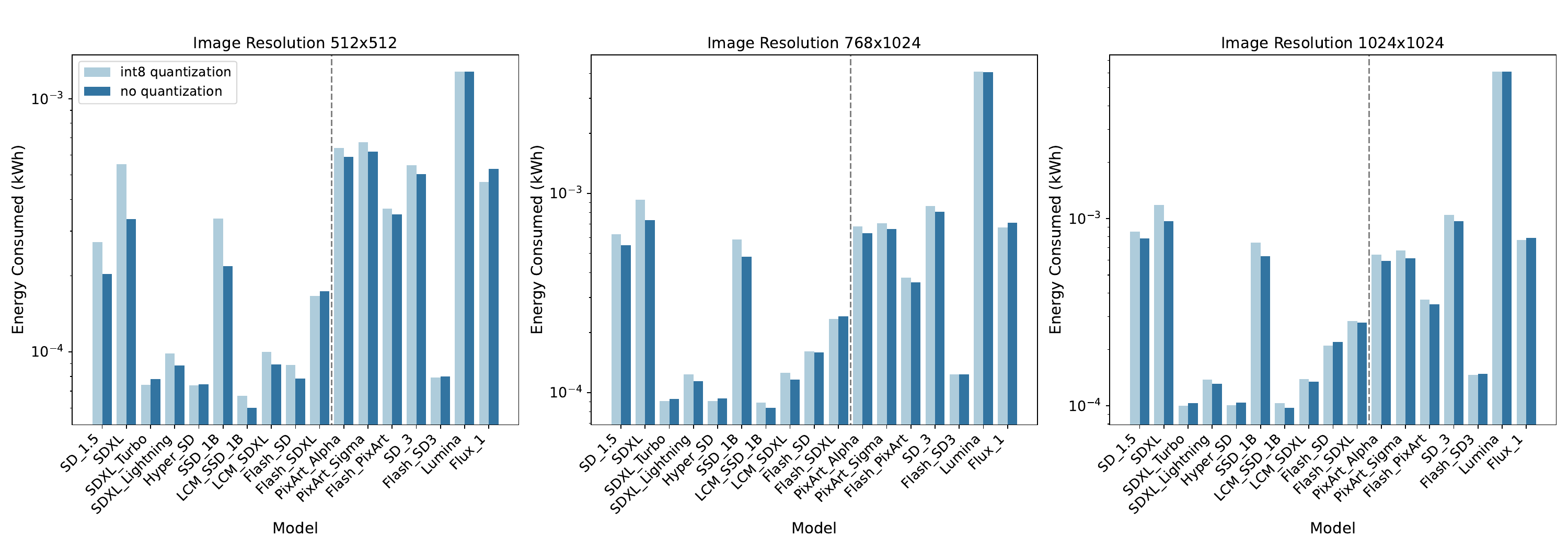}
    \caption{Energy consumption (kWh) for image generation process of diffusion models at varying quantization levels. Bar charts (log scale) compare int8-quantized vs. non-quantized models at fixed resolutions. The dotted grey lines separate U-Net (left) from Transformer-based models (right). Each bar shows the median over 30 prompts.}
    \label{fig:quantization}
\end{figure*}

\Cref{fig:quantization} illustrates, on a logarithmic scale, the energy consumption (in kWh) of the analyzed models across three image resolutions, comparing their performance when quantization using the \textit{int8} data type and \textit{no quantization} is applied. 
The energy consumed values for each model are the medians across the 30 prompts.
The results indicate that most models exhibit increased energy consumption when quantized to the int8 data type, with the exception of four models—SDXL\_Turbo, Hyper\_SD, Flash\_SD3, and Flux\_1—which show a modest reduction in energy usage (up to 12.13\%). 
This observation suggests that quantization primarily optimizes memory usage without optimizing the energy efficiency of the image generation process.
Furthermore, as we can see from \Cref{tab:quant}, the percentage difference in energy consumption between the quantized and non-quantized versions is generally negligible, except for few models, e.g., SD\_1.5, SDXL, and SSD\_1B. 

\begin{highlight}
    \faLeaf~\textbf{Model Quantization:} Counterintuitively, model quantization leads in the vast majority of cases to an energy consumption increase (up to 64.54\%) and only in rare cases to a negligible energy saving (up to 12\%). 
\end{highlight}

\begin{table}[t]

\caption{Percentage variation of energy consumption of diffusion models at varying quantization levels. Variation is computed as the difference between non-quantized and quantized model for each resolution. Negative values indicate lower energy consumption in non-quantized models.
 }\label{tab:quant}
\centering
\adjustbox{width=0.6\textwidth}{
\begin{tabular}{l|l|ccc}
\toprule
 \multirow{2}{*}{\textbf{Type}} & \multirow{2}{*}{\textbf{Model}} & \multicolumn{3}{c}{\textbf{Image Resolution}} \\
 &  & \multicolumn{1}{l}{\textbf{512$\times$512}} & \multicolumn{1}{l}{\textbf{768$\times$1024}} & \multicolumn{1}{l}{\textbf{1024$\times$1024}} \\
 \midrule
\multirow{10}{*}{\rotatebox{90}{\textbf{U-Net-based}}} & SD\_1.5 & -33.33\% & -13.67\% & -8.46\% \\
 & SDXL & -64.58\% & -26.61\% & -22.34\% \\
 & SDXL\_Turbo & 5.32\% & 2.13\% & 3.51\% \\
 & SDXL\_Lightning & -11.80\% & -8.08\% & -6.09\% \\
 & Hyper\_SD & 1.49\% & 3.60\% & 2.59\% \\
 & SSD\_1B & -54.42\% & -22.22\% & -17.84\% \\
 & LCM\_SSD\_1B & -11.42\% & -6.43\% & -5.44\% \\
 & LCM\_SDXL & -12.74\% & -7.85\% & -3.47\% \\
 & Flash\_SD & -13.38\% & 0.96\% & 5.30\% \\
 & Flash\_SDXL & 3.77\% & 2.66\% & -1.73\% \\
 \midrule
\multirow{7}{*}{\rotatebox{90}{\textbf{Transformer-based}}} & PixArt\_Alpha & -8.30\% & -8.76\% & -7.95\% \\
 & PixArt\_Sigma & -8.88\% & -7.11\% & -9.75\% \\
 & Flash\_PixArt & -5.25\% & -6.40\% & -6.02\% \\
 & SD\_3 & -8.35\% & -7.13\% & -7.64\% \\
 & Flash\_SD3 & 1.05\% & 0.67\% & 0.75\% \\
 & Lumina & -0.15\% & -0.16\% & -0.20\% \\
 & Flux\_1 & 12.13\% & 5.97\% & 2.88\% \\
 \bottomrule
\end{tabular}}
\end{table}

\subsection{Image Resolution}
\begin{figure*}[t]
    \centering
    \includegraphics[width=\textwidth]{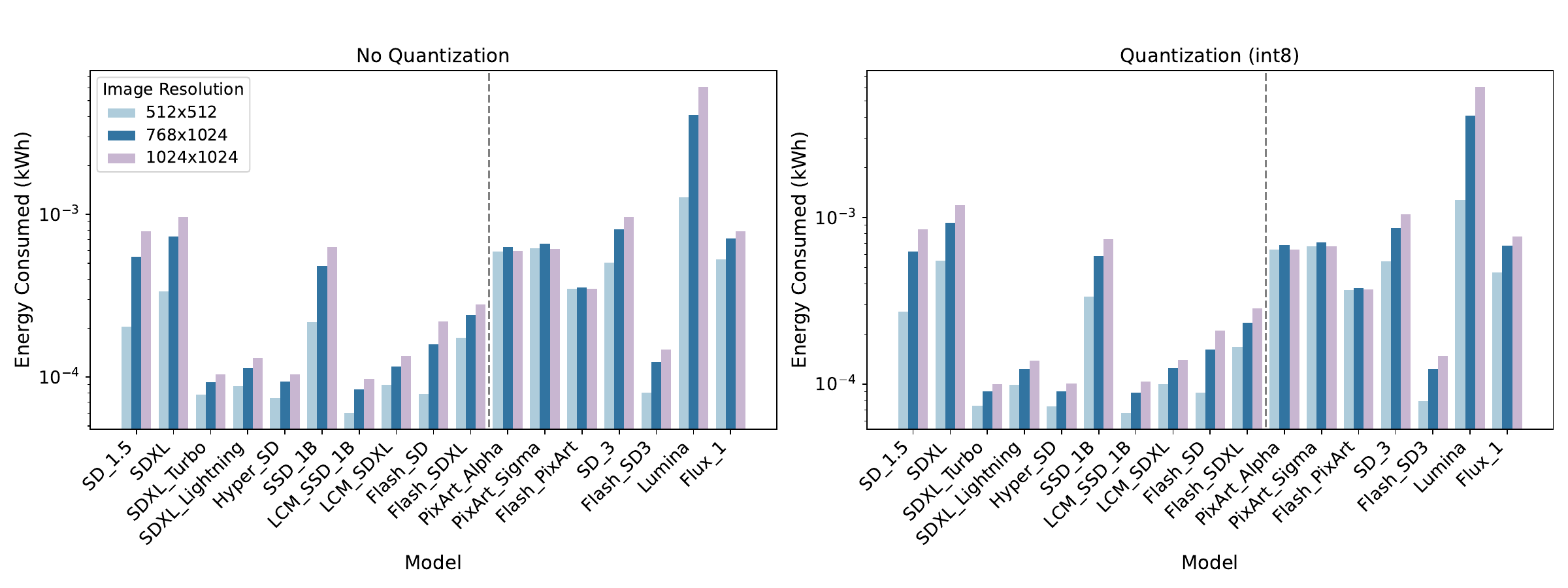}
    \caption{Energy consumption (kWh) for image generation process of diffusion models at varying image resolution. Bar charts (log scale) compare 512$\times$512, 768$\times$1024, and 1024$\times$1024 image resolutions at fixed quantization setting. The dotted grey lines separete U-net (left) from Transformer-based models (right). Each bar shows the median over 30 prompts.
    }
    \label{fig:resolutions}
\end{figure*}

\Cref{fig:resolutions} shows the impact of image resolution on the energy consumption. 
Without any surprise, most of the investigated models exhibit the expected trend of increased energy consumption as image resolution increases. 

\Cref{tab:img_res} highlights the varying energy consumption scaling of diffusion models as image resolution increases fourfold, from 512$\times$512 to 1024$\times$1024. 
Energy consumption variation is calculated as the ratio of the energy used to generate a 1024$\times$1024 resolution image to the energy used for 512$\times$512 resolution image. 
Regarding U-Net-based models, some of them exhibit a relatively low scaling ratio (around 1.3), such as SDXL\_Turbo and Hyper\_SD, indicating a more efficient handling of higher resolutions.
In contrast, models like SD\_1.5 (3.85) and SSD\_1B (2.90) show a much higher ratio, suggesting that their energy consumption is almost directly proportional to image resolution. 
For Transformer-based models, the PixArt family maintains a ratio close to 1.00, indicating that their energy consumption remains constant across resolutions. 
Other models, such as SD\_3 and Flash\_SD3, have slightly higher ratios ($\sim$1.9), implying a modest consumption increase. 
Lumina stands out with the highest ratio (4.75), exceeding even the expected quadratic scaling, which may suggest additional computational overhead or inefficiencies when processing high-resolution images.
As peculiar result, the PixArt-based models showcased a higher energy consumption for non-square formats. Further experiments conducted by considering other 11 formats between the range of 256$\times$256 to 1536$\times$1536 confirmed this trend. 
We conjecture this trend is due to the training data used for the model, which could have been formatted only in square resolution. Additional details on the follow-up experiments are available in the supplementary material of this work.

\begin{highlight}
\faLeaf~\textbf{Image resolution:} Image resolution affects differently the energy consumption of the models, with some showcasing low scaling ratio (e.g., 1.3x when image size is doubled) while other a considerable increase (e.g., 4.7x). Pixart family models remain close to constant across all square resolutions, albeit consuming considerably more for non-square ones. 
\end{highlight}

\begin{table}[t]
\caption{Scaling ratios of energy consumption for image generation at 1024$\times$1024 resolution compared to 512$\times$512 resolution across different diffusion models.}
\label{tab:img_res}
\centering
\adjustbox{width=0.6\textwidth}{
\begin{tabular}{l|l|cc}
\toprule
 \textbf{Type} & \textbf{Model} & \textbf{No Quantization} & \textbf{Quantization (int8)} \\ \midrule
\multirow{10}{*}{\rotatebox{90}{\textbf{U-Net-based}}} & SD\_1.5 & 3.85 & 3.13 \\
 & SDXL & 2.88 & 2.14 \\
 & SDXL\_Turbo & 1.33 & 1.35 \\
 & SDXL\_Lightning & 1.48 & 1.41 \\
 & Hyper\_SD & 1.38 & 1.37 \\
 & SSD\_1B & 2.90 & 2.21 \\
 & LCM\_SSD\_1B & 1.62 & 1.54 \\
 & LCM\_SDXL & 1.51 & 1.39 \\
 & Flash\_SD & 2.80 & 2.35 \\
 & Flash\_SDXL & 1.61 & 1.70 \\ \midrule
\multirow{7}{*}{\rotatebox{90}{\textbf{Transformer-based}}} & PixArt\_Alpha & 1.00 & 1.00 \\
 & PixArt\_Sigma & 0.99 & 1.00 \\
 & Flash\_PixArt & 1.00 & 1.01 \\
 & SD\_3 & 1.92 & 1.91 \\
 & Flash\_SD3 & 1.85 & 1.86 \\
 & Lumina & 4.75 & 4.75 \\
 & Flux\_1 & 1.50 & 1.63 \\ \bottomrule
\end{tabular}}
\end{table}

\subsection{Prompt Length}
From a Quantile-Quantile plot analysis the energy consumption of each model does not result to follow a normal distribution. Therefore, we study the correlation between energy consumption and prompt length \textit{via} the Kruskal-Wallis non-parametric test. 
Results showcase that for all models the p-values are notably above the conventional significance level of 0.05, implying that prompt length is not a determining factor in the energy consumption of the examined models.
Further details regarding this experiment are provided in the supplementary material of this work.%

\begin{highlight}
    \faLeaf~\textbf{Prompt Length:} Prompt length does not  impact in a statistically significant manner the energy consumed by image generation.
\end{highlight}

\subsection{Image Quality Assessment}

\begin{figure*}[!htb]
    \centering
    \includegraphics[width=\columnwidth]{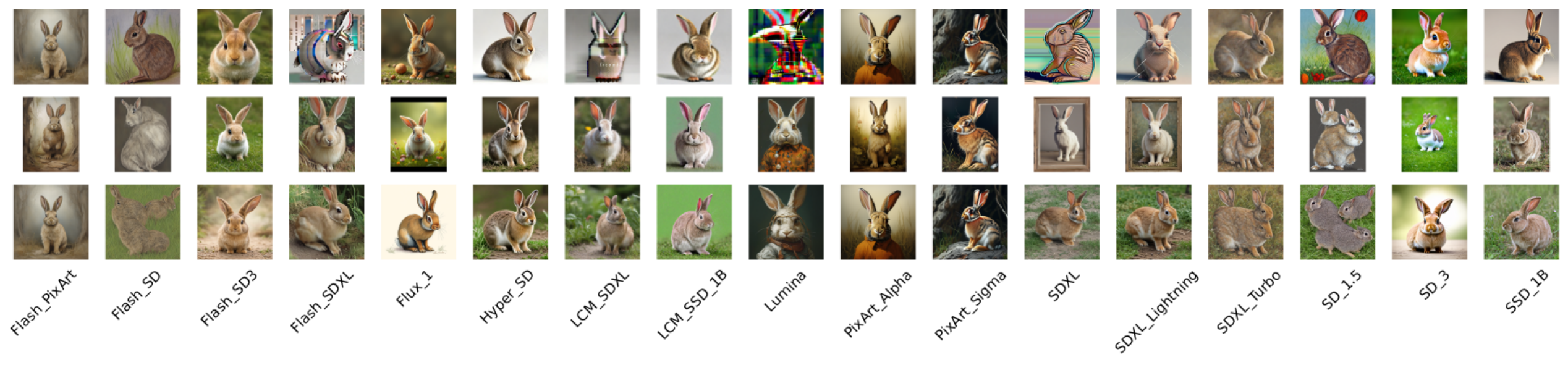}
    \caption{
    Example images generated by each model at different resolutions. Each row corresponds to a different resolution: 512×512 (top), 768×1024 (middle), and 1024×1024 (bottom). The images were generated using the short prompt ``rabbit" with int8 quantization.}\label{fig:all_models_short_prompt_5_int8}
\end{figure*}

\begin{figure}[hbpt]
    \centering
    \includegraphics[width=\linewidth]{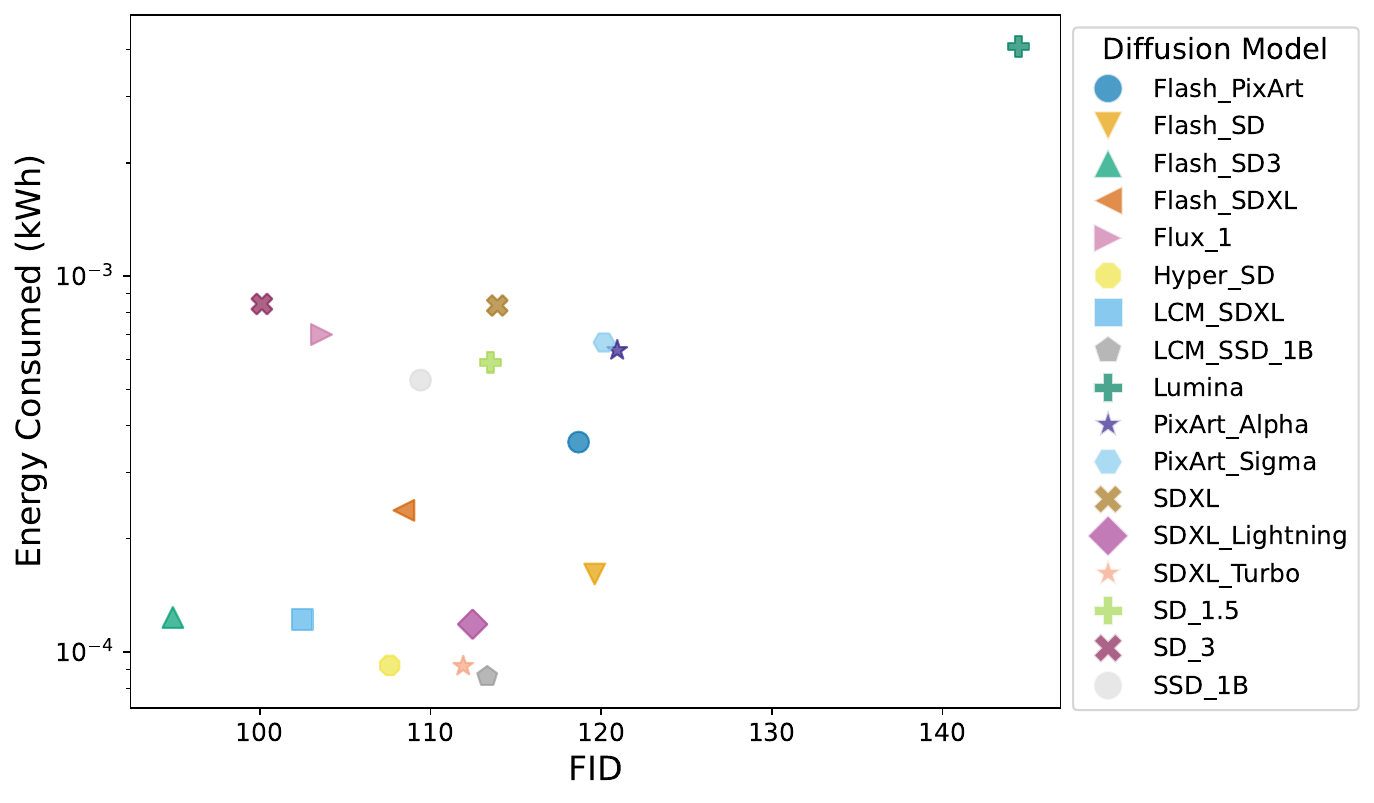}
    \caption{Energy consumption (kWh) vs. FID scores for diffusion models. Y-axis is on a logarithmic scale. Each dot represents the median energy consumption per model.} 
    \label{fig:fid_vs_energy}
\end{figure}

\Cref{fig:fid_vs_energy} documents the performance of each diffusion model in terms of FID and energy consumption. Low values of FID correspond to better quality.
Notably, Lumina emerges as the least efficient model, both from the quality and the energy perspective. On the other hand, Flash\_SD3 offers the best compromise, achieving the lowest FID among all models while maintaining low energy consumption.
As illustrated in \Cref{fig:all_models_short_prompt_5_int8}, where the same subject is represented at different resolutions, the final output of a diffusion model can vary significantly depending on the resolution. 
In particular, Lumina's high FID score is largely due to the poor quality of its 512$\times$512 images, as the model is fine-tuned on resolutions of 1024$\times$1024 and above, decreasing its ability to generate 512-resolution images effectively.
Similar effects are observed for other models, such as Flash\_SD, SD\_1.5 and SDXL\_Turbo, which exhibit intermediate FID scores despite producing semantically altered images at 768$\times$1024 and 1024$\times$1024 resolutions.

\begin{figure}[!ht]
    \centering
    \includegraphics[width=\linewidth]{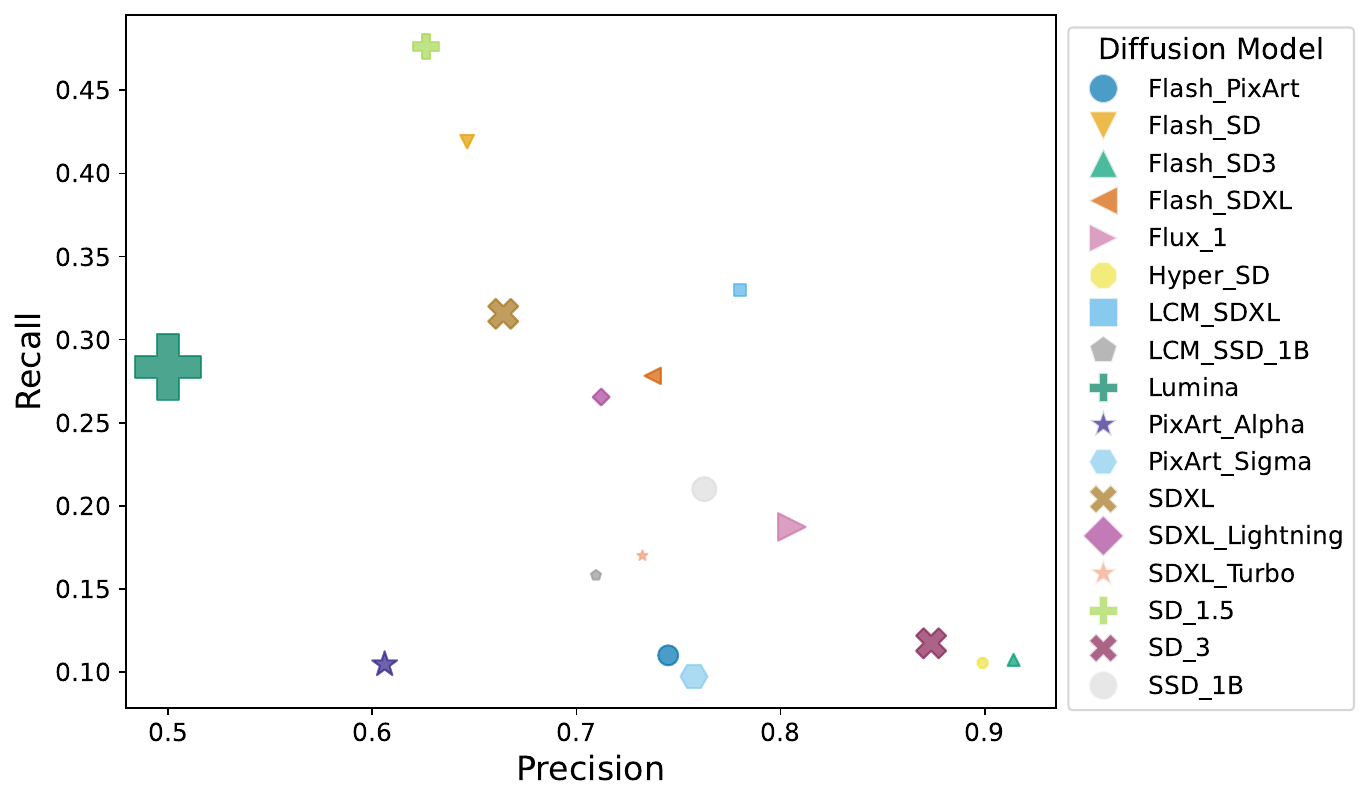}
    \caption{Precision vs. Recall for diffusion models. 
    Shape size is proportional to the median energy consumption per model.
    } 
    \label{fig:pr_vs_energy}
\end{figure}

FID captures both fidelity (realism) and diversity (variations in the original data) of generative models but struggles to separate their contributions when FID scores are similar. Precision and Recall overcome this limitation by independently measuring fidelity and diversity, providing a more nuanced evaluation of image quality.
In \Cref{fig:pr_vs_energy} each diffusion model is represented as a dot in the Precision-Recall space, with size indicating energy consumption. As we can observe, Lumina appears as the least precise model. 
Differently, while SD\_1.5 and Hyper\_SD achieve similar scores for FID in \Cref{fig:fid_vs_energy}, their distinct positions in the Precision-Recall space highlight qualitative differences. SD\_1.5 shows lower precision but higher recall with respect to Hyper\_SD. 
This behavior can be seen as a drawback, as high recall often results from generating numerous diverse but unrealistic samples rather than effectively capturing subject heterogeneity \cite{coverage_density}. Based on this observation, Hyper\_SD, SD\_3, and Flash\_SD3 stand out as the highest-quality models, combining strong performance with good energy efficiency. 

Additional discussions on quality metrics with resolution-specific plots are provided in the supplementary material of this work.%

\begin{highlight}
\faLeaf~\textbf{Image Quality:} Higher image quality does not come at the cost of higher energy consumption, with some of the most energy efficient models also producing among the highest quality images.

\end{highlight}

\section{Conclusions}
In this study, we analyze the energy consumption of 17 diffusion models, considering multiple influencing factors at interplay. The measured energy varies significantly across models (up to a 47x increase), with U-Net-based models generally being more efficient than Transformer-based ones.
Counterintuitively, model quantization often increases energy consumption (up to 64.54\%), with only a few cases showing minor savings. Image resolution affects energy consumption inconsistently, ranging from a 1.3x to 4.7x increase when doubling resolution. Considering the quality-energy connection, Lumina ranks lowest, while Flash SD3 excels in both aspects. 
In conclusion, with this research, we %
present a fine-grained empirical perspective on the energy required to generate images. With our study, we hope to have set a stepping stone to progress image generation not only by producing more eye pleasing images, but also by improving the environmental impact hidden behind each image we generate.

\bibliographystyle{elsarticle-num} 
\bibliography{biblio}

\clearpage
\appendix
\section{Preliminary Experiment}\label{app:preliminary_exp}
\begin{figure}[h]
     \centering
     \begin{subfigure}[b]{0.23\textwidth}
         \centering
         \includegraphics[width=\textwidth]{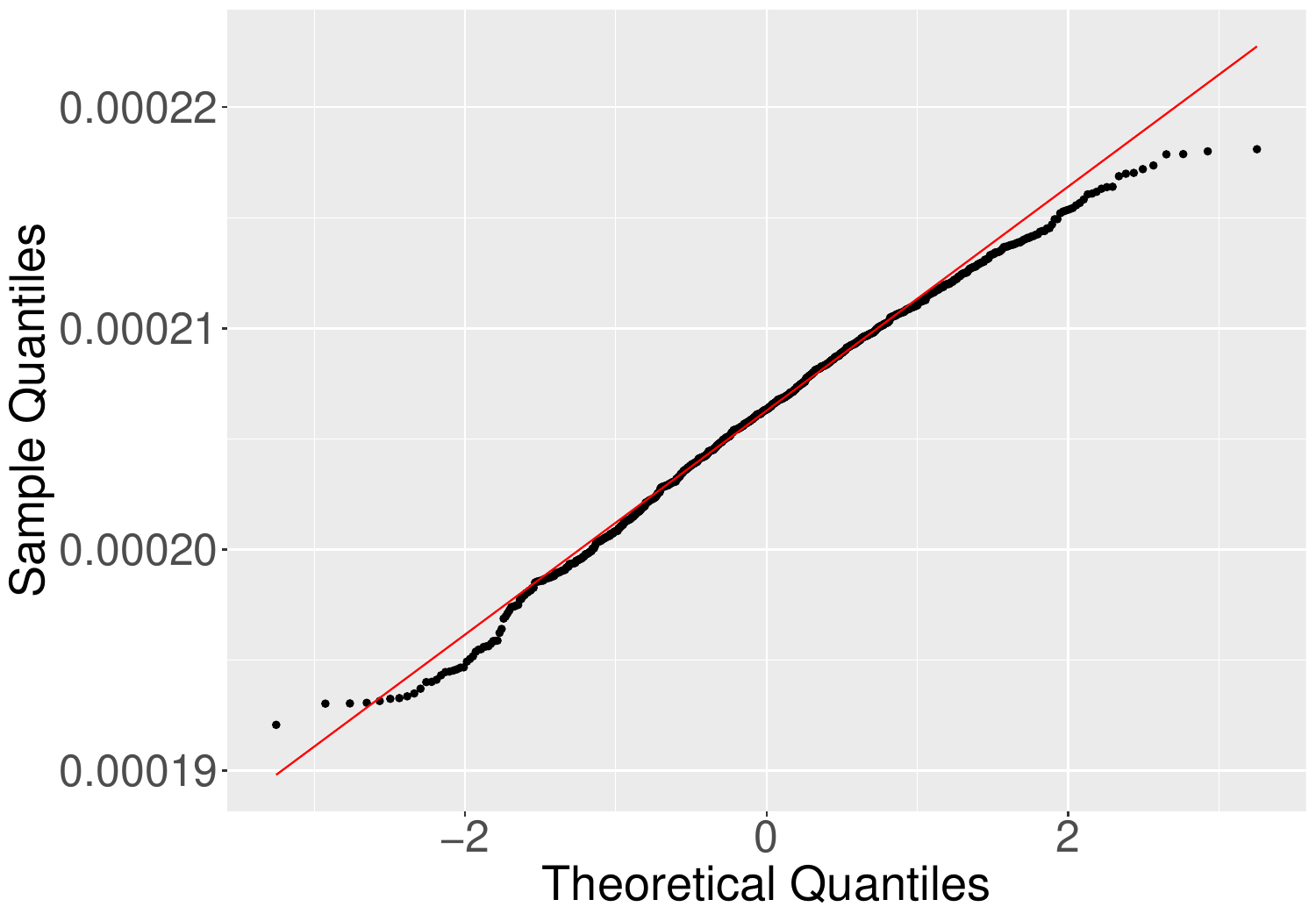}
         \caption{SD\_1.5}
     \end{subfigure}
     \hfill
     \begin{subfigure}[b]{0.23\textwidth}
         \centering
         \includegraphics[width=\textwidth]{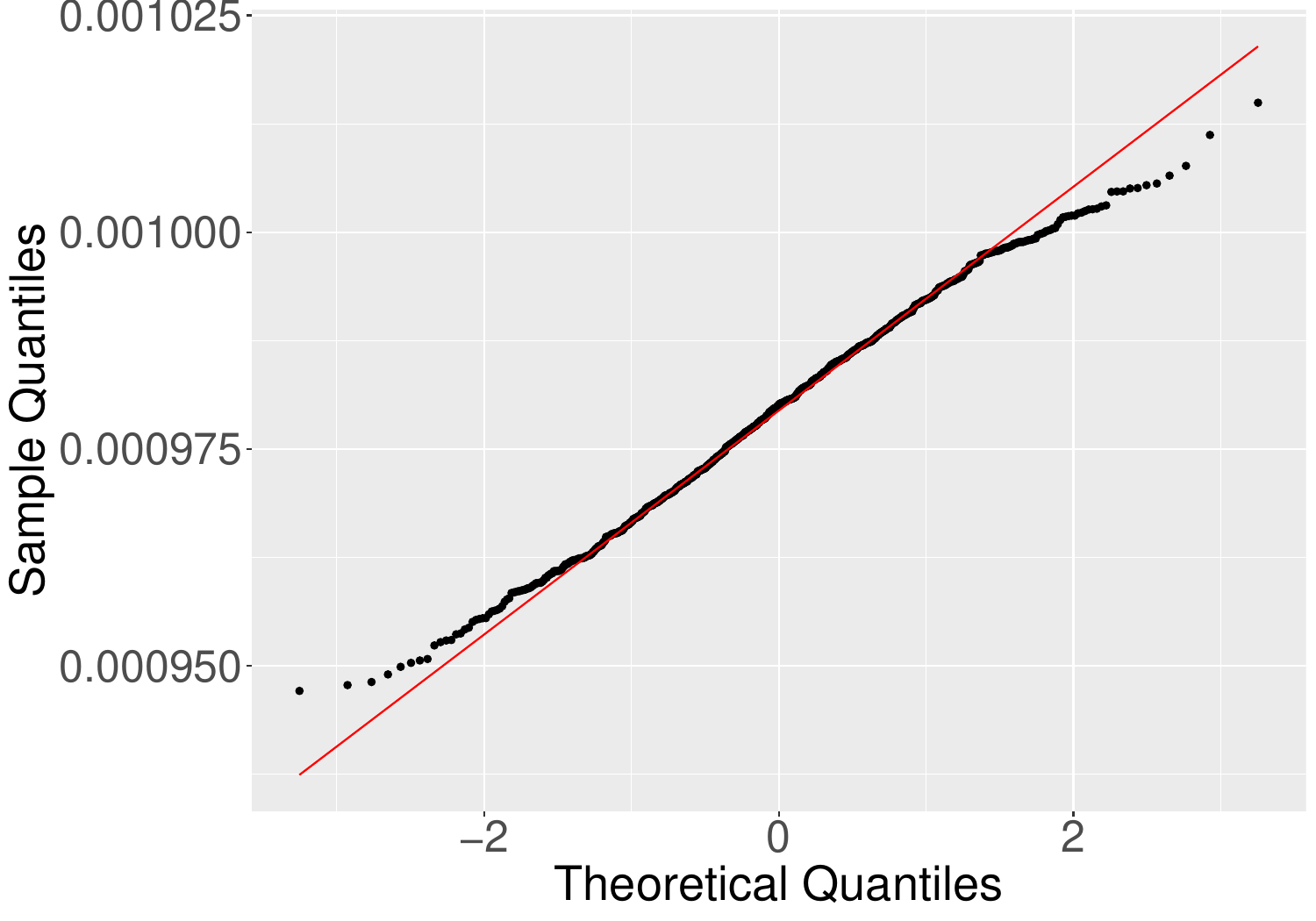}
         \caption{SDXL}
     \end{subfigure}
     \hfill
     \begin{subfigure}[b]{0.23\textwidth}
         \centering
         \includegraphics[width=\textwidth]{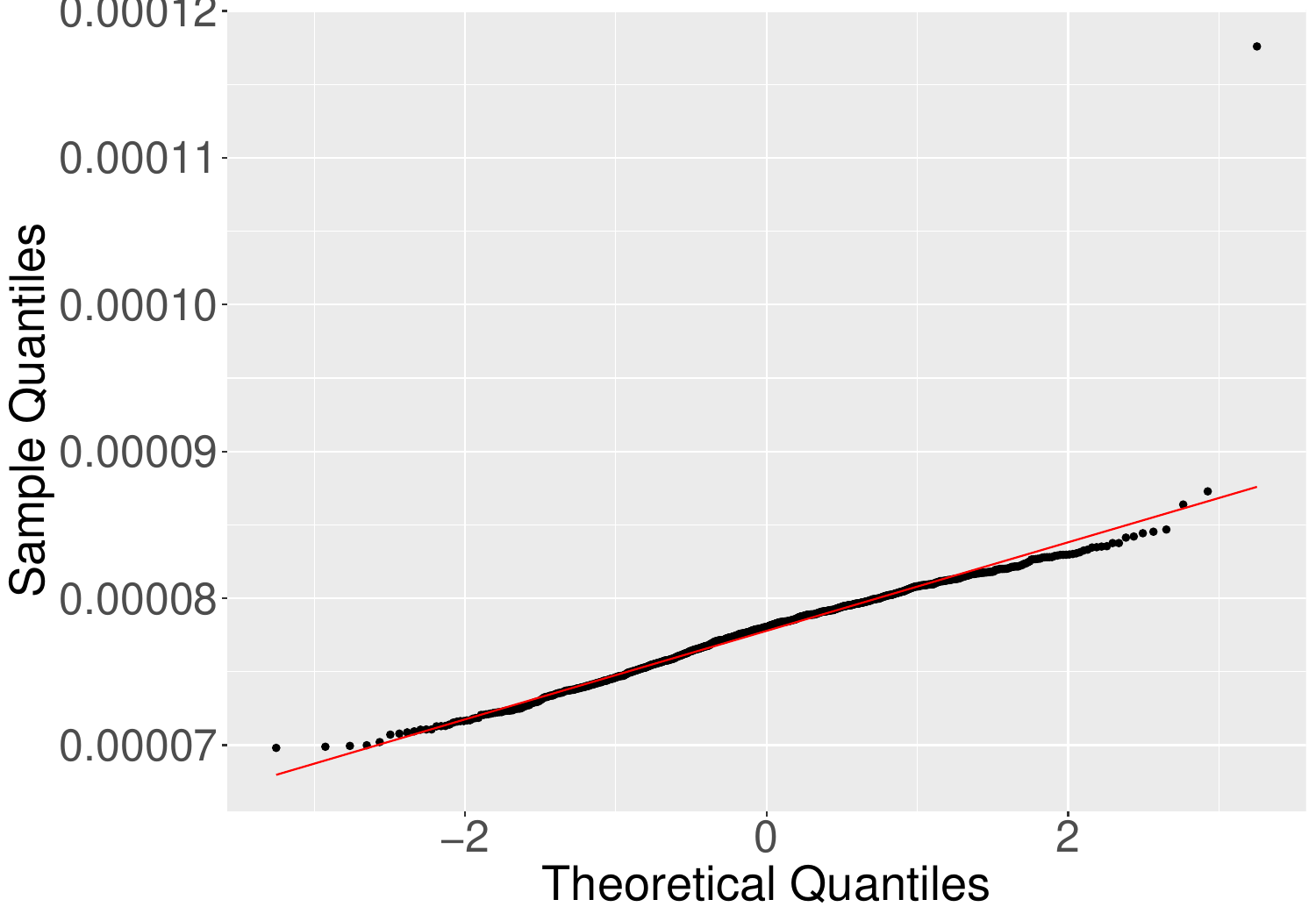}
         \caption{SDXL\_Turbo}
     \end{subfigure}
     \hfill
     \begin{subfigure}[b]{0.23\textwidth}
         \centering
         \includegraphics[width=\textwidth]{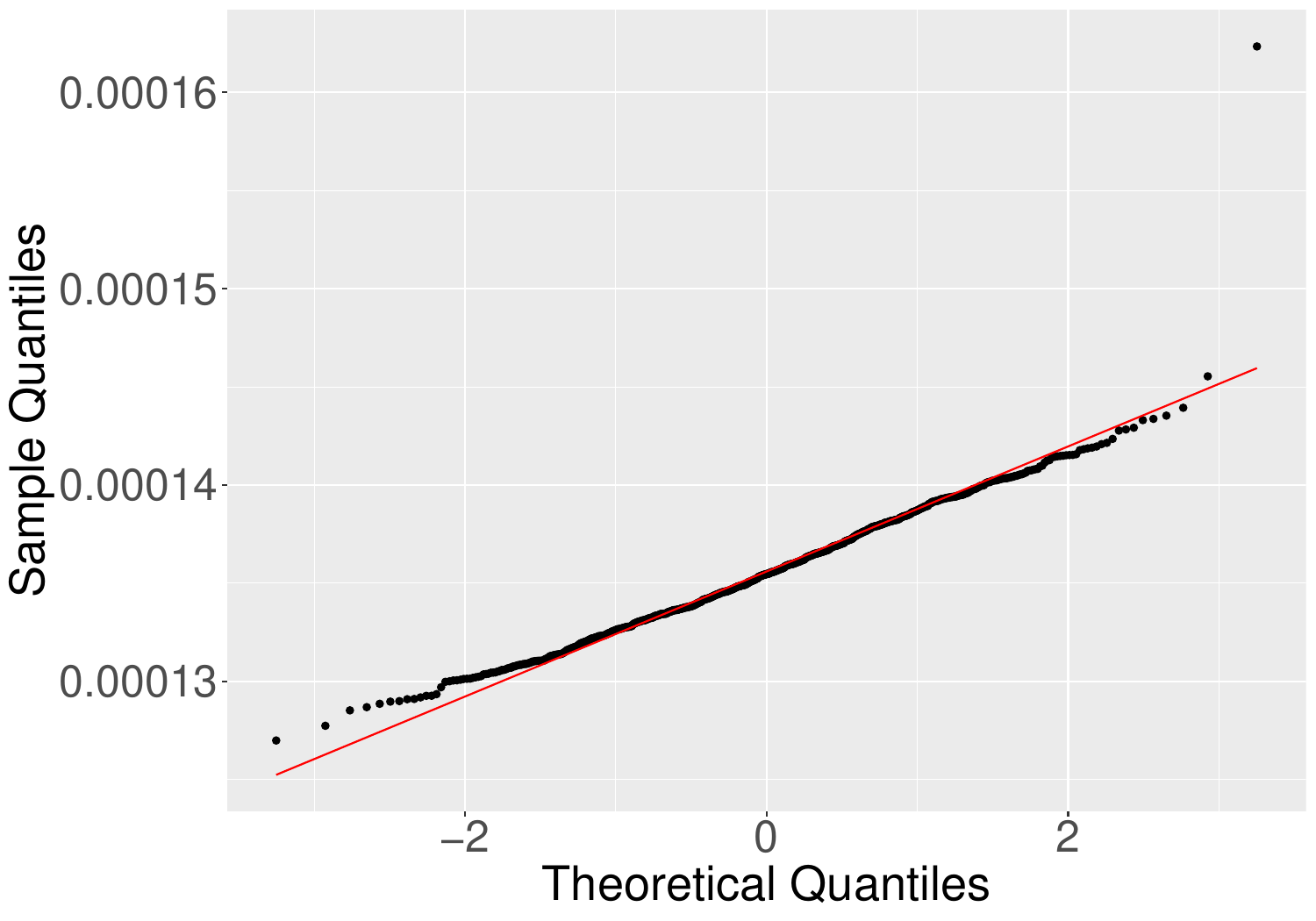}
         \caption{SDXL\_Lightning}
     \end{subfigure}
     \begin{subfigure}[b]{0.23\textwidth}
         \centering
         \includegraphics[width=\textwidth]{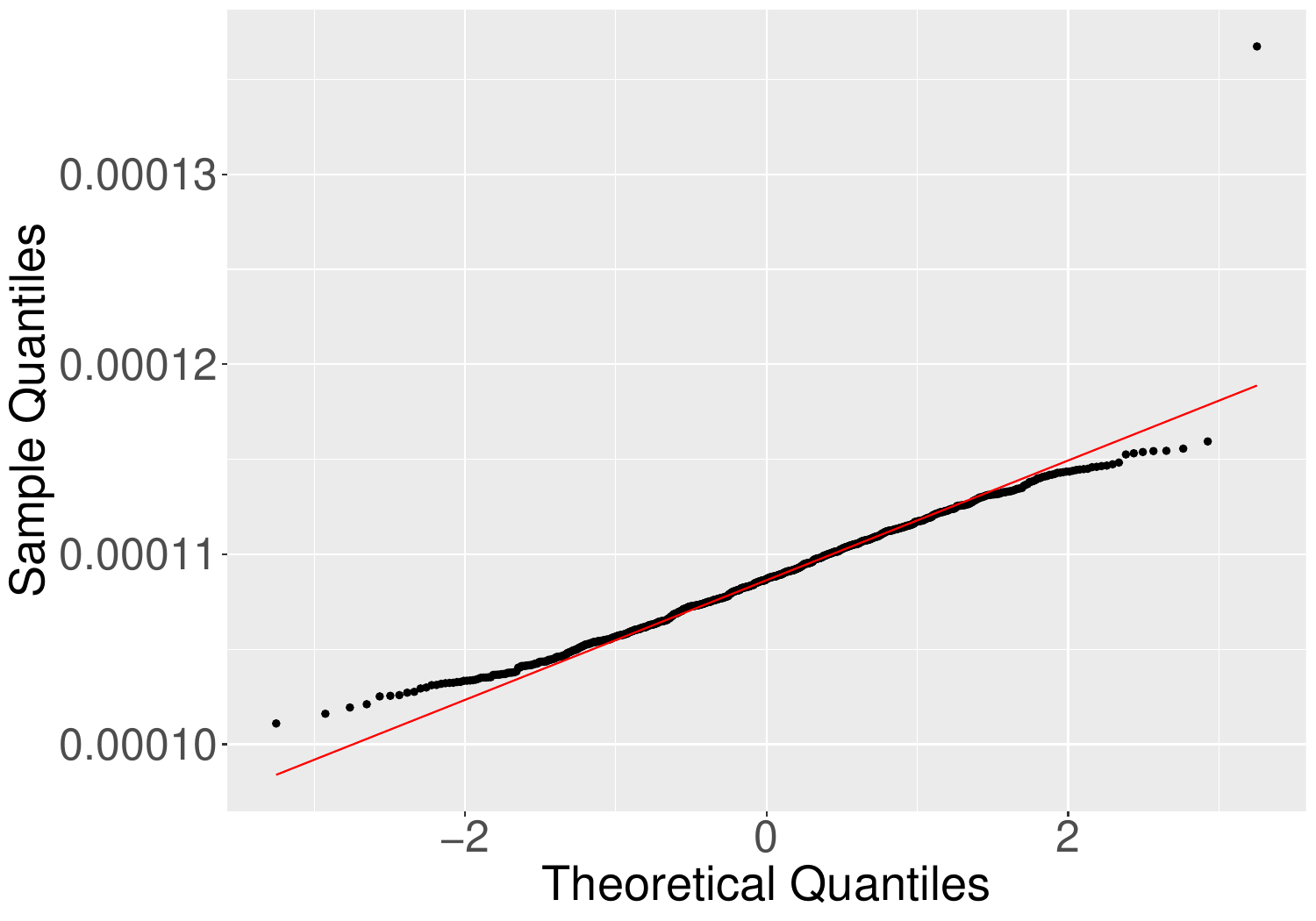}
         \caption{Hyper\_SD}
     \end{subfigure}
     \hfill
     \begin{subfigure}[b]{0.23\textwidth}
         \centering
         \includegraphics[width=\textwidth]{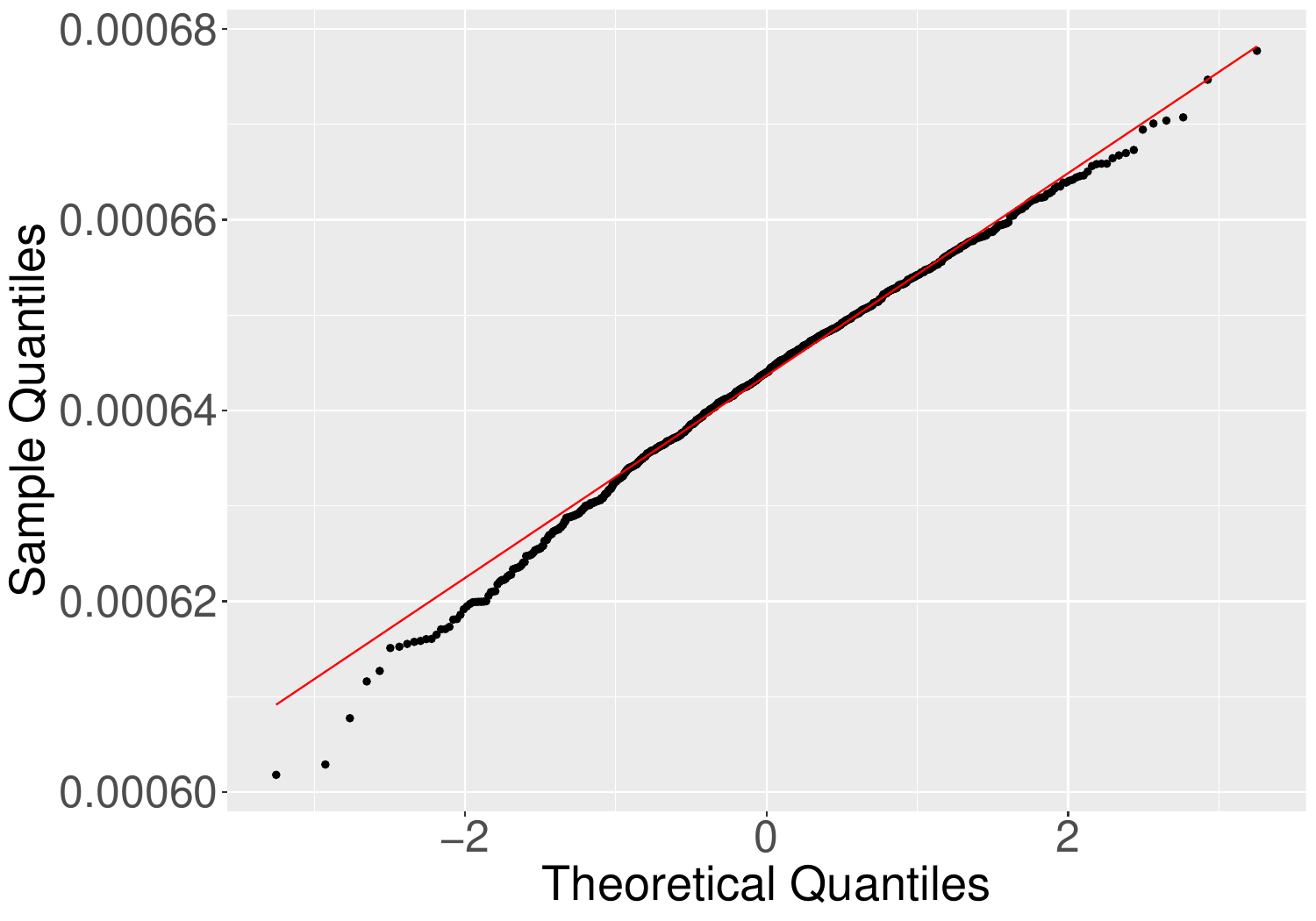}
         \caption{SSD\_1B}
     \end{subfigure}
     \hfill
     \begin{subfigure}[b]{0.23\textwidth}
         \centering
         \includegraphics[width=\textwidth]{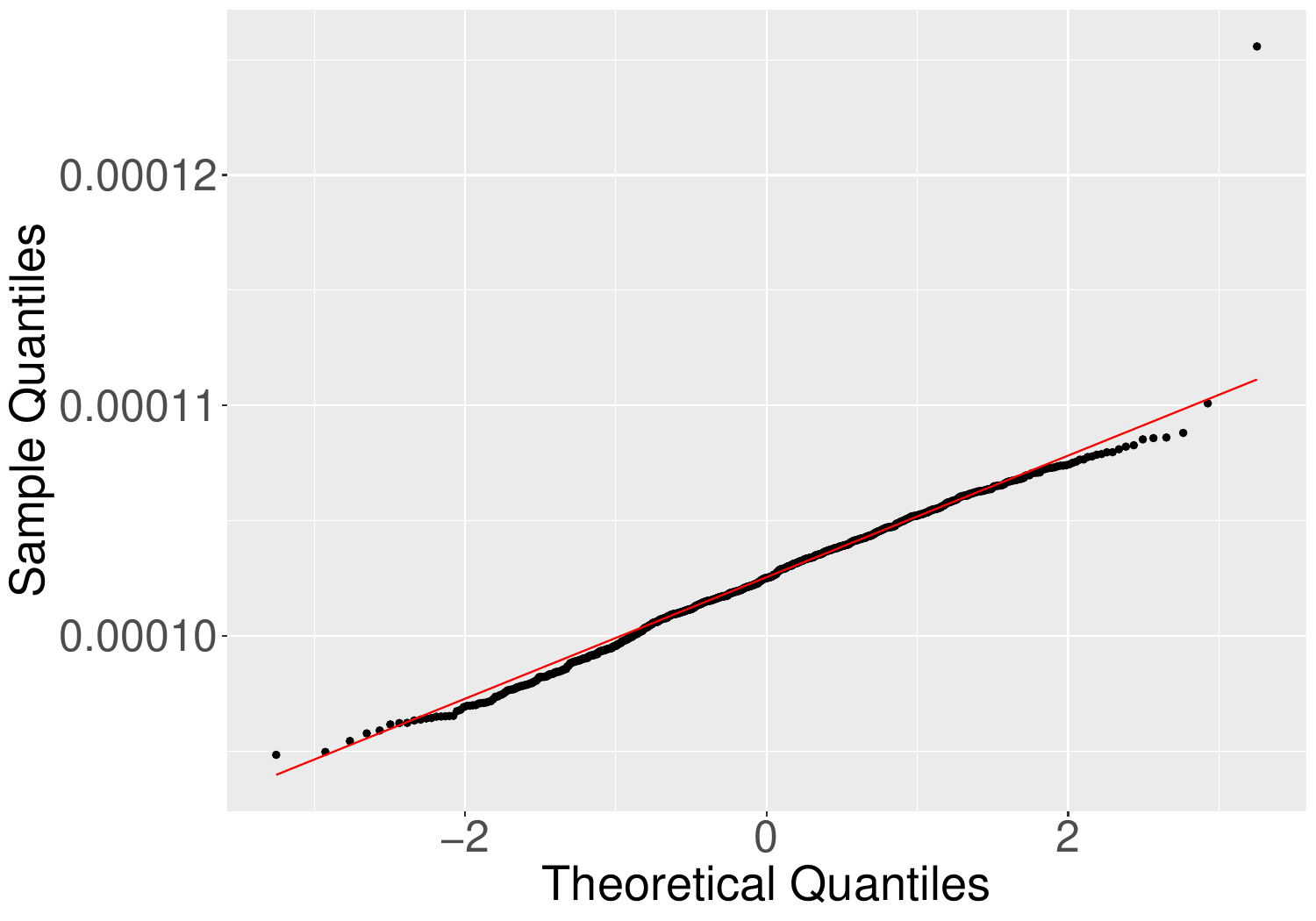}
         \caption{LCM\_SSD\_1B}
     \end{subfigure}
     \hfill
     \begin{subfigure}[b]{0.23\textwidth}
         \centering
         \includegraphics[width=\textwidth]{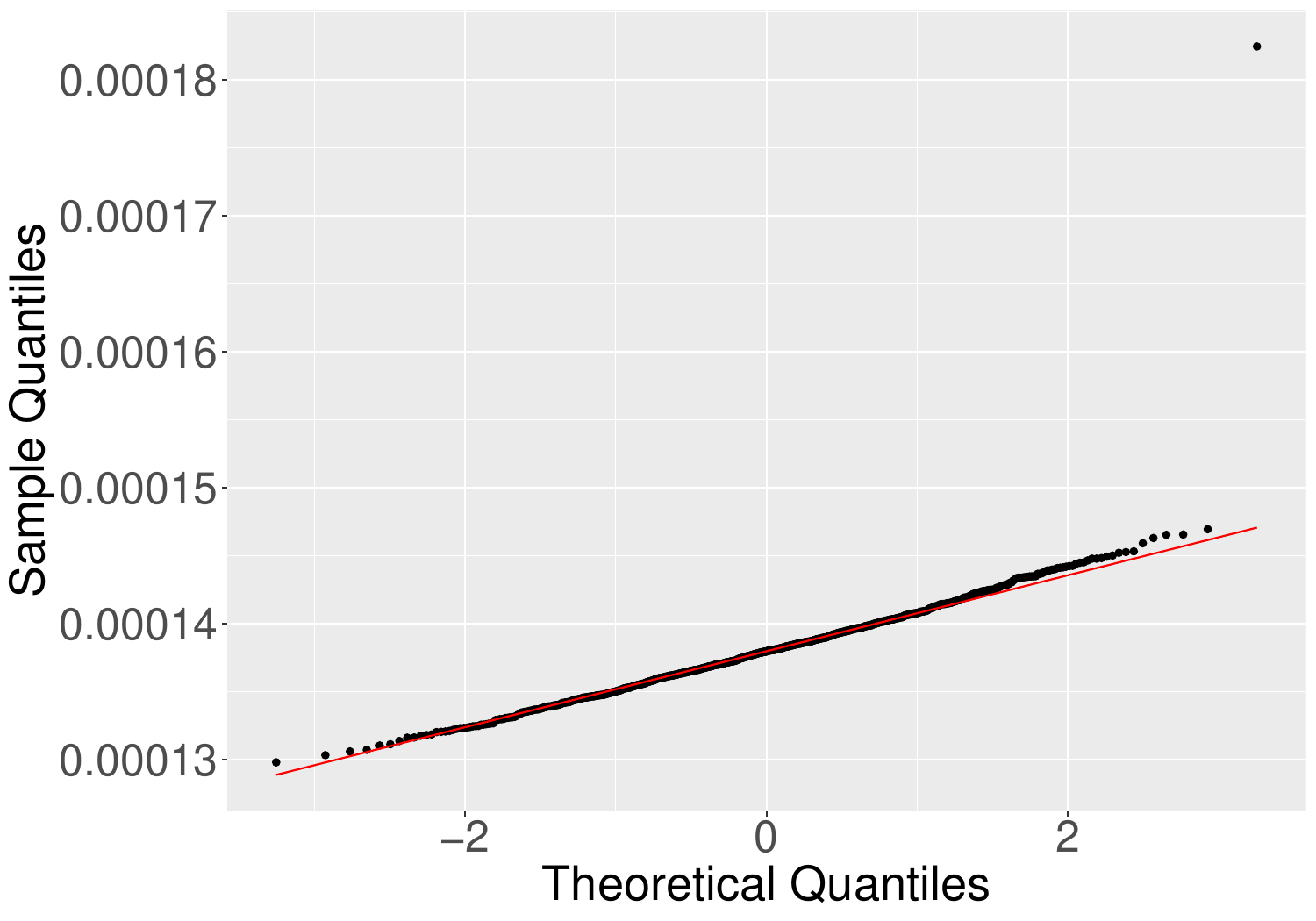}
         \caption{LCM\_SDXL}
     \end{subfigure}
     \begin{subfigure}[b]{0.23\textwidth}
         \centering
         \includegraphics[width=\textwidth]{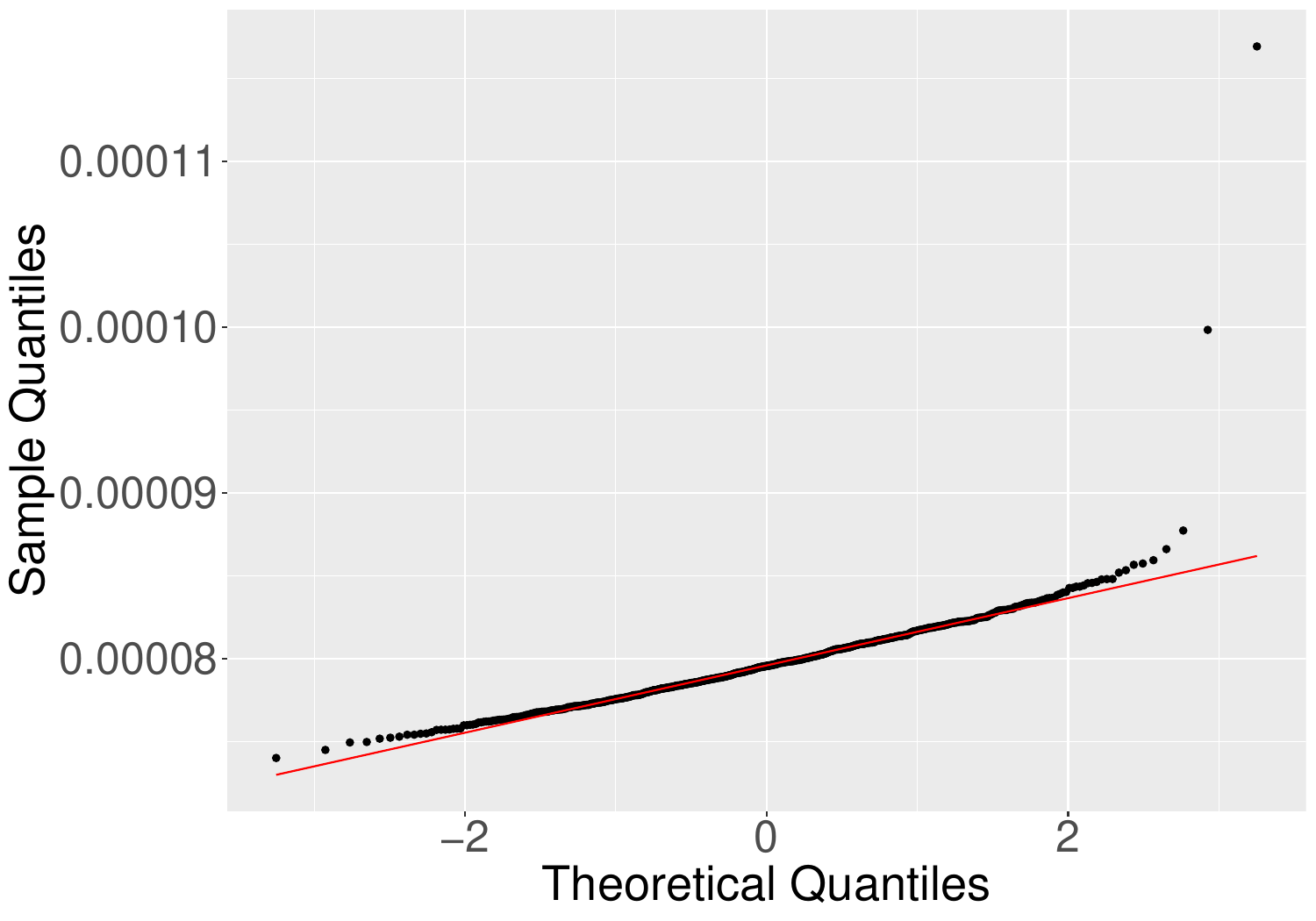}
         \caption{Flash\_SD}
     \end{subfigure}
     \hfill
     \begin{subfigure}[b]{0.23\textwidth}
         \centering
         \includegraphics[width=\textwidth]{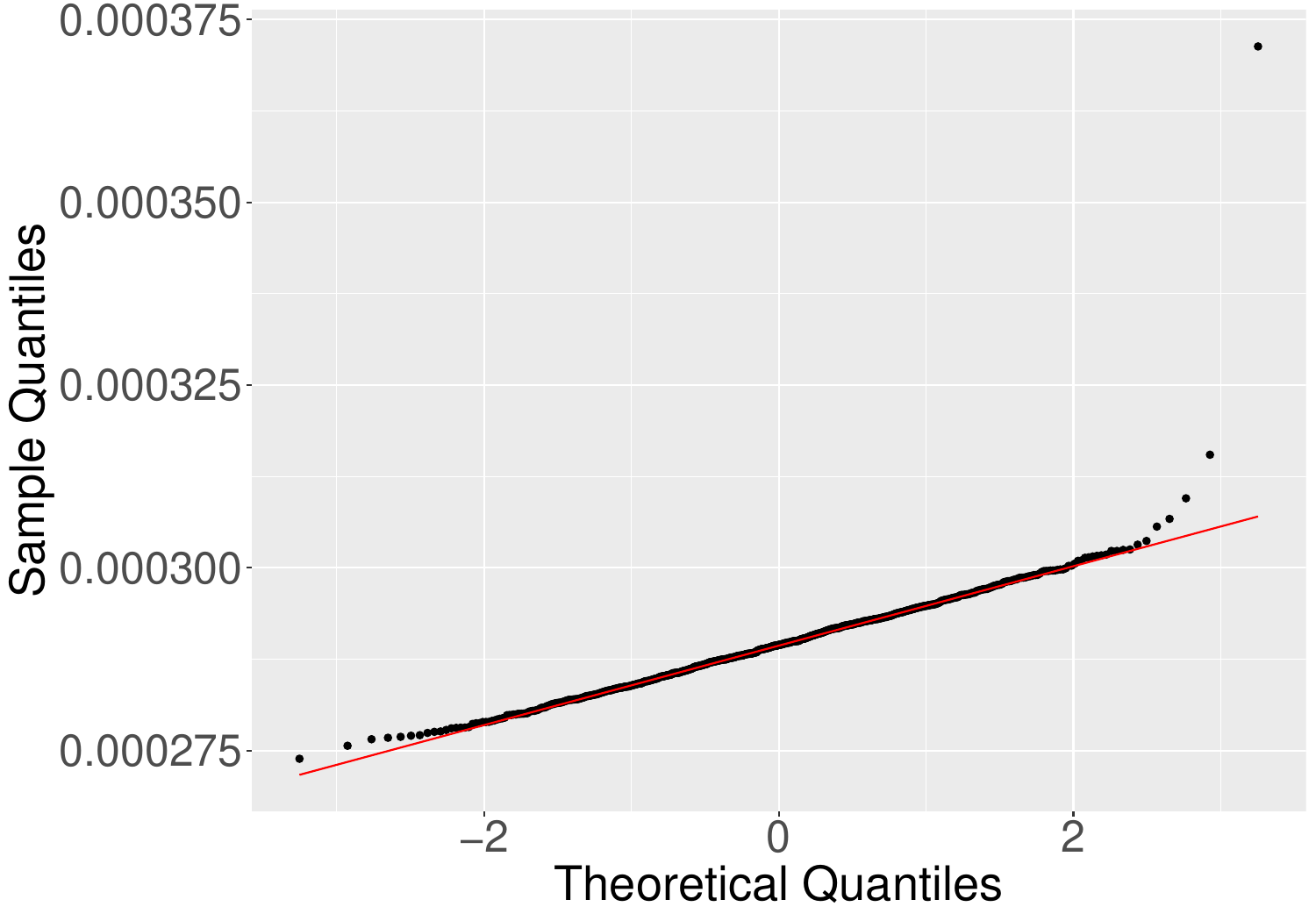}
         \caption{Flash\_SDXL}
     \end{subfigure}
     \hfill
     \begin{subfigure}[b]{0.23\textwidth}
         \centering
         \includegraphics[width=\textwidth]{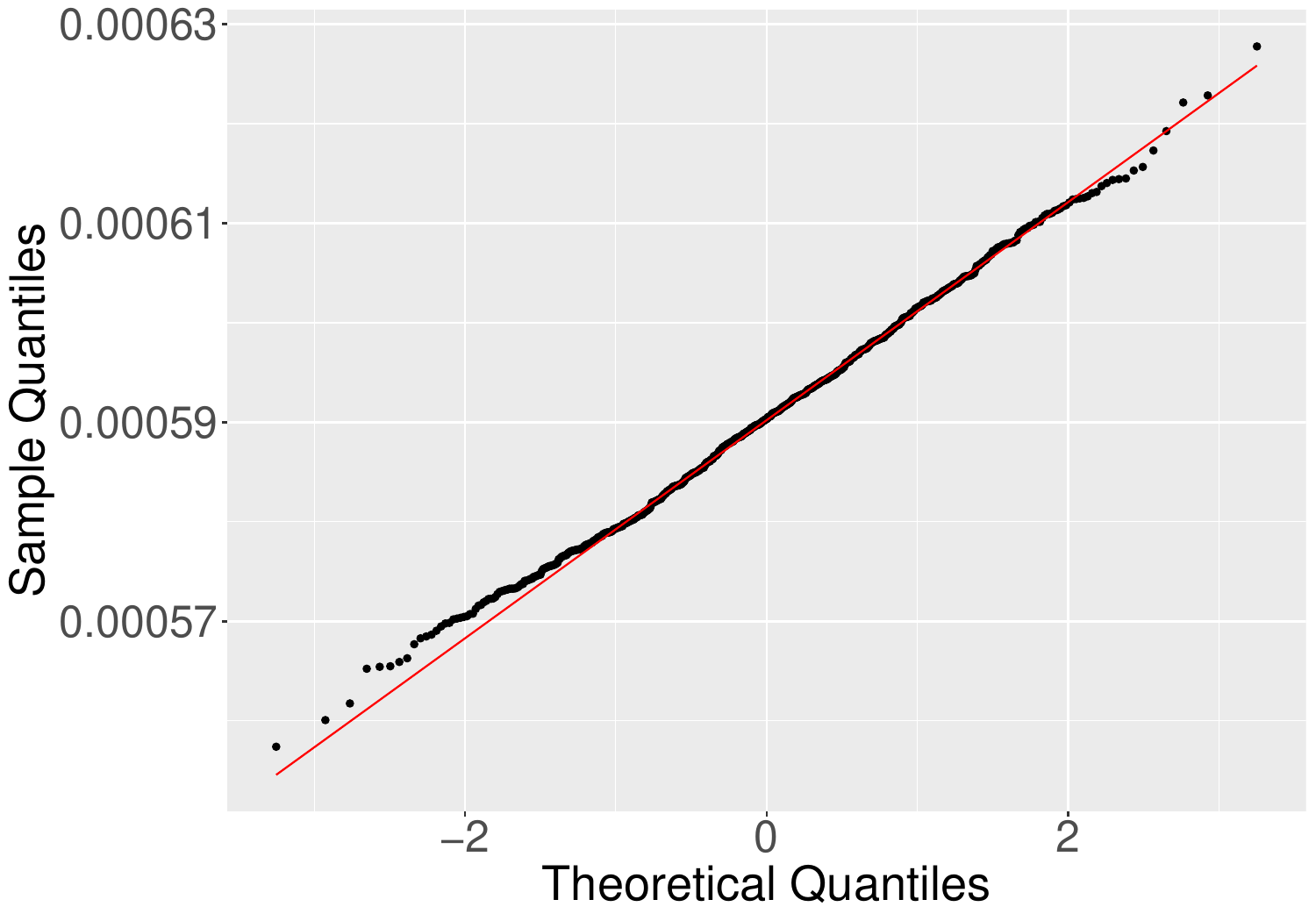}
         \caption{PixArt\_Alpha}
     \end{subfigure}
     \hfill
     \begin{subfigure}[b]{0.23\textwidth}
         \centering
         \includegraphics[width=\textwidth]{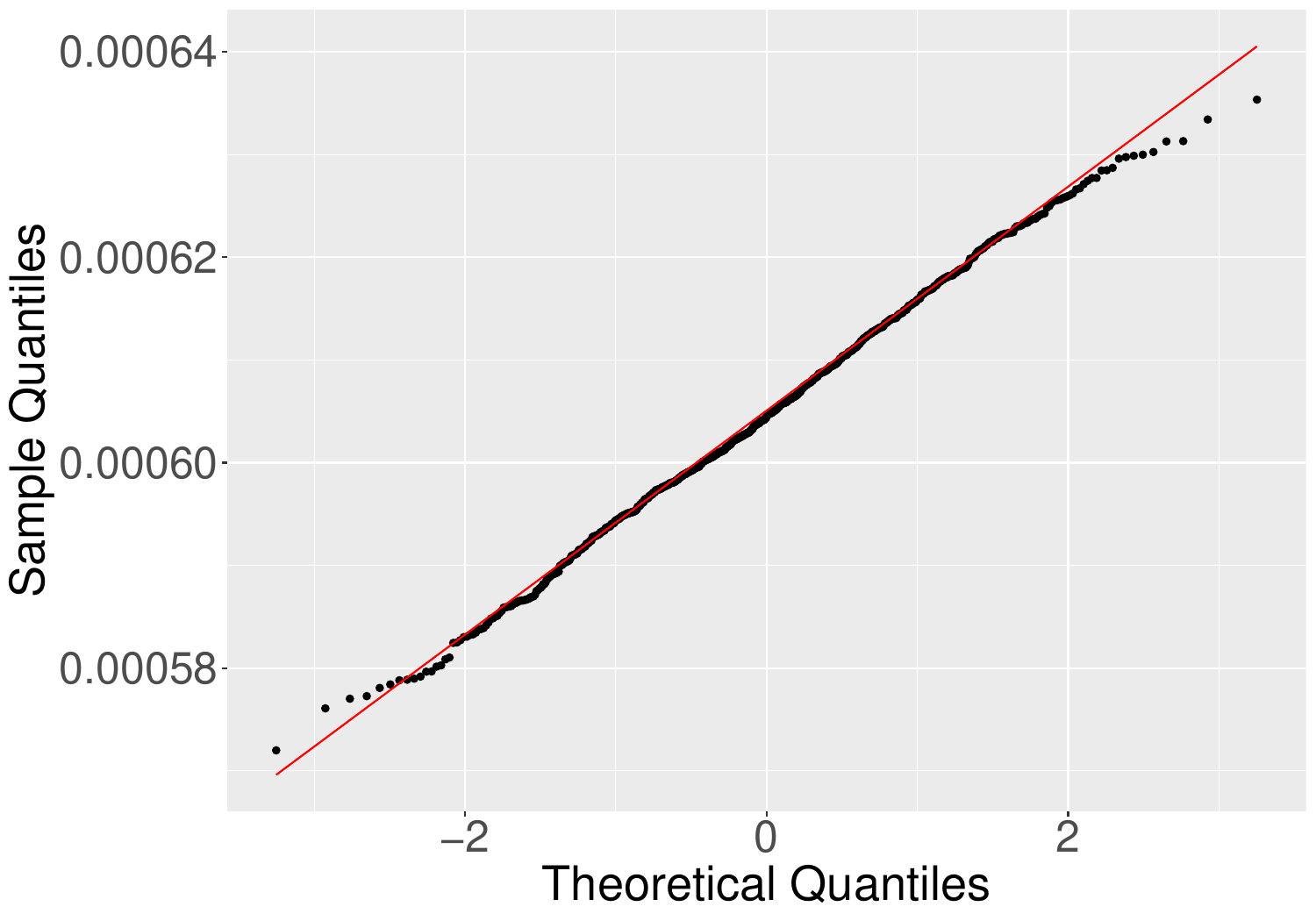}
         \caption{PixArt\_Sigma}
     \end{subfigure}
     \hfill
     \begin{subfigure}[b]{0.23\textwidth}
         \centering
         \includegraphics[width=\textwidth]{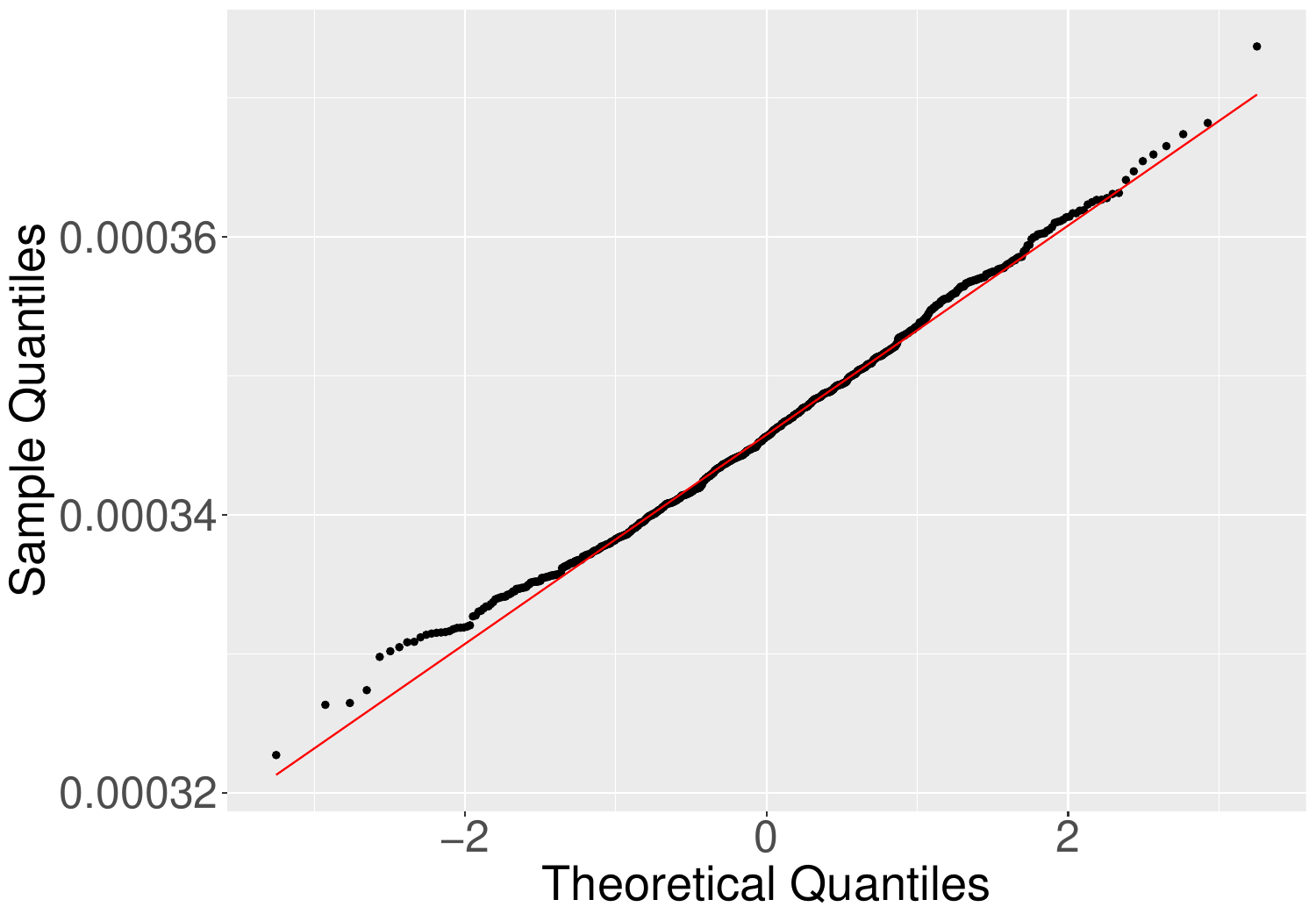}
         \caption{Flash\_PixArt}
     \end{subfigure}
     \hfill
     \begin{subfigure}[b]{0.23\textwidth}
         \centering
         \includegraphics[width=\textwidth]{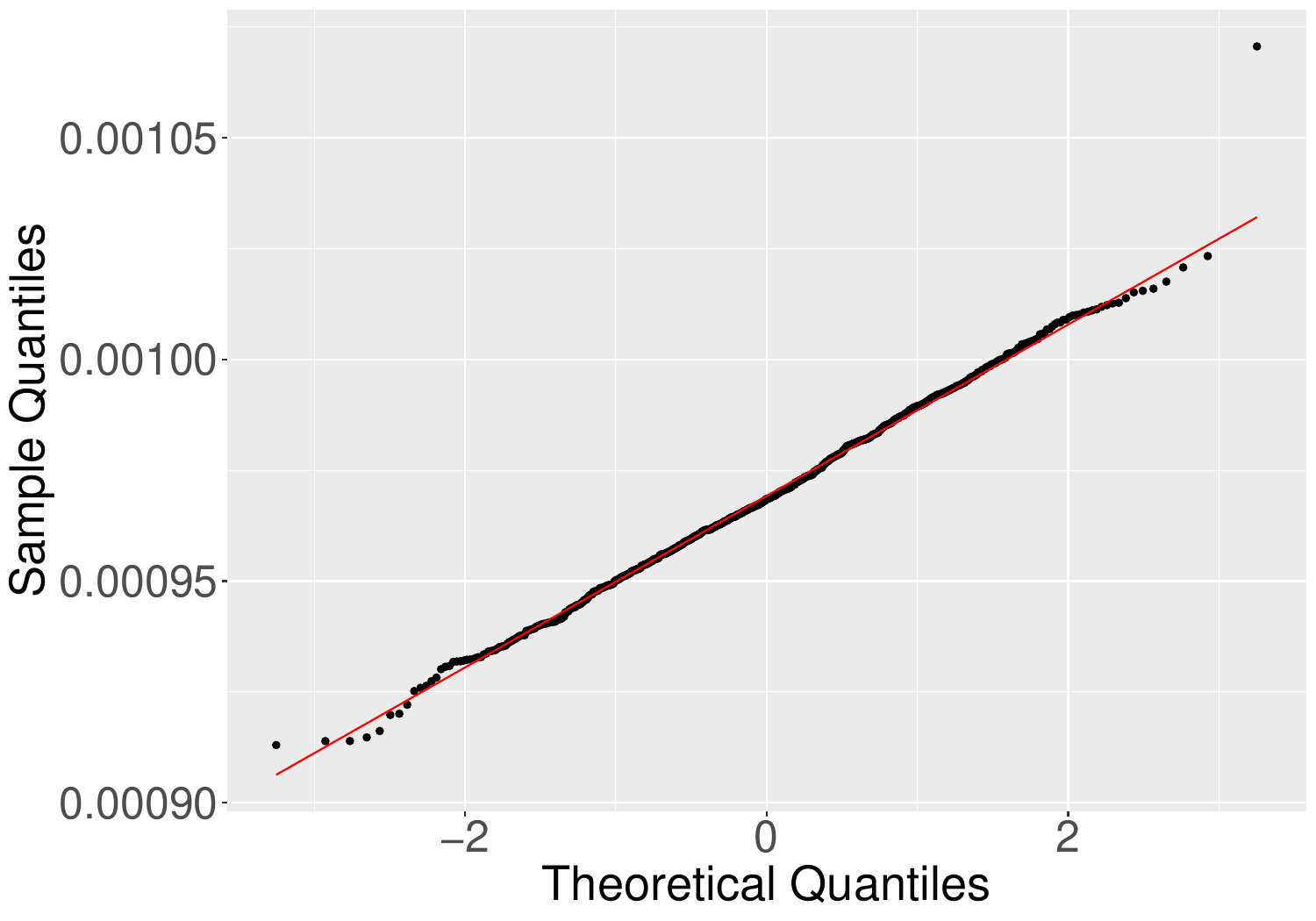}
         \caption{SD\_3}
     \end{subfigure}
     \hfill
     \begin{subfigure}[b]{0.23\textwidth}
         \centering
         \includegraphics[width=\textwidth]{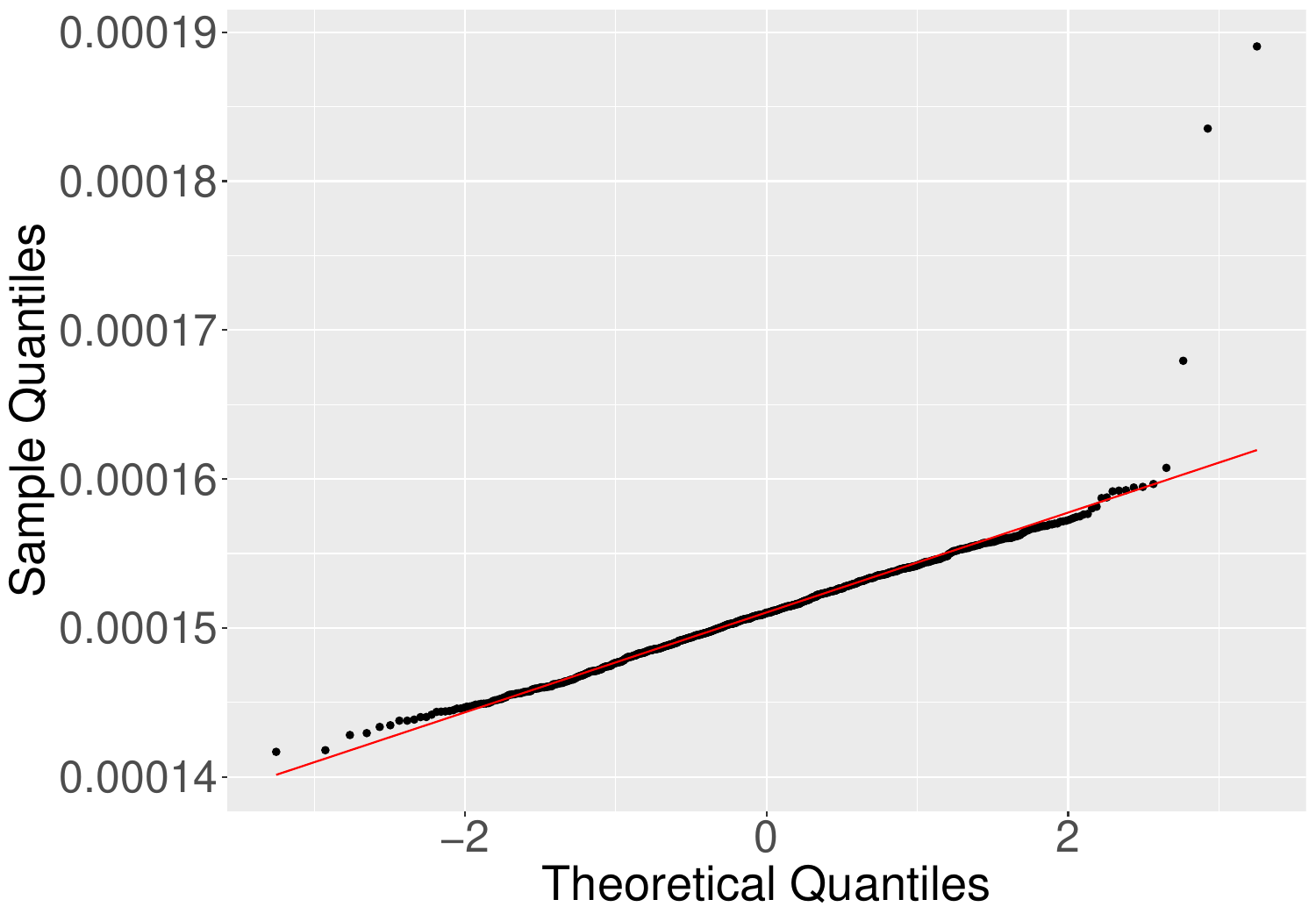}
         \caption{Flash\_SD3}
     \end{subfigure}
     \hfill
     \begin{subfigure}[b]{0.23\textwidth}
         \centering
         \includegraphics[width=\textwidth]{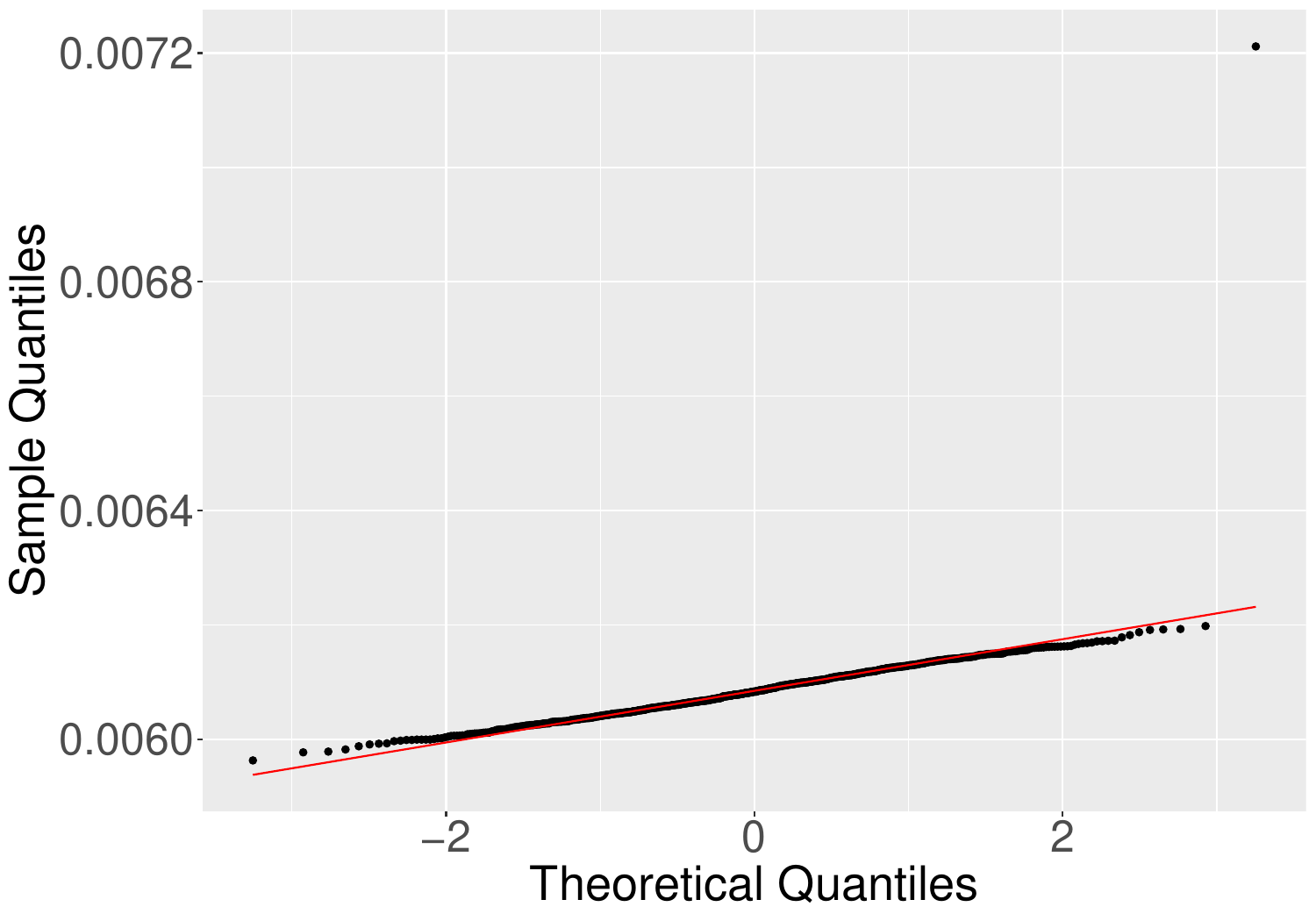}
         \caption{Lumina}
     \end{subfigure}
     \hfill
     \begin{subfigure}[b]{0.23\textwidth}
         \centering
         \includegraphics[width=\textwidth]{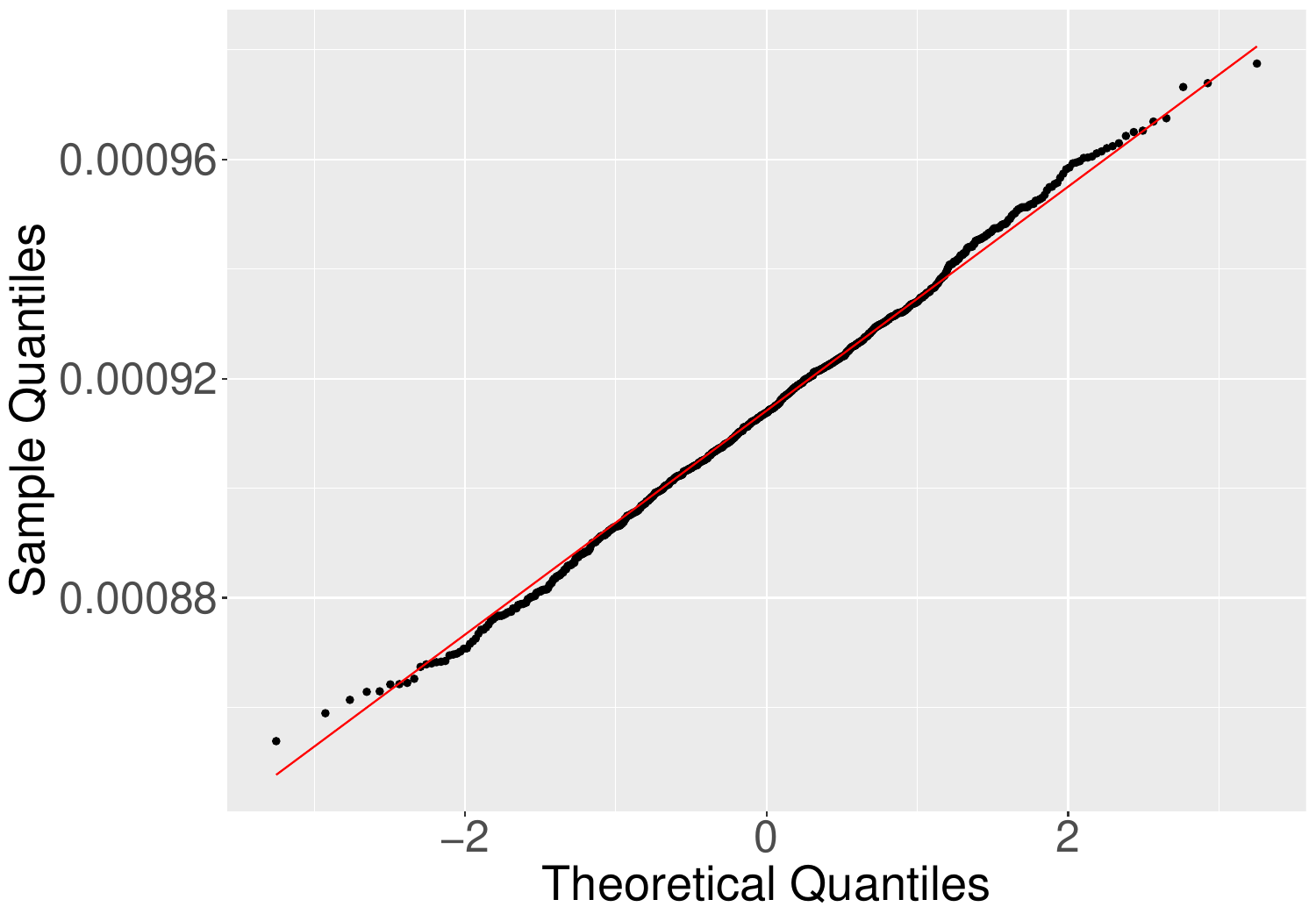}
         \caption{Flux\_1}
     \end{subfigure}
        \caption{Q-Q plots comparing the energy consumption of the analyzed models across various prompt contents to the normal distribution. Each plot visualizes the distribution of energy consumption values for a specific model, assessing how closely they follow a normal distribution.}
        \label{fig:qqplot_content}
\end{figure}

The results of the correlation between the semantic content of prompts and the energy consumed during the image generation process are presented in \Cref{tab:prompt_content}.
Before performing the correlation analysis, we verified that the energy consumption for each model followed a normal distribution, by examining Quantile-Quantile (QQ) plots, which are illustrated in \Cref{fig:qqplot_content}. 
If the data closely follows the normal distribution, the points on the Q-Q plot will move on a diagonal line (depicted in red), while deviations from the reference line indicate departures from the expected distribution.
\Cref{fig:qqplot_content} shows that for all the analyzed models, data follows a normal distribution.
This step ensured the validity of applying Pearson’s correlation coefficient, which relies on the assumption of normally distributed data.

\begin{figure}[p]
    \centering
    \begin{subfigure}[b]{0.45\textwidth}
        \centering
        \includegraphics[width=\textwidth]{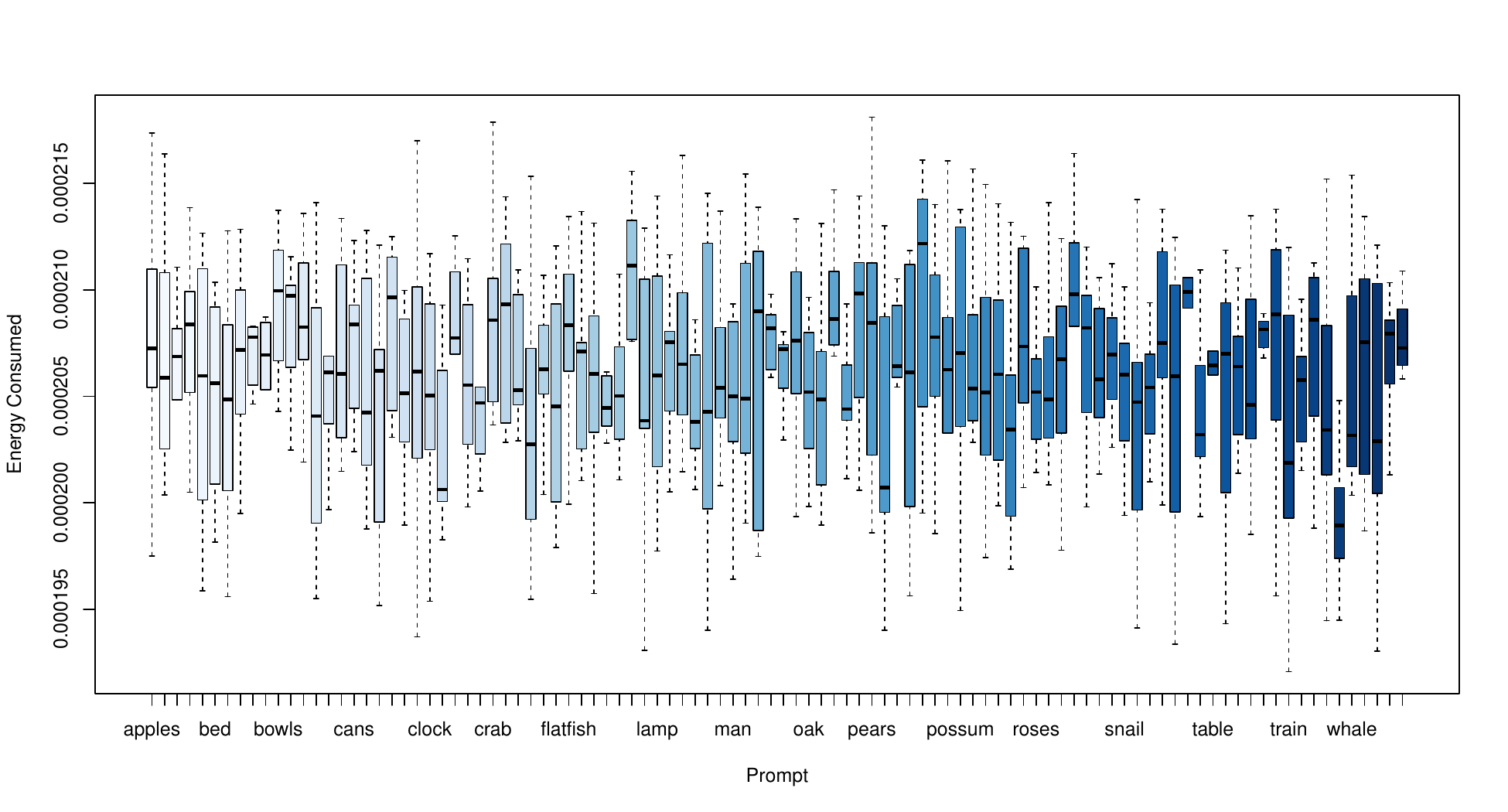}
        \caption{SD\_1.5}
    \end{subfigure}
    \hfill
    \begin{subfigure}[b]{0.45\textwidth}
        \centering
        \includegraphics[width=\textwidth]{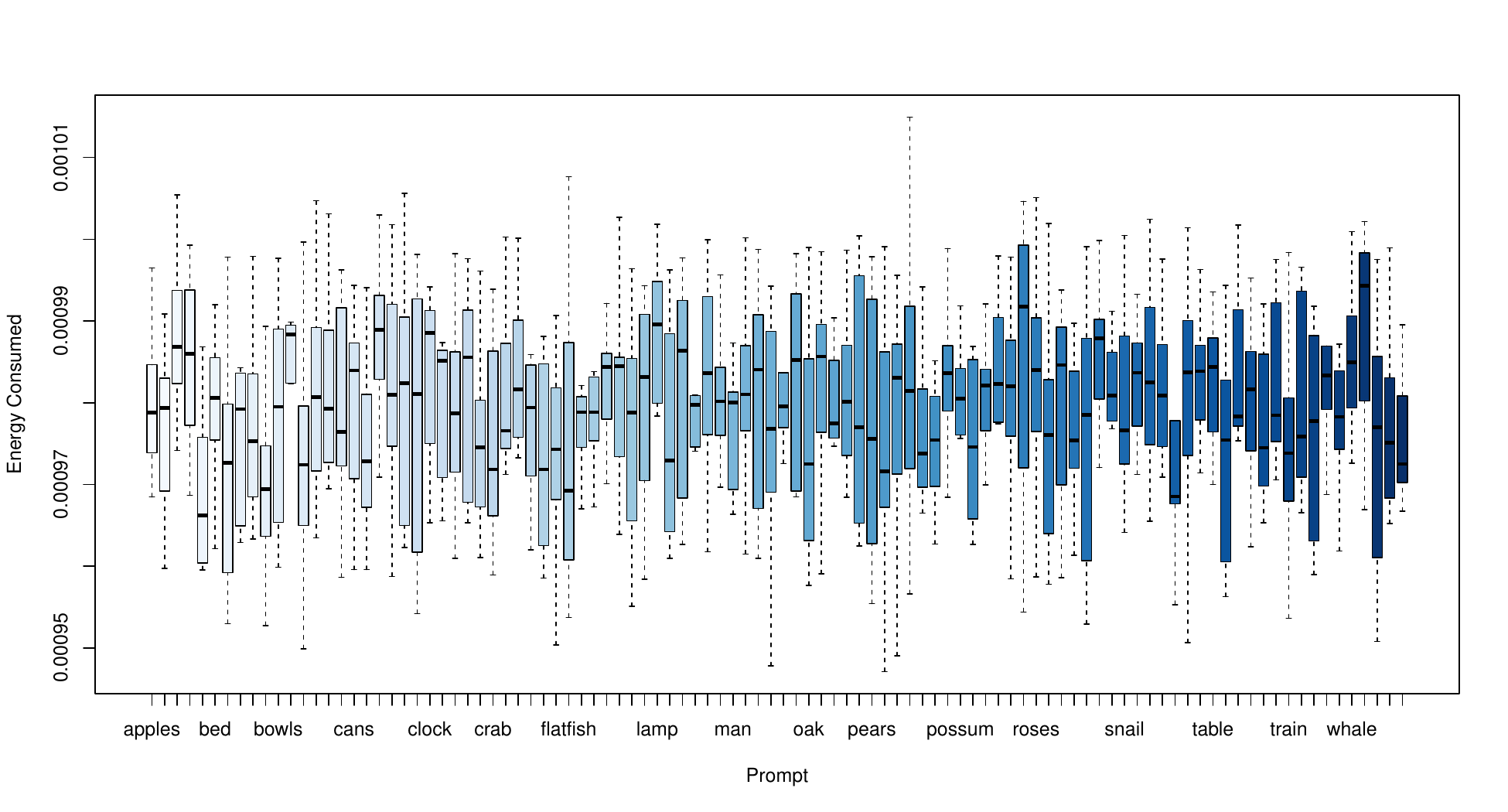}
        \caption{SDXL}
    \end{subfigure}
    \hfill
    \begin{subfigure}[b]{0.45\textwidth}
        \centering
        \includegraphics[width=\textwidth]{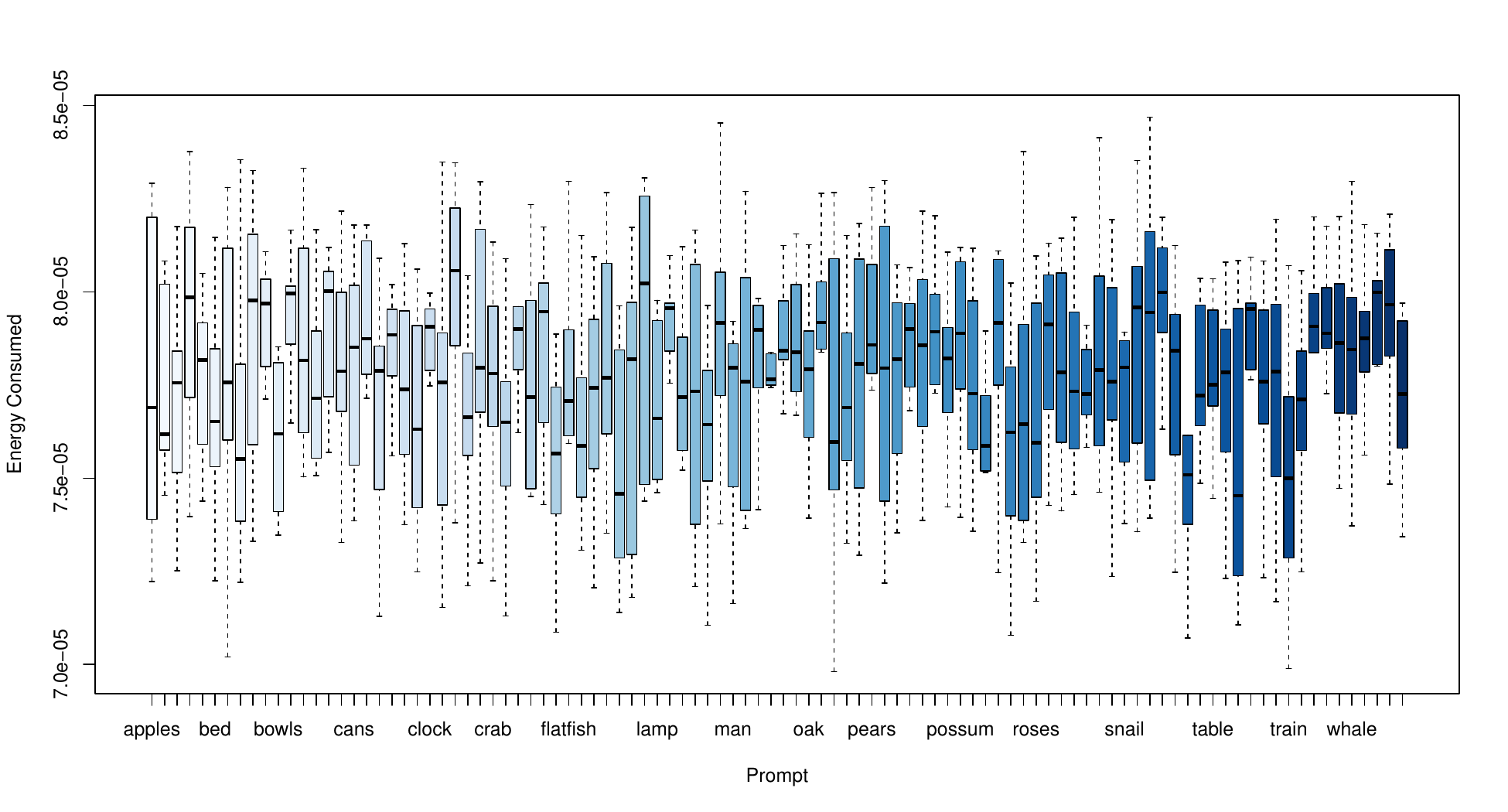}
        \caption{SDXL\_Turbo}
    \end{subfigure}
    \hfill
    \begin{subfigure}[b]{0.45\textwidth}
        \centering
        \includegraphics[width=\textwidth]{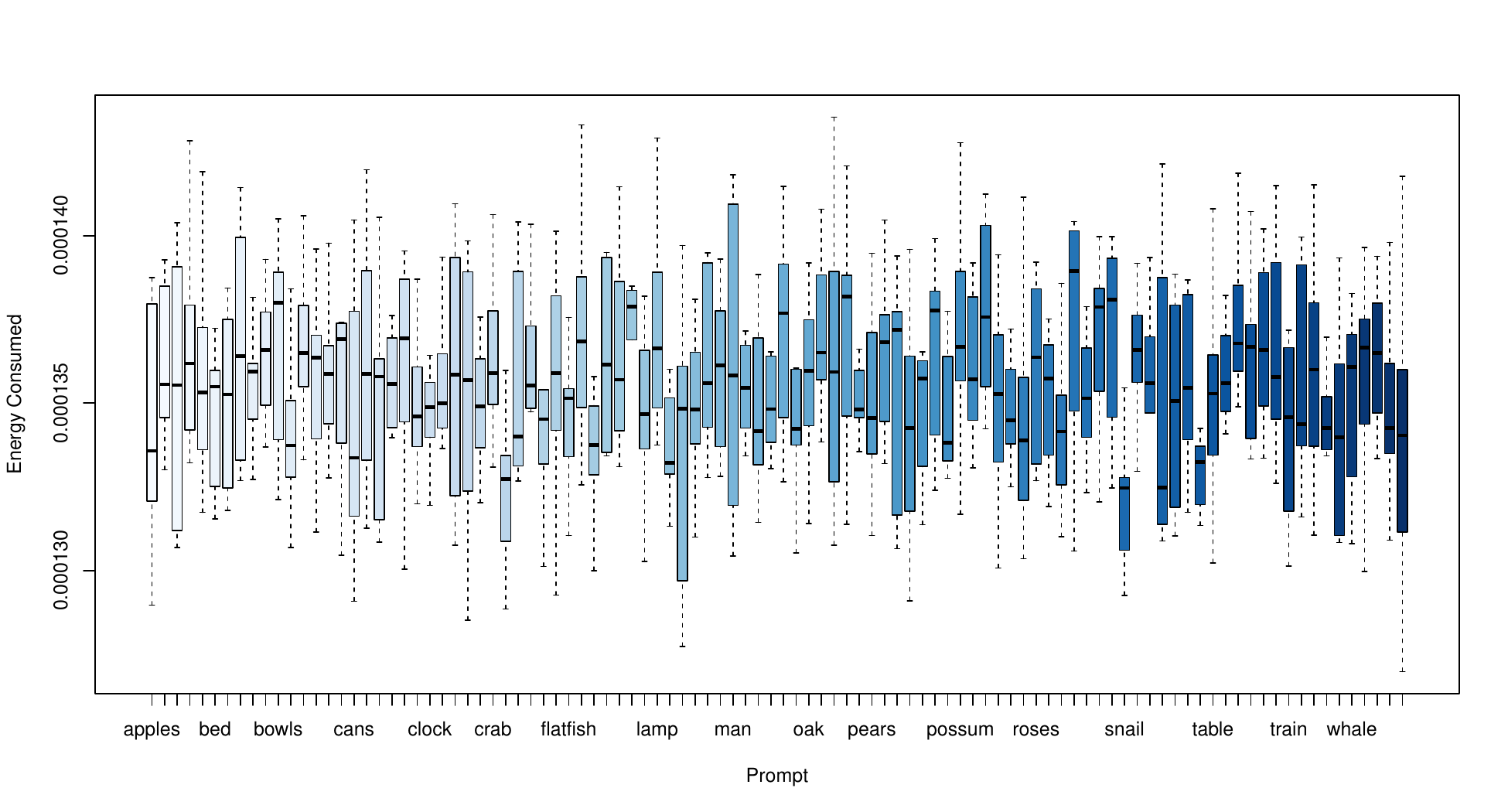}
        \caption{SDXL\_Lightning}
    \end{subfigure}
    \hfill
    \begin{subfigure}[b]{0.45\textwidth}
        \centering
        \includegraphics[width=\textwidth]{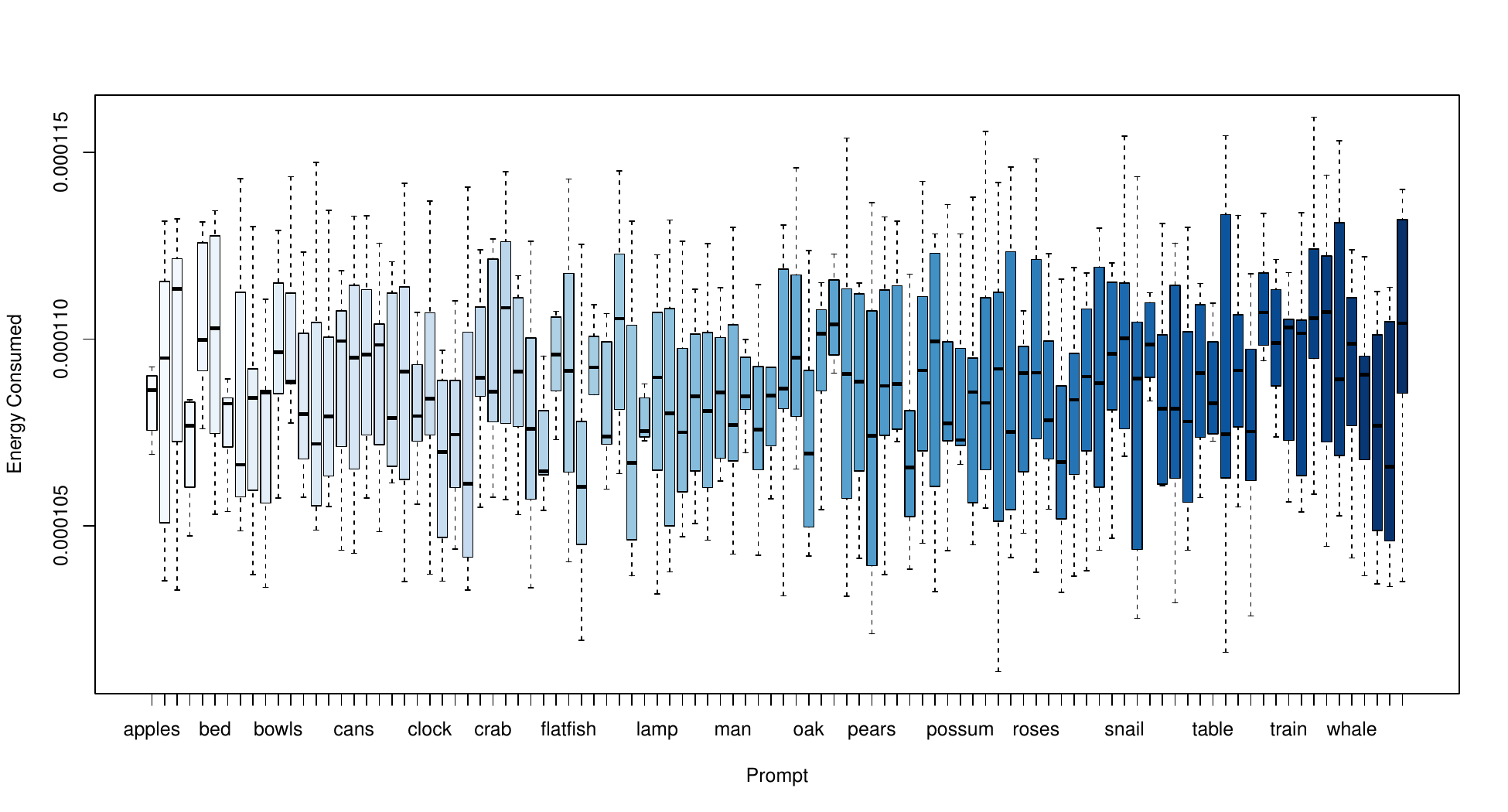}
        \caption{Hyper\_SD}
    \end{subfigure}
    \hfill
    \begin{subfigure}[b]{0.45\textwidth}
        \centering
        \includegraphics[width=\textwidth]{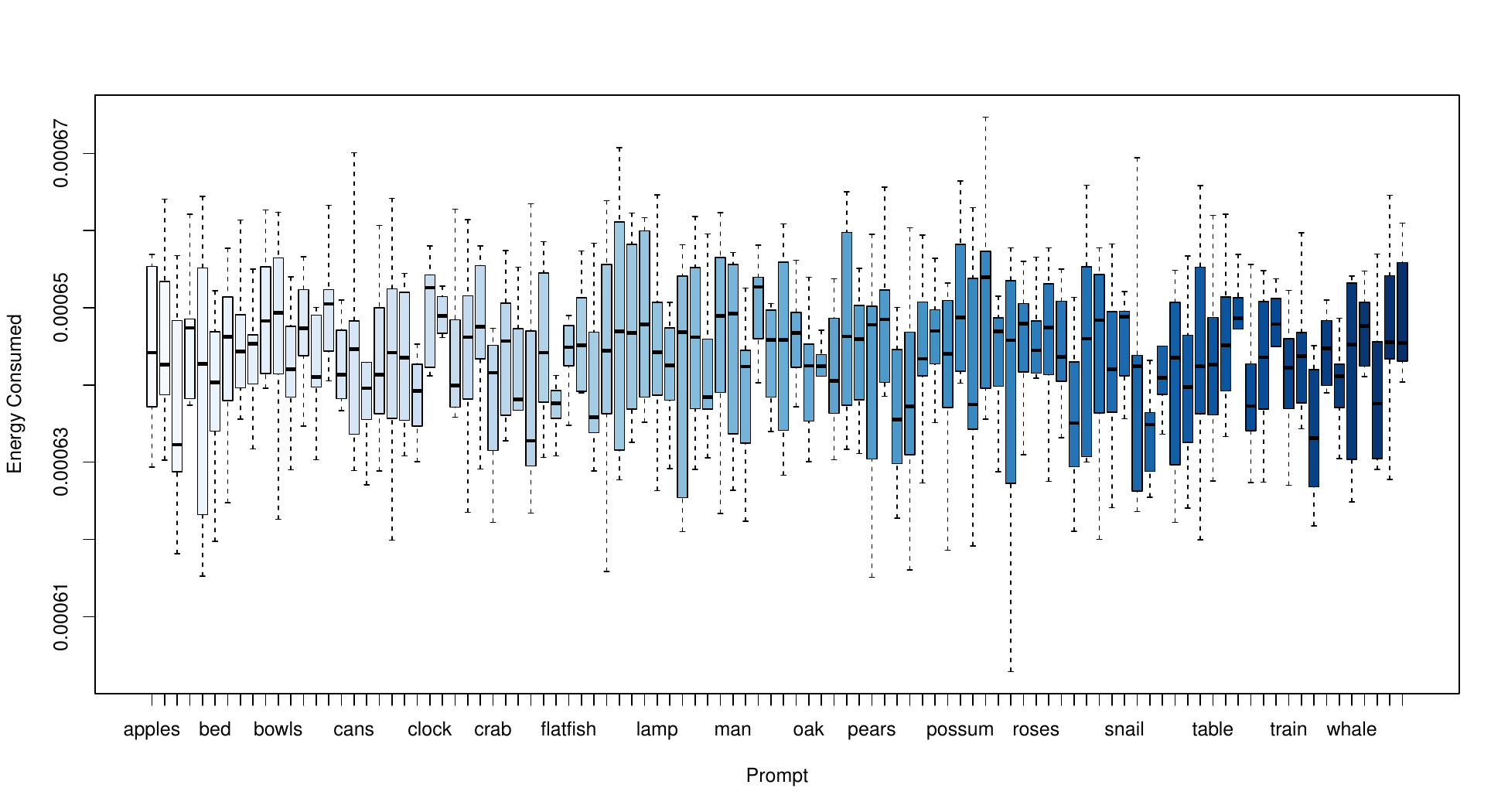}
        \caption{SSD\_1B}
    \end{subfigure}
    \caption{Box plots of the energy consumption across various prompt lengths for the analyzed models. (Group 1).}
    \label{fig:boxplot_content_g1}
\end{figure}

\begin{figure}[p]
    \centering
    \begin{subfigure}[b]{0.45\textwidth}
         \centering
         \includegraphics[width=\textwidth]{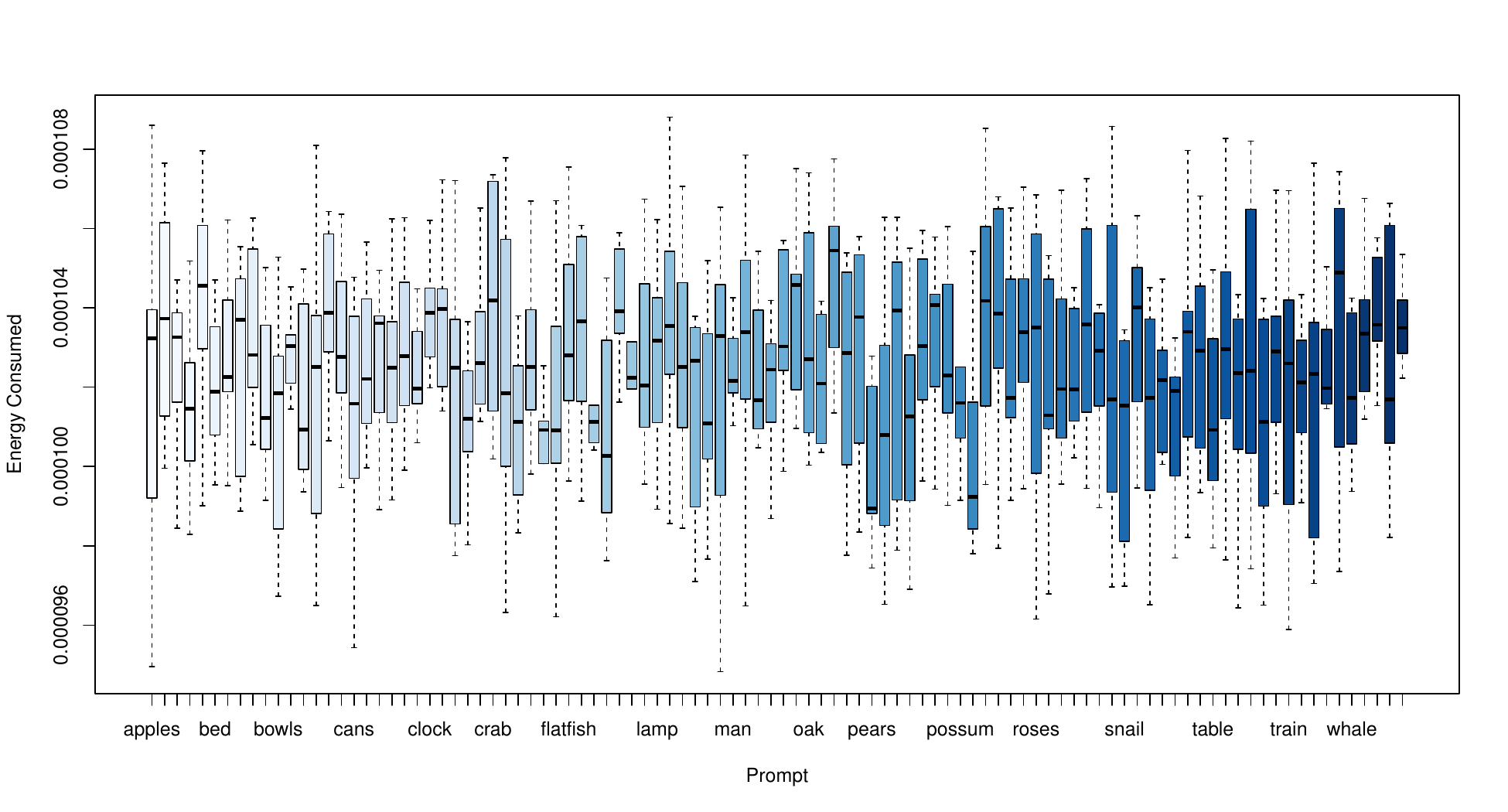}
         \caption{LCM\_SSD\_1B}
     \end{subfigure}
     \hfill
     \begin{subfigure}[b]{0.45\textwidth}
         \centering
         \includegraphics[width=\textwidth]{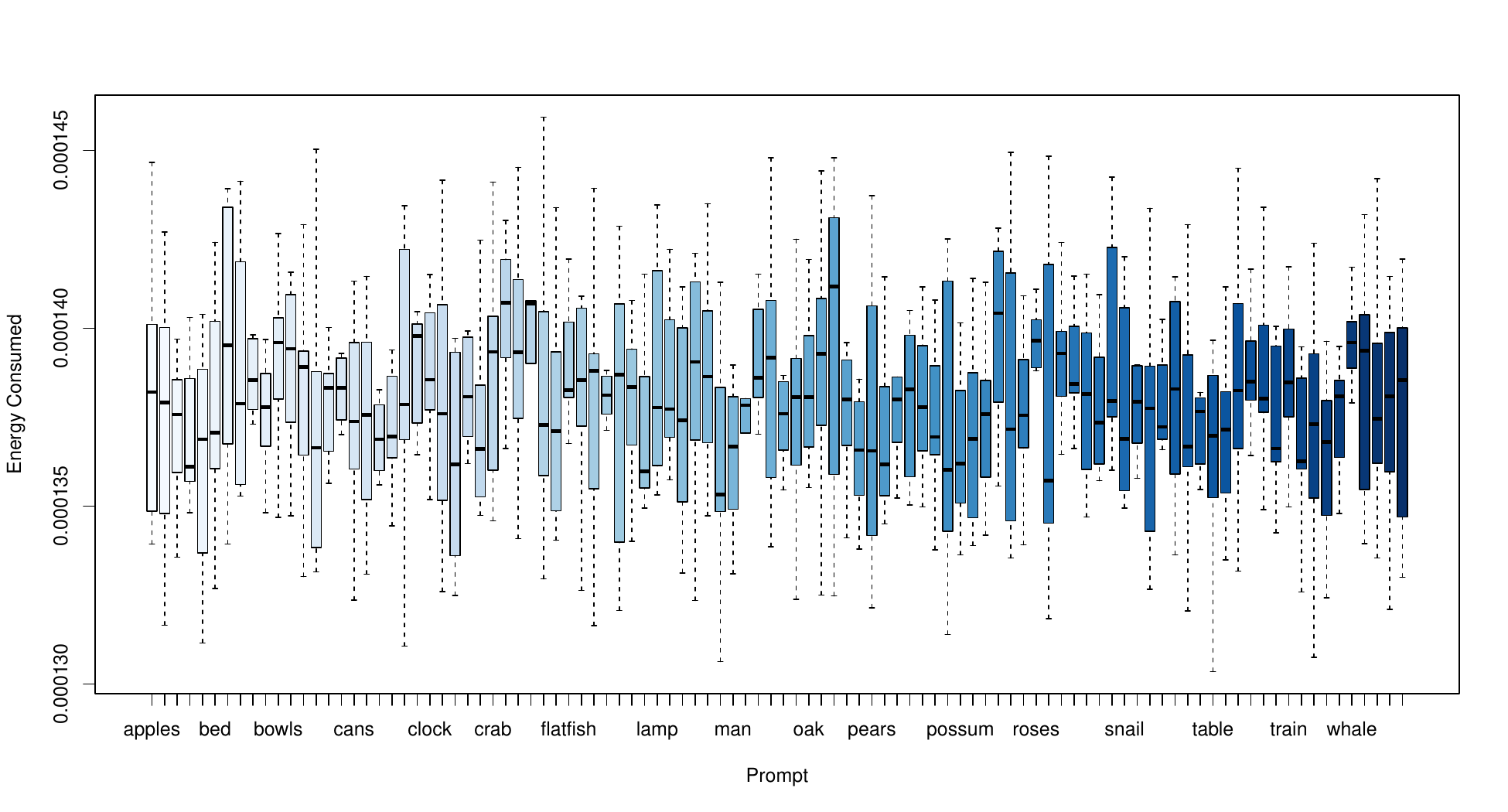}
         \caption{LCM\_SDXL}
     \end{subfigure}
     \begin{subfigure}[b]{0.45\textwidth}
         \centering
         \includegraphics[width=\textwidth]{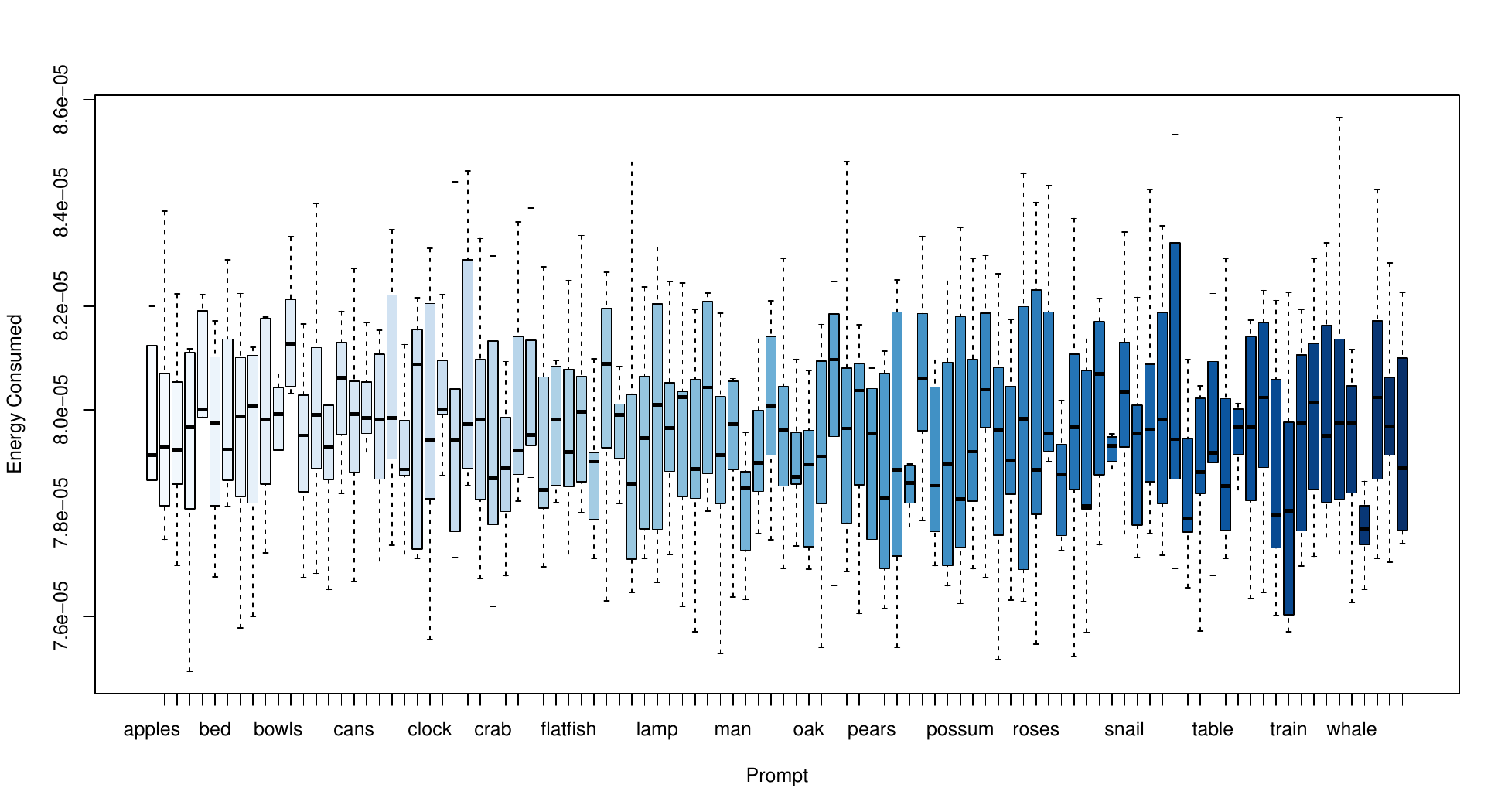}
         \caption{Flash\_SD}
     \end{subfigure}
     \hfill
     \begin{subfigure}[b]{0.45\textwidth}
         \centering
         \includegraphics[width=\textwidth]{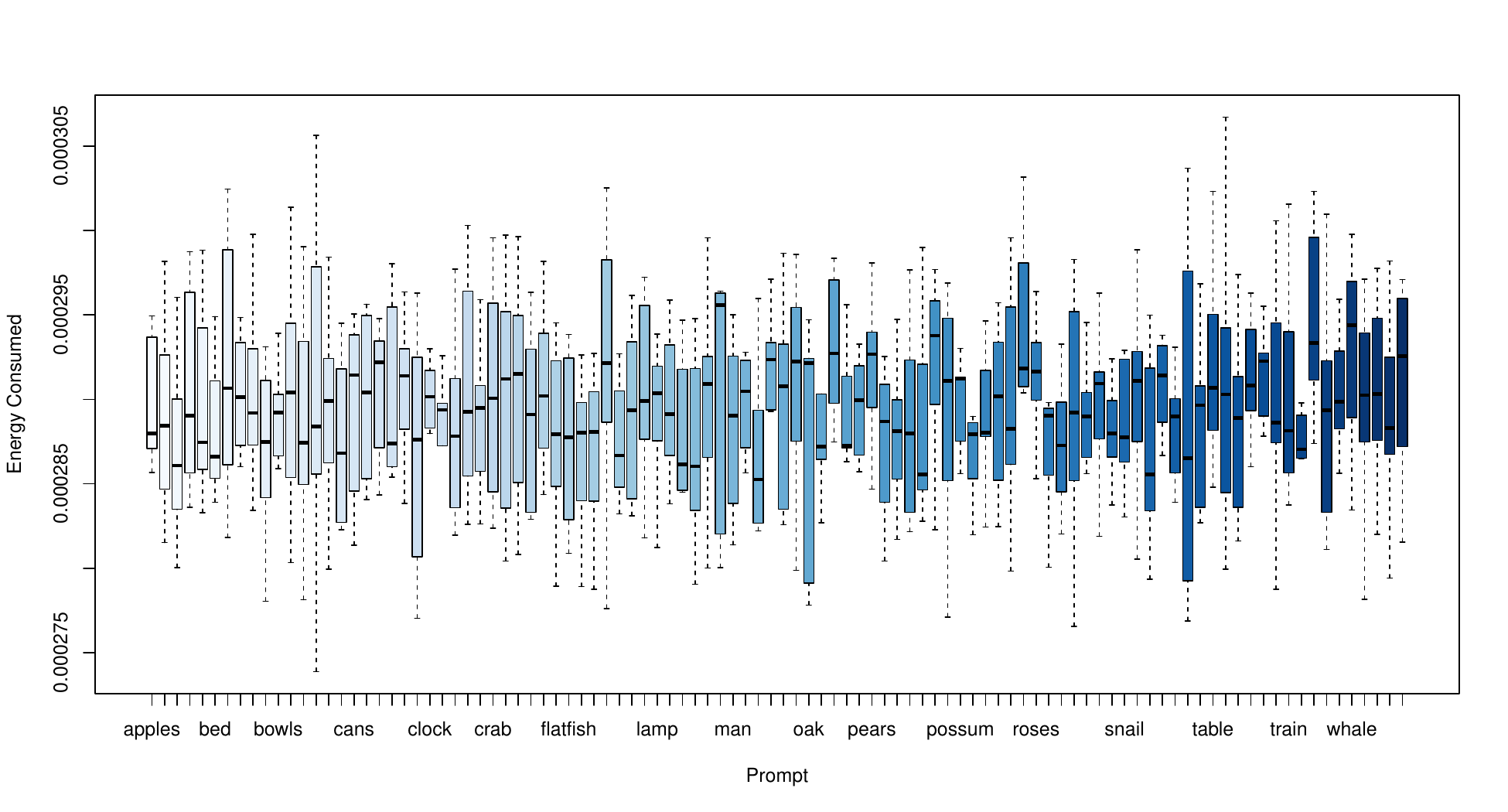}
         \caption{Flash\_SDXL}
     \end{subfigure}
     \hfill
     \begin{subfigure}[b]{0.45\textwidth}
         \centering
         \includegraphics[width=\textwidth]{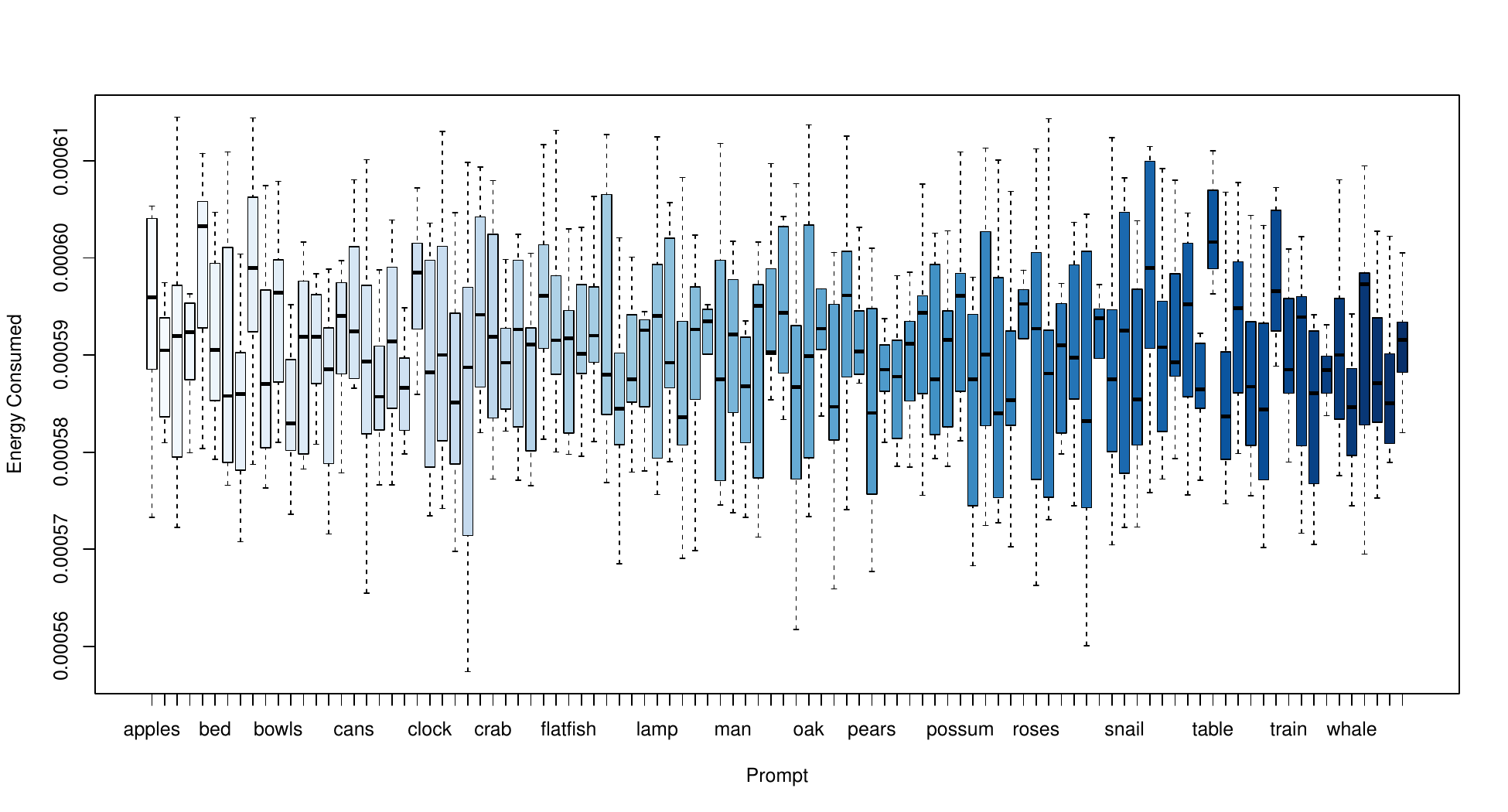}
         \caption{PixArt\_Alpha}
     \end{subfigure}
     \hfill
     \begin{subfigure}[b]{0.45\textwidth}
         \centering
         \includegraphics[width=\textwidth]{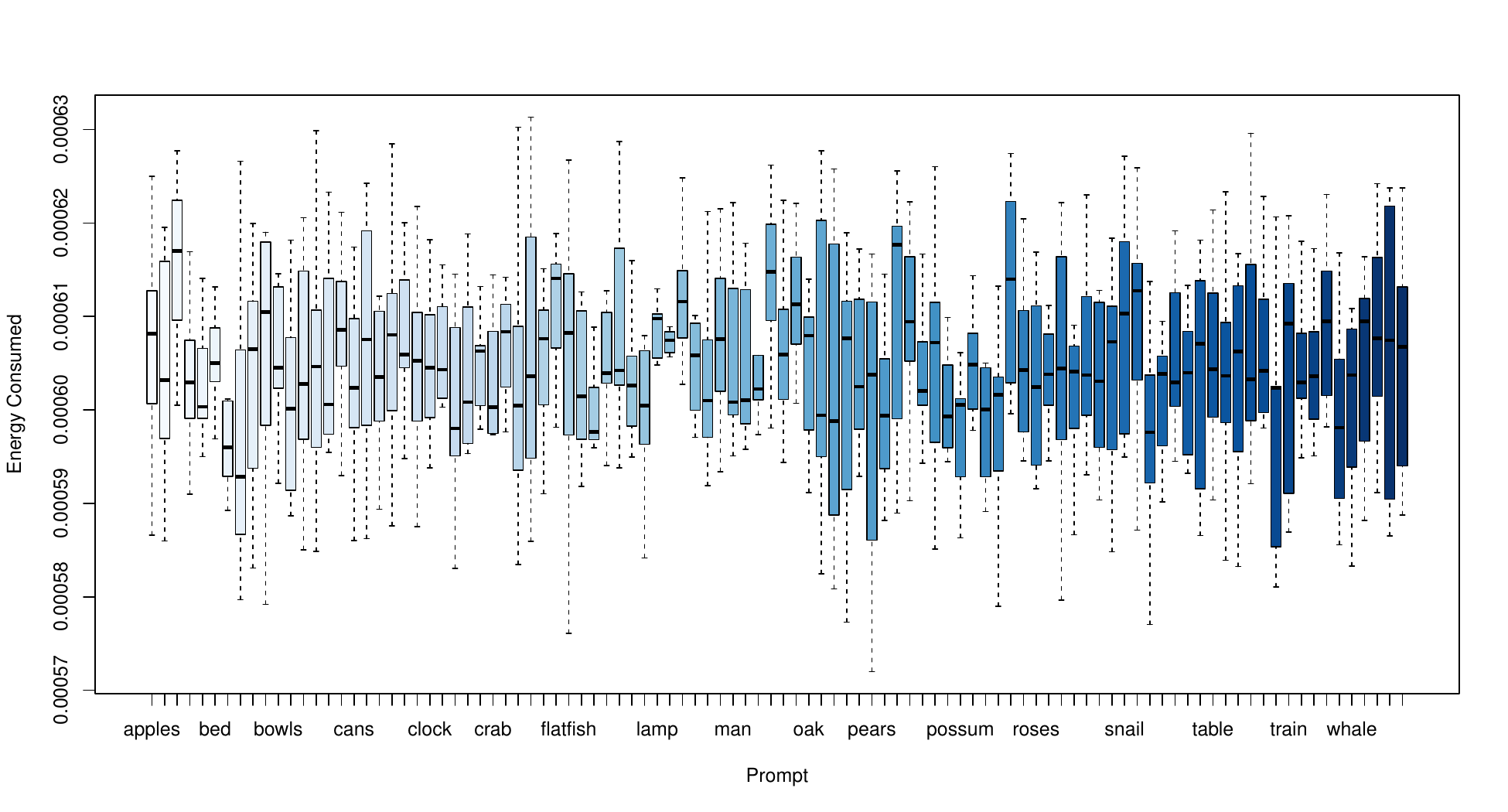}
         \caption{PixArt\_Sigma}
     \end{subfigure}
    \caption{Box plots of the energy consumption across various prompt lengths for the analyzed models. (Group 2).}
    \label{fig:boxplot_content_g2}
\end{figure}

\begin{figure}[p]
    \centering
    \begin{subfigure}[b]{0.45\textwidth}
         \centering
         \includegraphics[width=\textwidth]{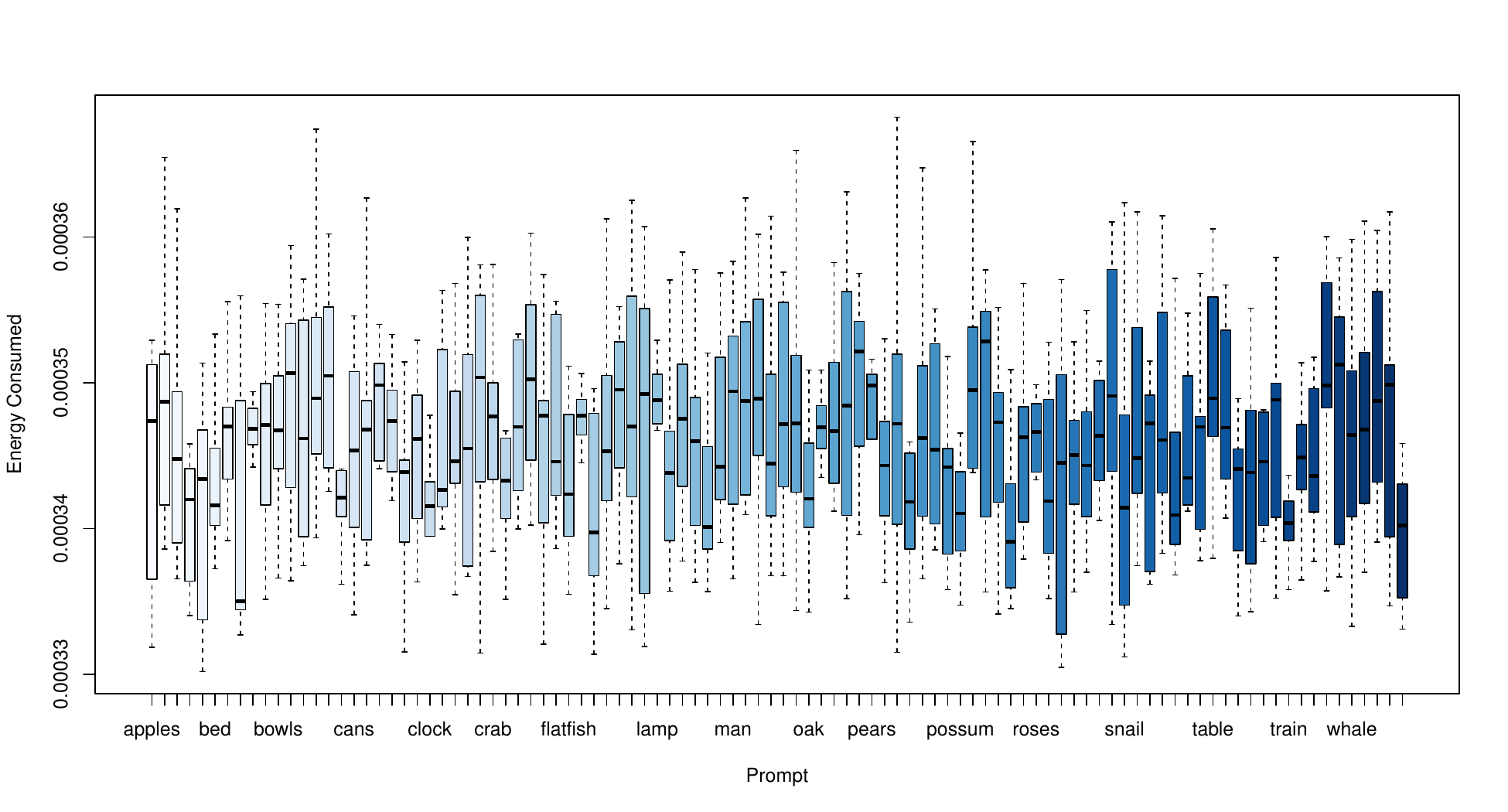}
         \caption{Flash\_PixArt}
     \end{subfigure}
     \hfill
     \begin{subfigure}[b]{0.45\textwidth}
         \centering
         \includegraphics[width=\textwidth]{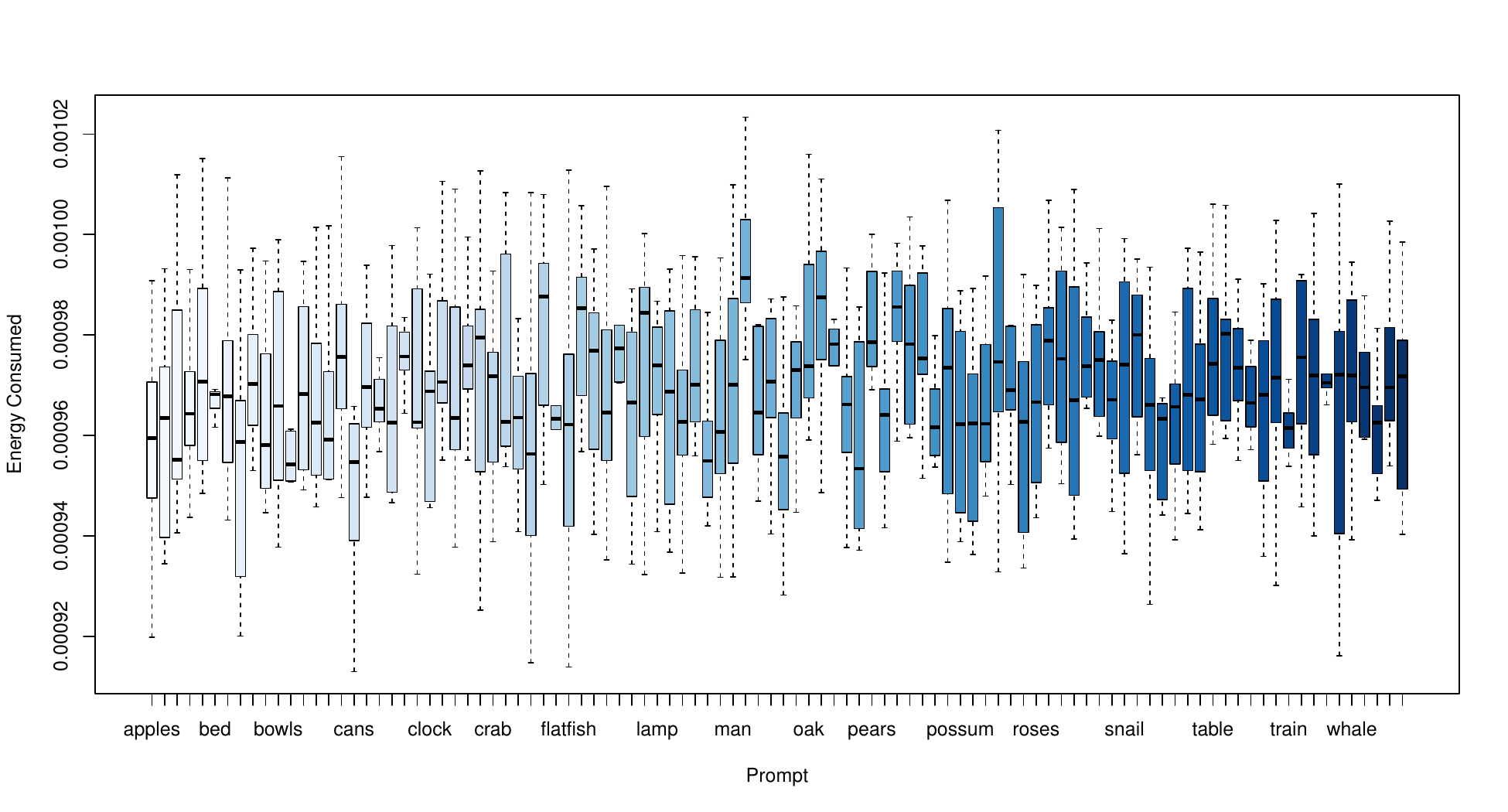}
         \caption{SD\_3}
     \end{subfigure}
     \hfill
     \begin{subfigure}[b]{0.45\textwidth}
         \centering
         \includegraphics[width=\textwidth]{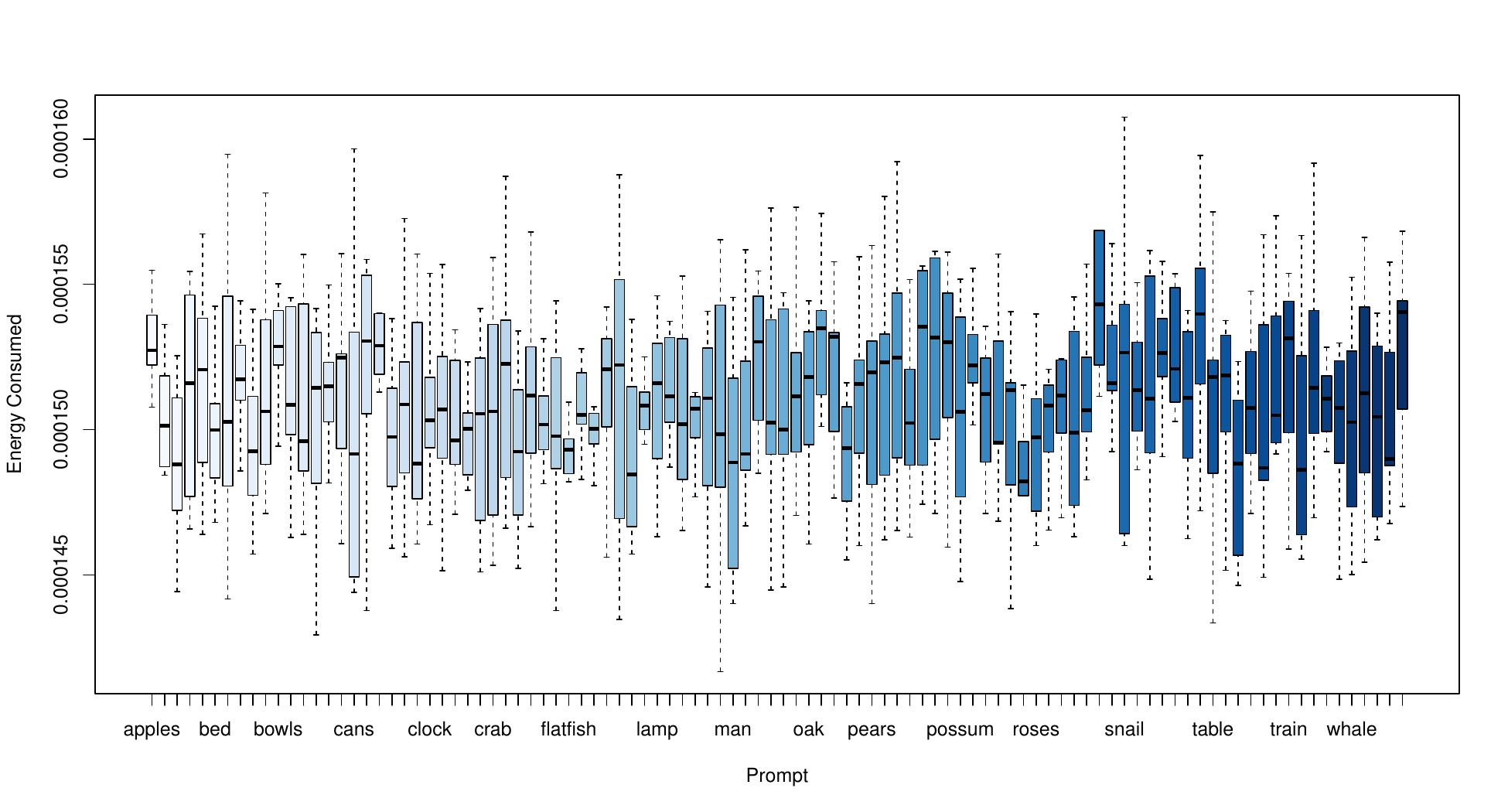}
         \caption{Flash\_SD3}
     \end{subfigure}
     \hfill
     \begin{subfigure}[b]{0.45\textwidth}
         \centering
         \includegraphics[width=\textwidth]{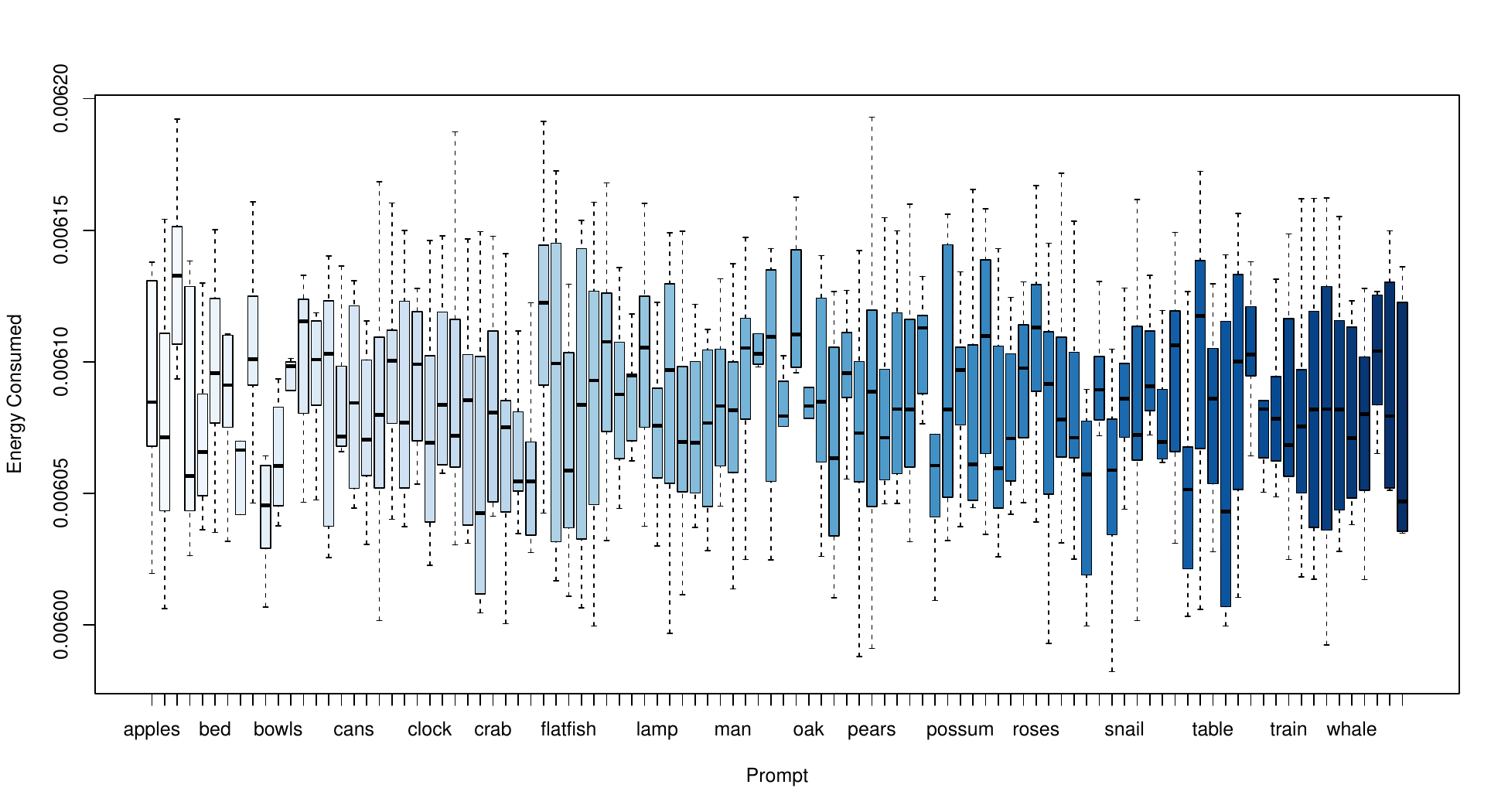}
         \caption{Lumina}
     \end{subfigure}
     \hfill
     \begin{subfigure}[b]{0.45\textwidth}
         \centering
         \includegraphics[width=\textwidth]{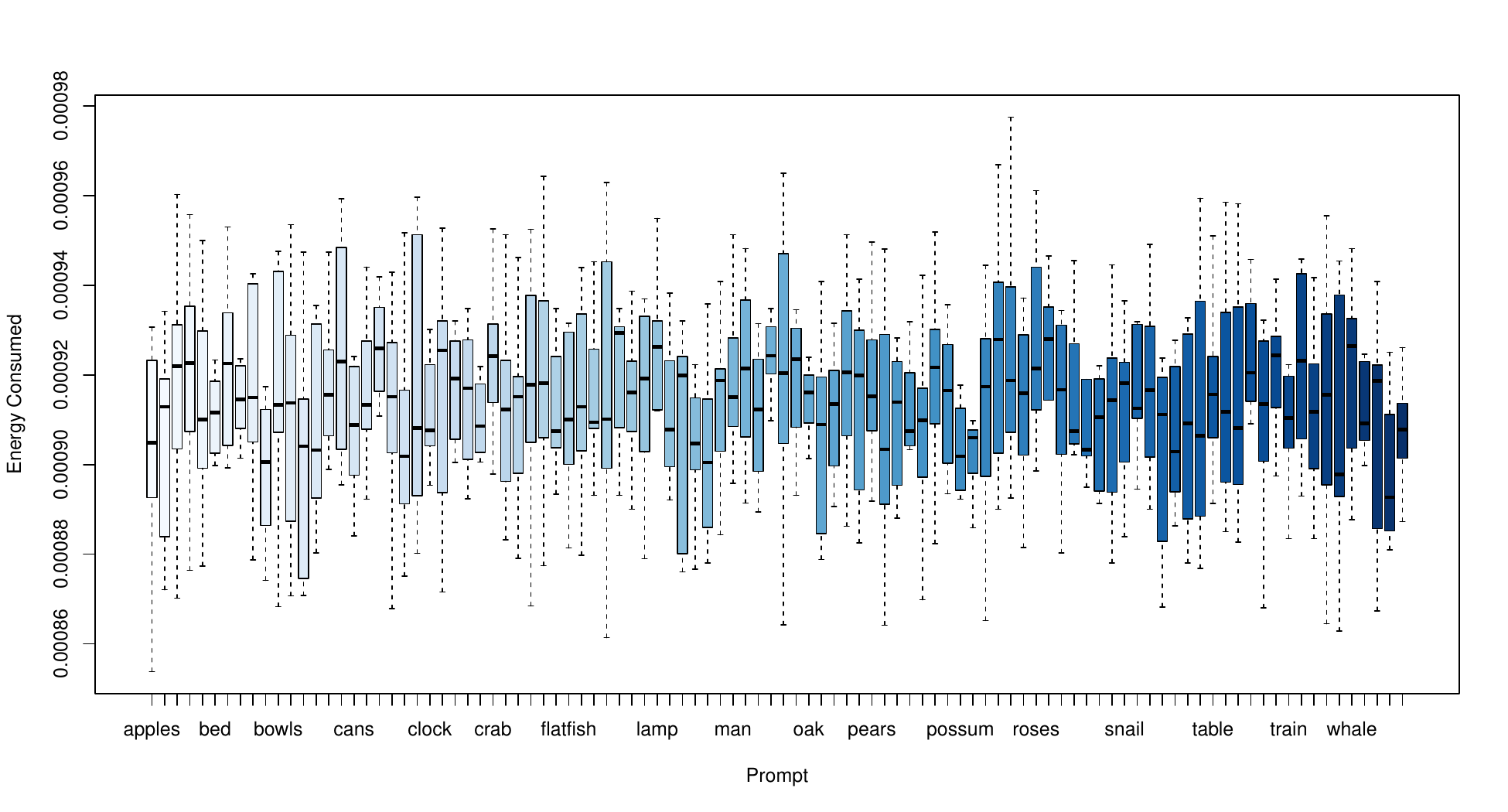}
         \caption{Flux\_1}
     \end{subfigure}
    \caption{Box plots of the energy consumption across various prompt lengths for the analyzed models. (Group 3).}
    \label{fig:boxplot_content_g3}
\end{figure}

From \Cref{tab:prompt_content}, we can observe that the Pearson correlation coefficients (PCC) for all models ranged from -0.042 to 0.036, with all the corresponding p-values exceeding the conventional significance level of 0.05. 
These findings indicate that there is no statistically significant relationship between the prompt content and the energy consumed during the generation process, supporting the validity of the followed experimental setup.

\begin{table}[h]
\caption{Pearson correlation coefficients (PCC) and corresponding p-values assessing the relationship between the semantic content of prompt and energy consumption during image generation process across different models. } \label{tab:prompt_content}
\centering
\adjustbox{width=0.45\textwidth}{
\begin{tabular}{l|c|c}
\toprule
\textbf{Model} & \multicolumn{1}{l|}{\textbf{PCC}} & \textbf{P-value} \\ 
\midrule
Flash\_PixArt & 0.033 & 0.329 \\
Flash\_SD & -0.035 & 0.296 \\
Flash\_SD3 & 0.035 & 0.307 \\
Flash\_SDXL & -0.038 & 0.263 \\
Flux\_1 & -0.024 & 0.485 \\
Hyper\_SD & -0.007 & 0.841 \\
LCM\_SDXL & -0.029 & 0.392 \\
LCM\_SSD\_1B & -0.009 & 0.795 \\
Lumina & 0.024 & 0.481 \\
PixArt\_Alpha & 0.021 & 0.532 \\
PixArt\_Sigma & 0.036 & 0.294 \\
SD\_1.5 & -0.042 & 0.213 \\
SD\_3 & -0.030 & 0.378 \\
SDXL & 0.027 & 0.426 \\
SDXL\_Lightning & -0.030 & 0.371 \\
SDXL\_Turbo & -0.011 & 0.751 \\
SSD\_1B & -0.003 & 0.934 \\ 
\bottomrule
\end{tabular}
}
\end{table}

Additionally, \Cref{fig:boxplot_content_g1,fig:boxplot_content_g2,fig:boxplot_content_g3} show the box plots which illustrate the distribution of energy consumption across different models for the 100 different prompts. 
As we can see, the median energy remains quite stable across all semantics, with negligible variations among prompts. 
This observation supports the finding from Pearson correlation test that prompt semantic does not impact on the energy consumption.

\section{Image Resolution}\label{app:image_resolution}

\begin{figure}[h]
     \centering
     \begin{subfigure}[b]{0.48\textwidth}
         \centering
         \includegraphics[width=\textwidth]{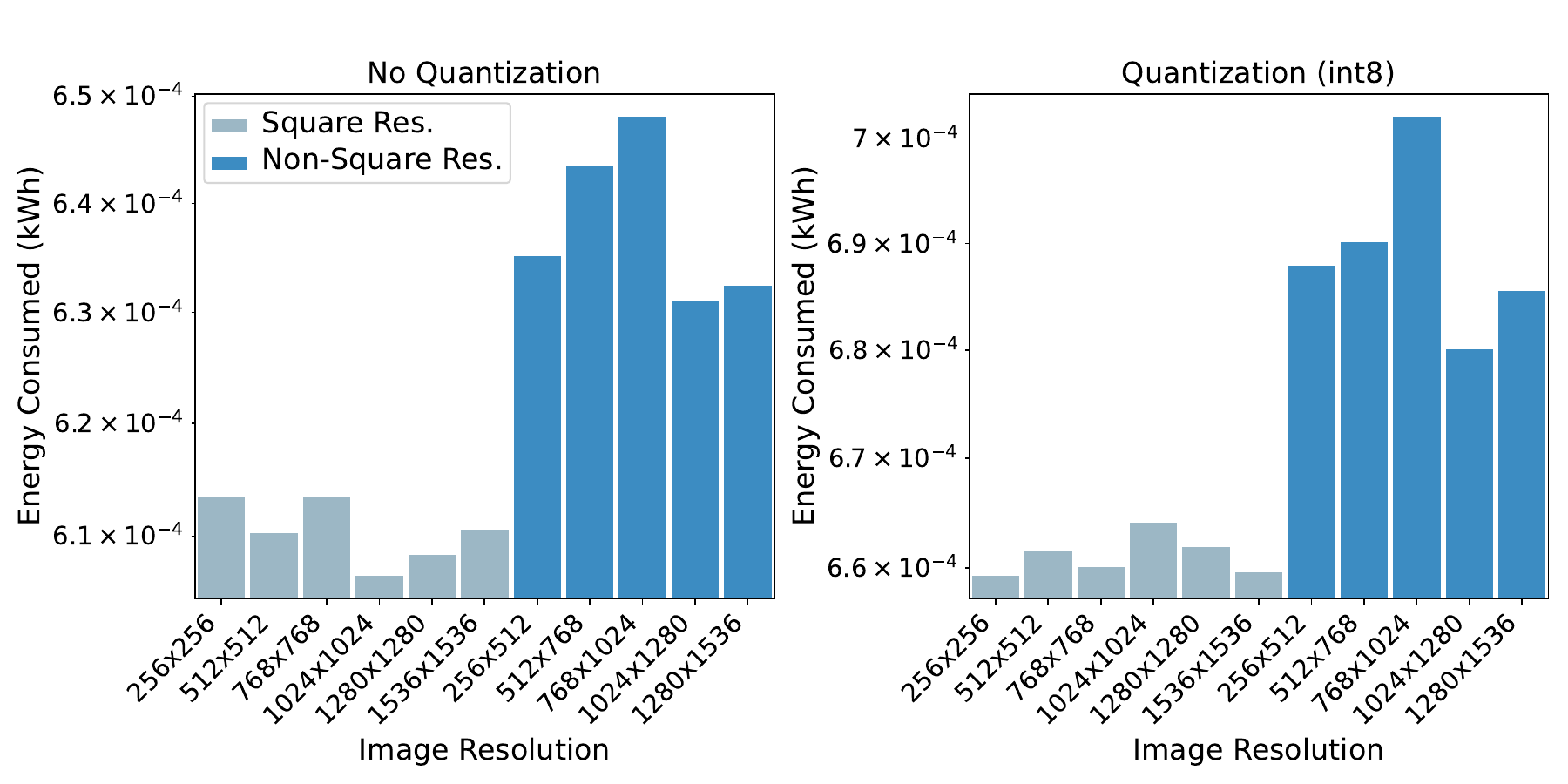}
         \caption{PixArt Alpha}
     \end{subfigure}
     \hfill
     \begin{subfigure}[b]{0.48\textwidth}
         \centering
         \includegraphics[width=\textwidth]{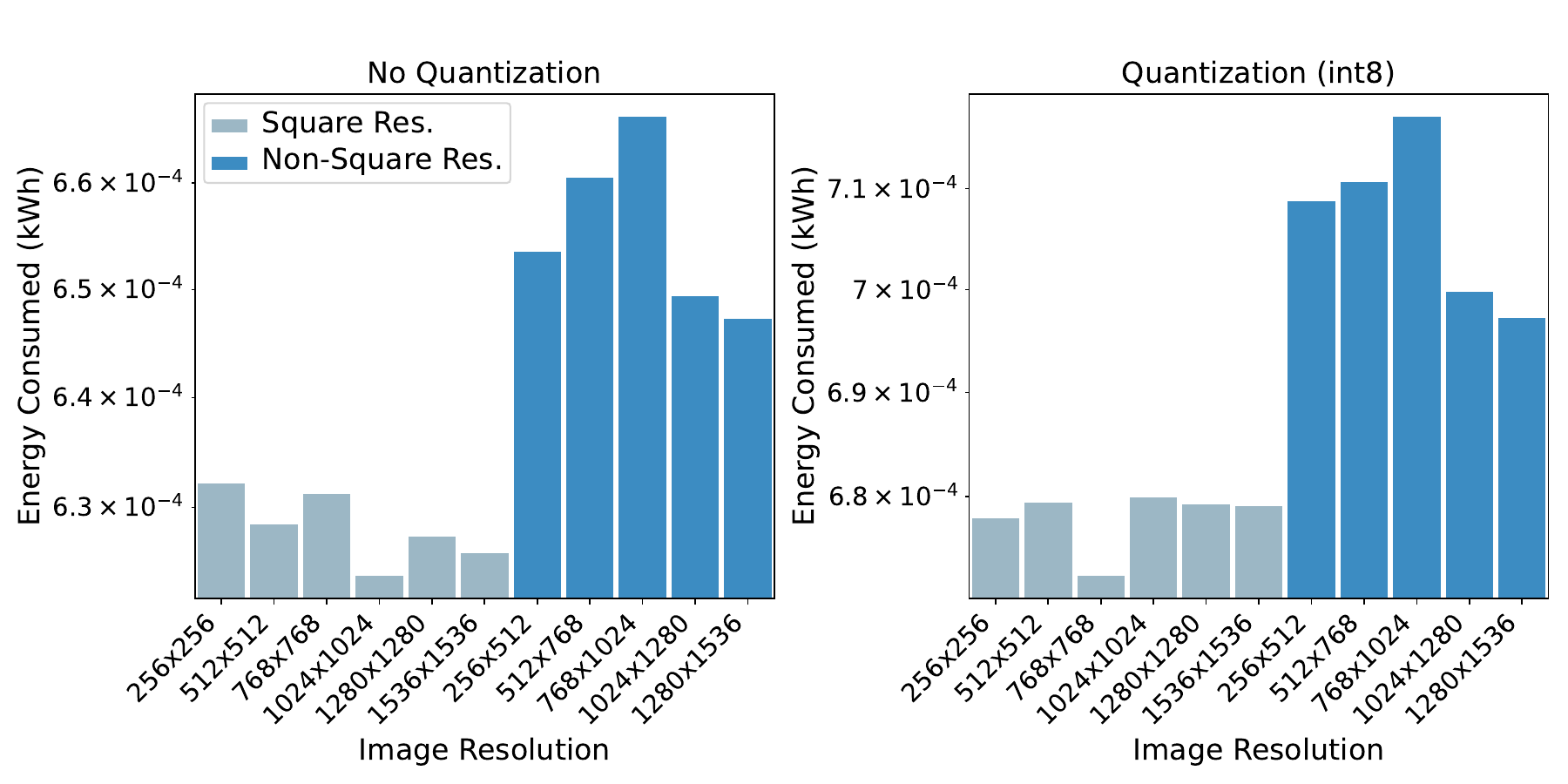}
         \caption{PixArt Sigma}
     \end{subfigure}
     \hfill
     \begin{subfigure}[b]{0.48\textwidth}
         \centering
         \includegraphics[width=\textwidth]{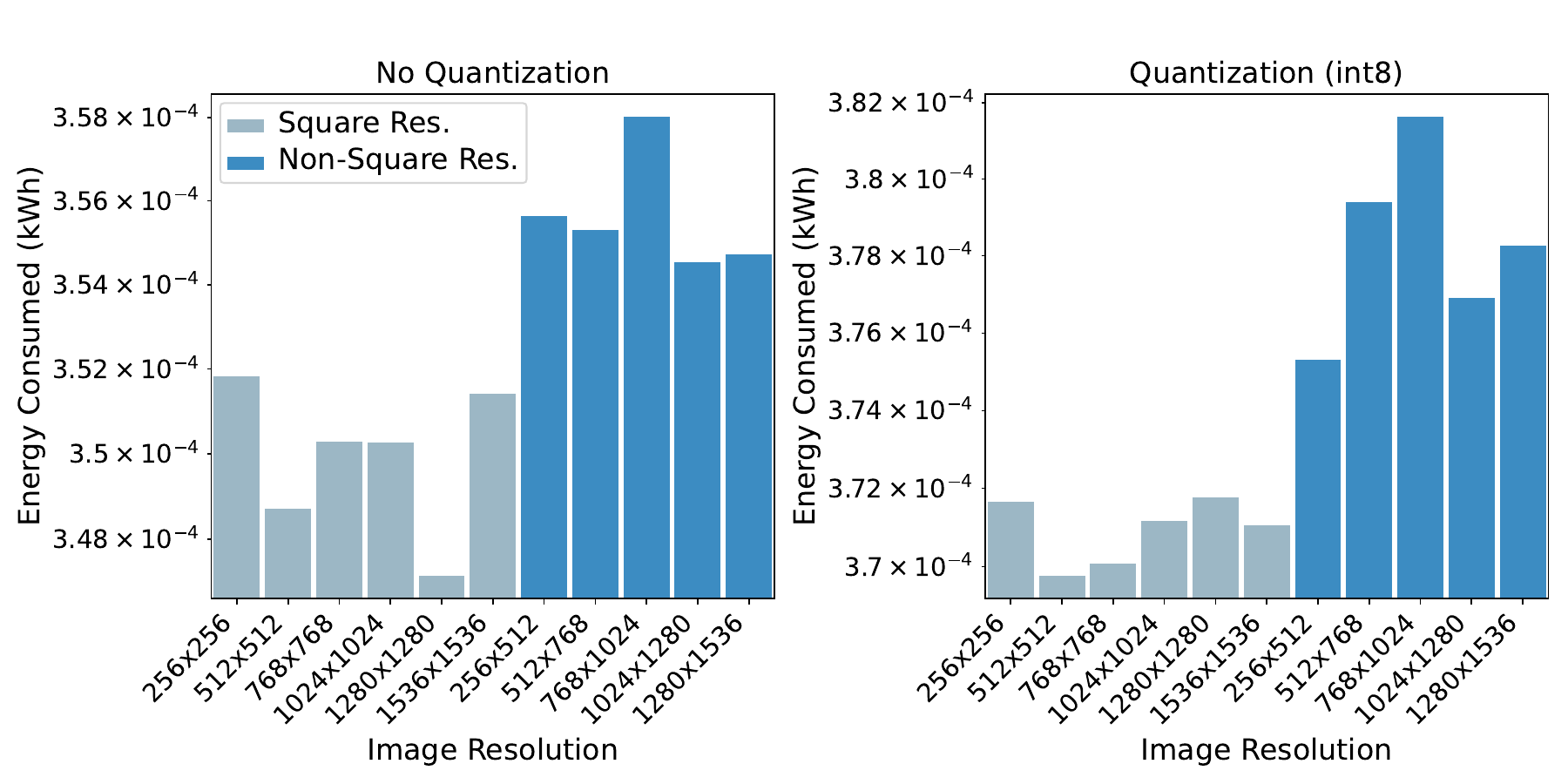}
         \caption{Flash PixArt}
     \end{subfigure}
     \caption{Energy consumption (kWh) of PixArt\_Alpha, PixArt\_Sigma, and Flash\_PixArt models at varying image resolution. Bar charts (log scale) show the energy consumed across different image resolutions, categorized as square and non-square resolutions, at fixed quantization setting. Each bar shows the median over 30 prompts.}
     \label{fig:pixart_resolutions}
     \end{figure}

\Cref{fig:pixart_resolutions} illustrates the additional experiments conducted on PixArt\_Alpha, PixArt\_Sigma, and Flash\_PixArt models to further investigate the impact of image resolution on energy consumption. 
We can notice that for all models the energy consumption for non-square resolution is much higher compared to square resolutions.

\section{Prompt Length}\label{app:prompt_length}

\begin{figure}[h]
     \centering
     \begin{subfigure}[b]{0.23\textwidth}
         \centering
         \includegraphics[width=\textwidth]{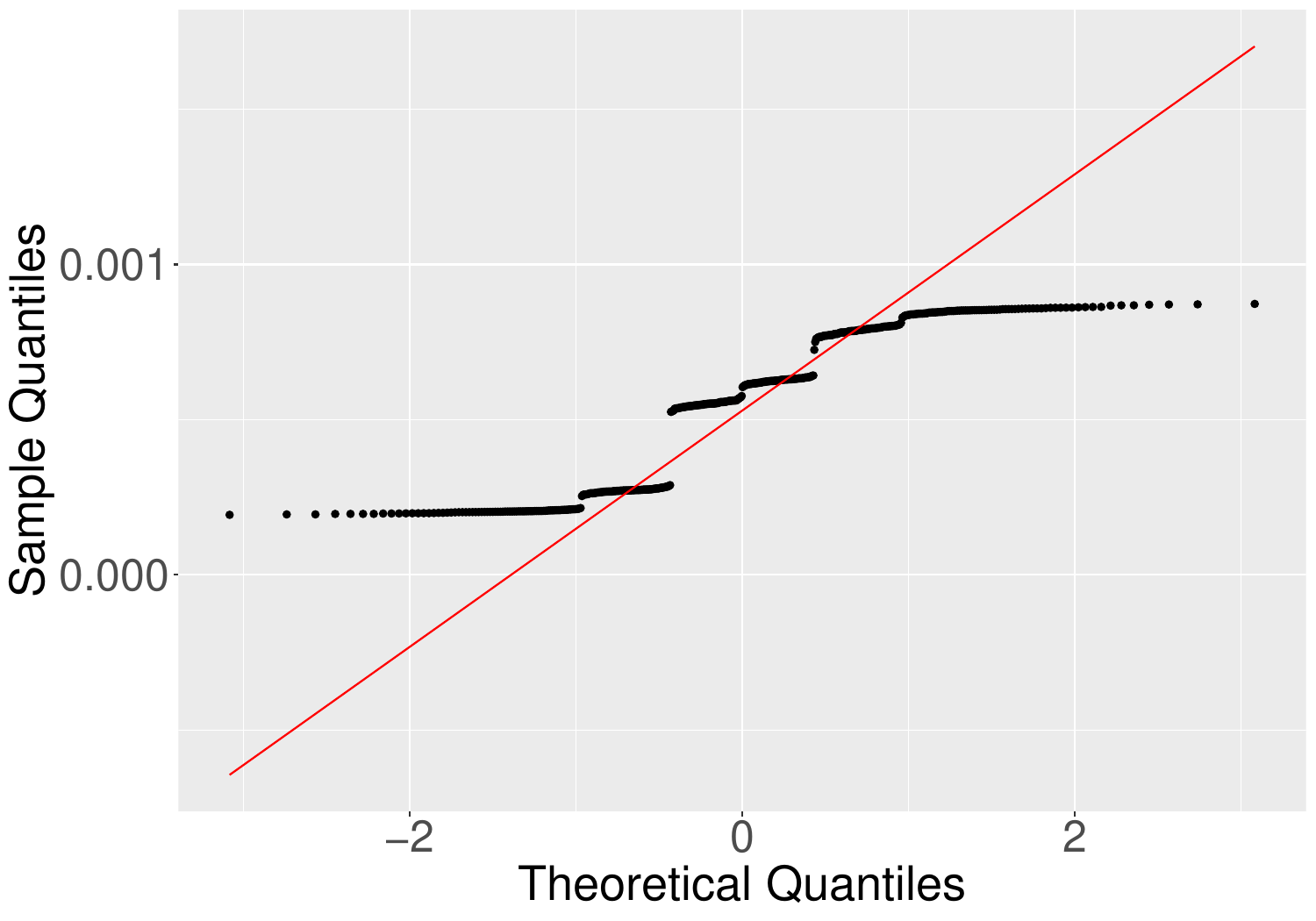}
         \caption{SD\_1.5}
     \end{subfigure}
     \hfill
     \begin{subfigure}[b]{0.23\textwidth}
         \centering
         \includegraphics[width=\textwidth]{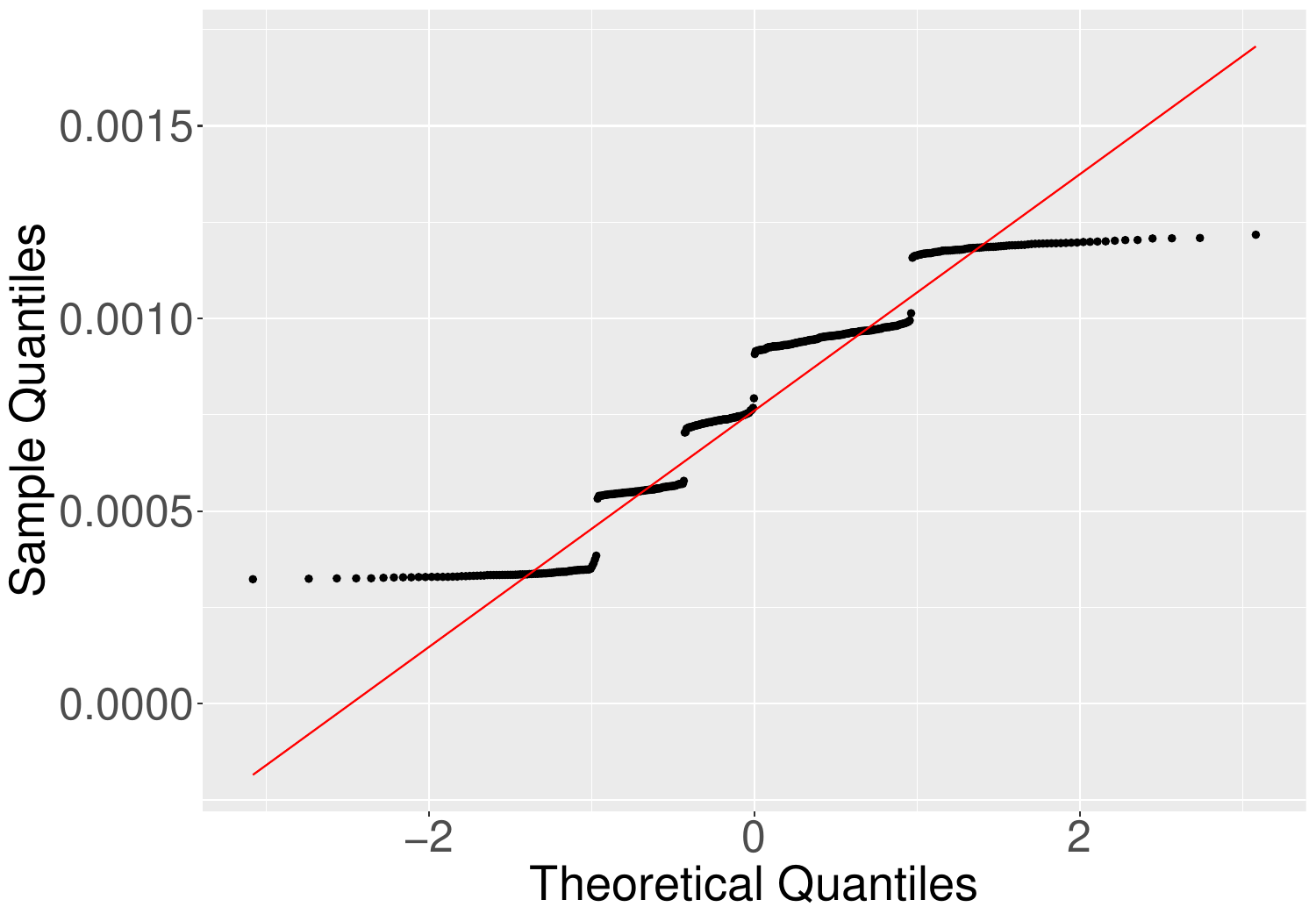}
         \caption{SDXL}
     \end{subfigure}
     \hfill
     \begin{subfigure}[b]{0.23\textwidth}
         \centering
         \includegraphics[width=\textwidth]{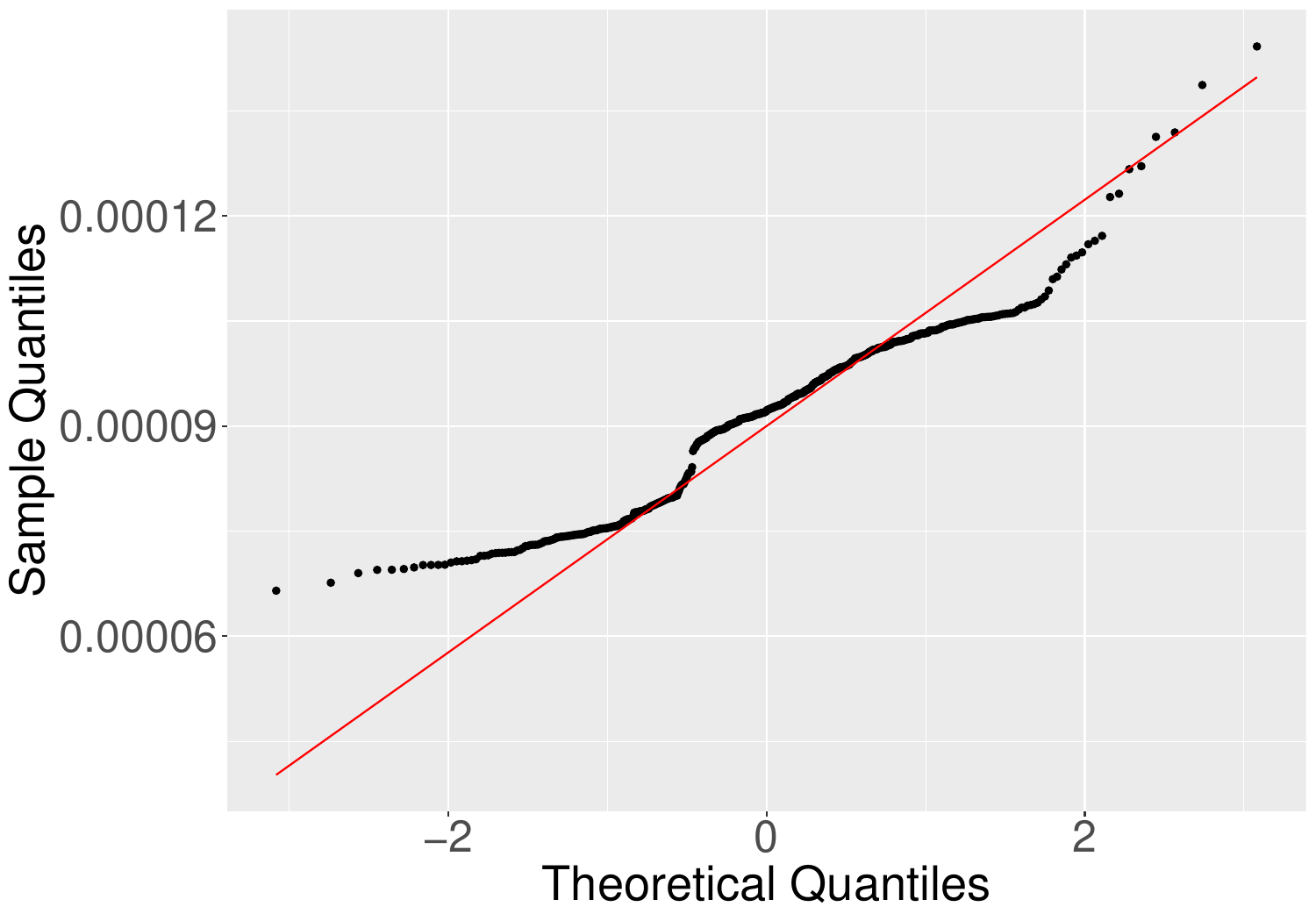}
         \caption{SDXL\_Turbo}
     \end{subfigure}
     \hfill
     \begin{subfigure}[b]{0.23\textwidth}
         \centering
         \includegraphics[width=\textwidth]{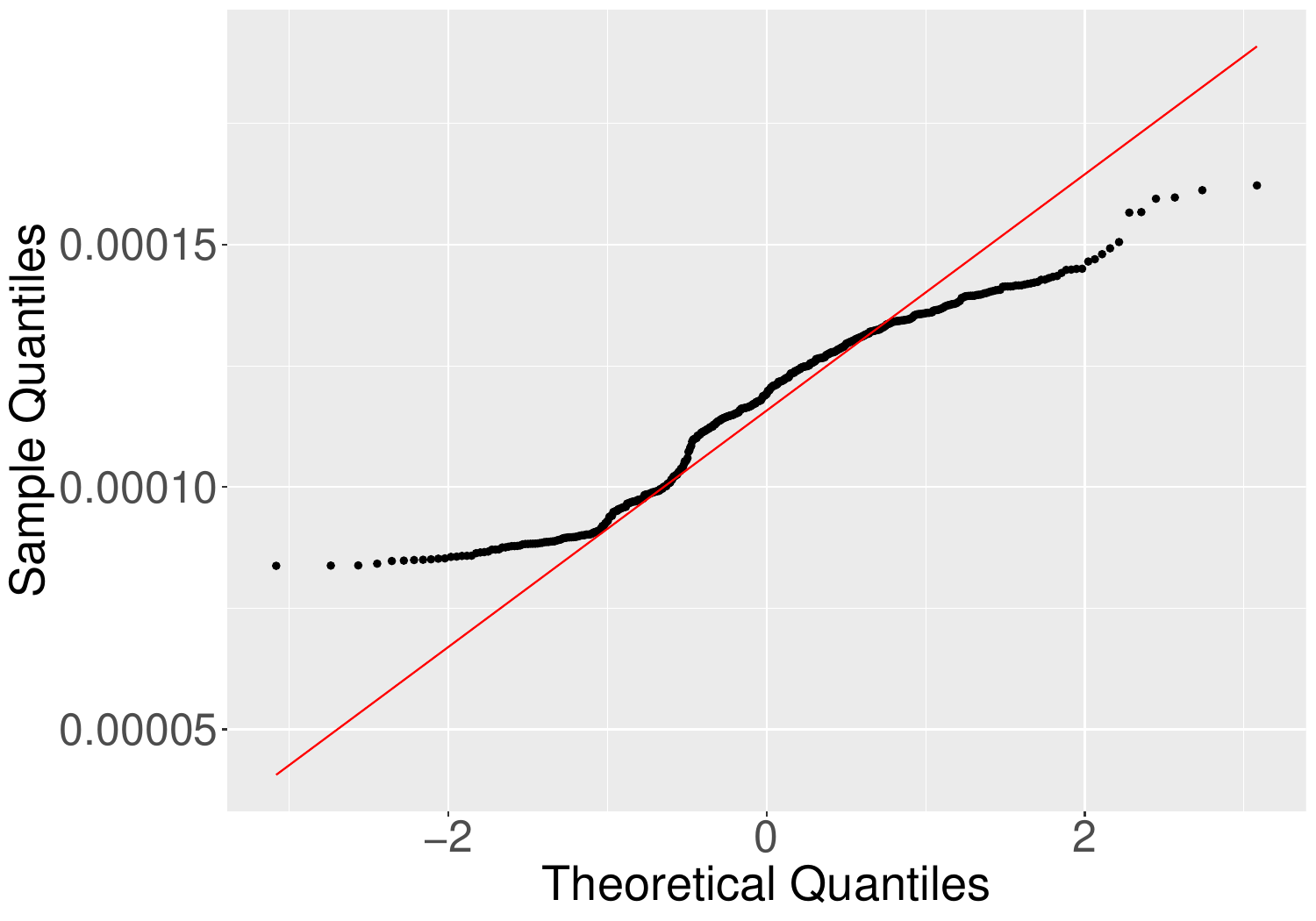}
         \caption{SDXL\_Lightning}
     \end{subfigure}
     \begin{subfigure}[b]{0.23\textwidth}
         \centering
         \includegraphics[width=\textwidth]{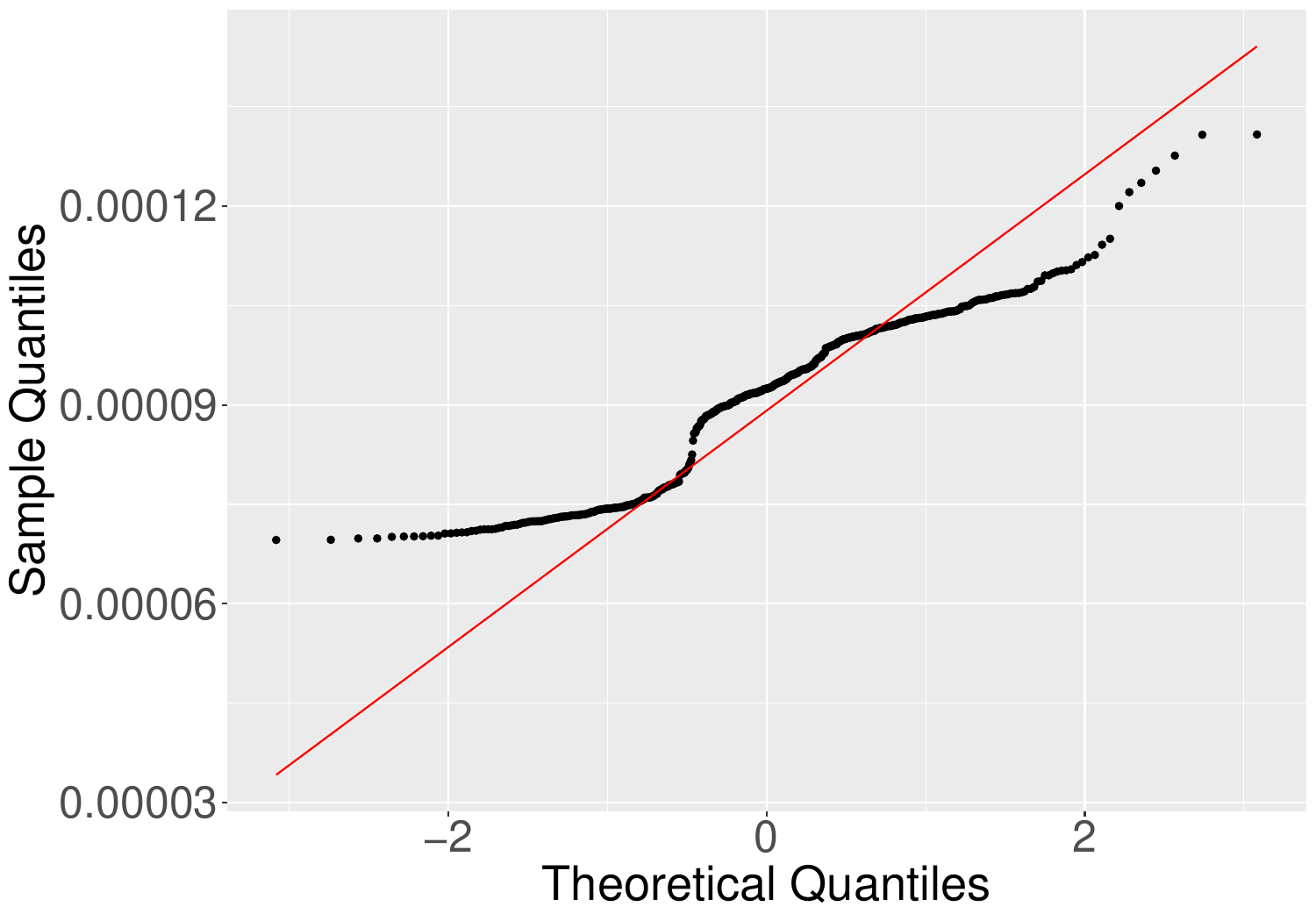}
         \caption{Hyper\_SD}
     \end{subfigure}
     \hfill
     \begin{subfigure}[b]{0.23\textwidth}
         \centering
         \includegraphics[width=\textwidth]{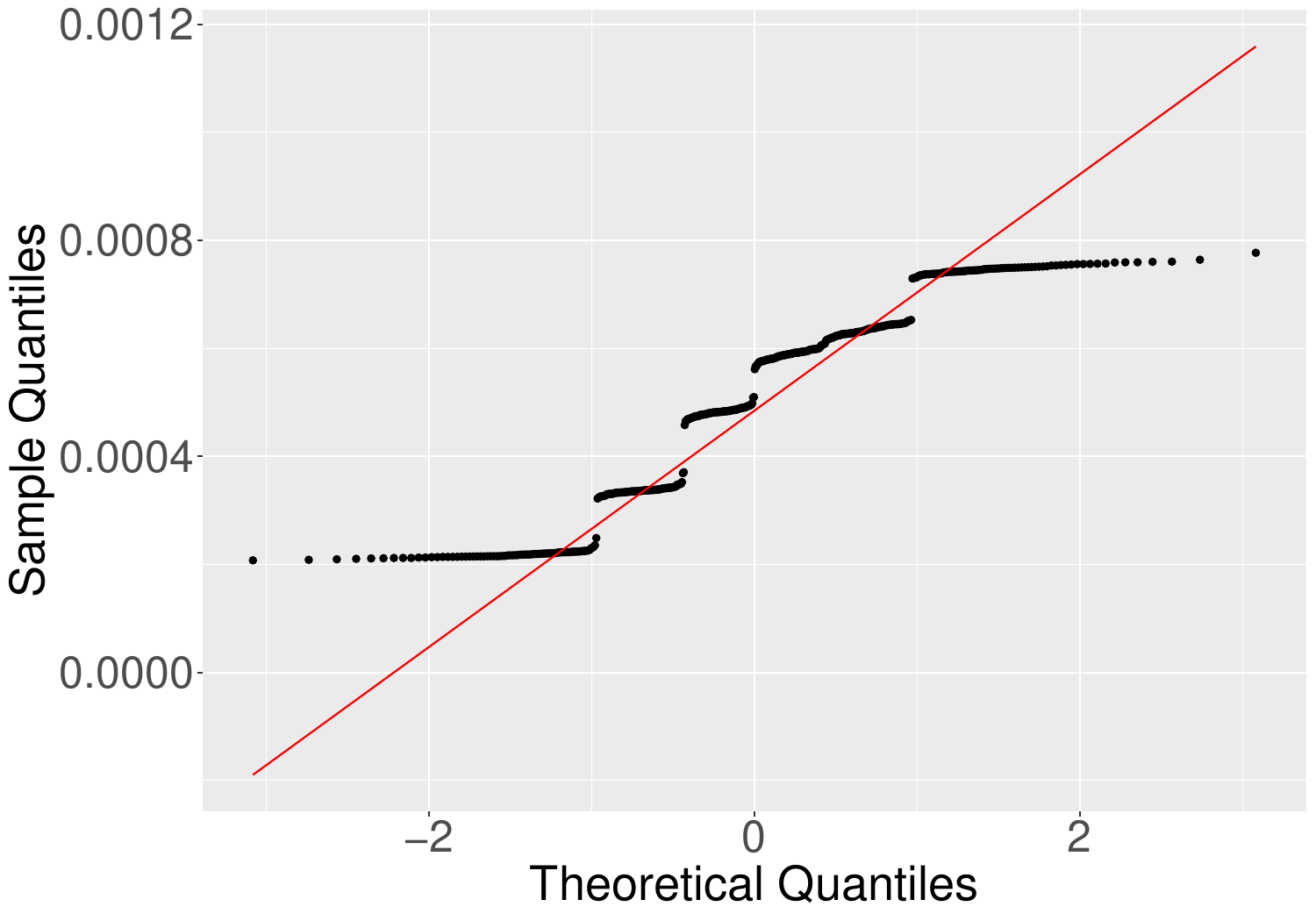}
         \caption{SSD\_1B}
     \end{subfigure}
     \hfill
     \begin{subfigure}[b]{0.23\textwidth}
         \centering
         \includegraphics[width=\textwidth]{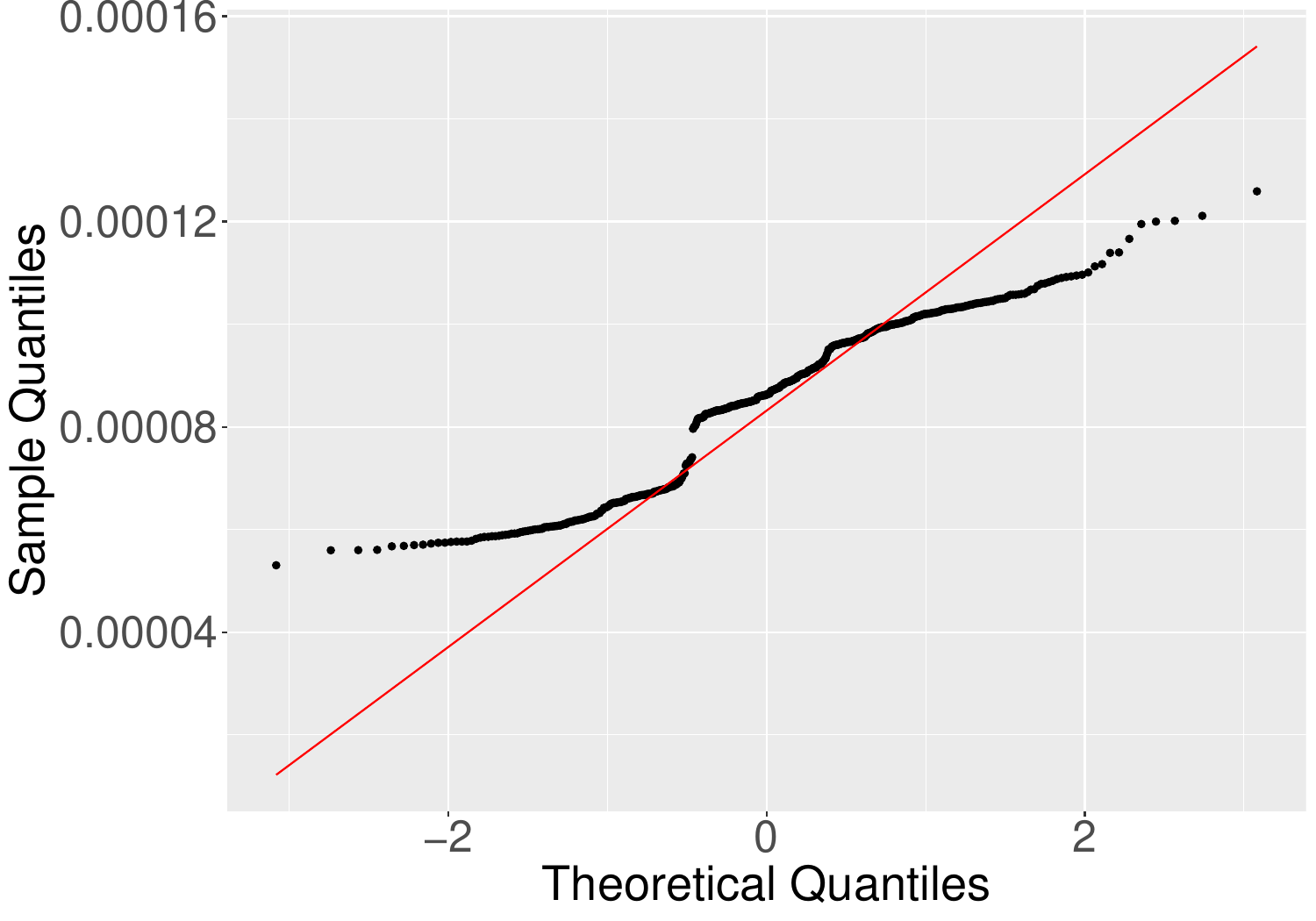}
         \caption{LCM\_SSD\_1B}
     \end{subfigure}
     \hfill
     \begin{subfigure}[b]{0.23\textwidth}
         \centering
         \includegraphics[width=\textwidth]{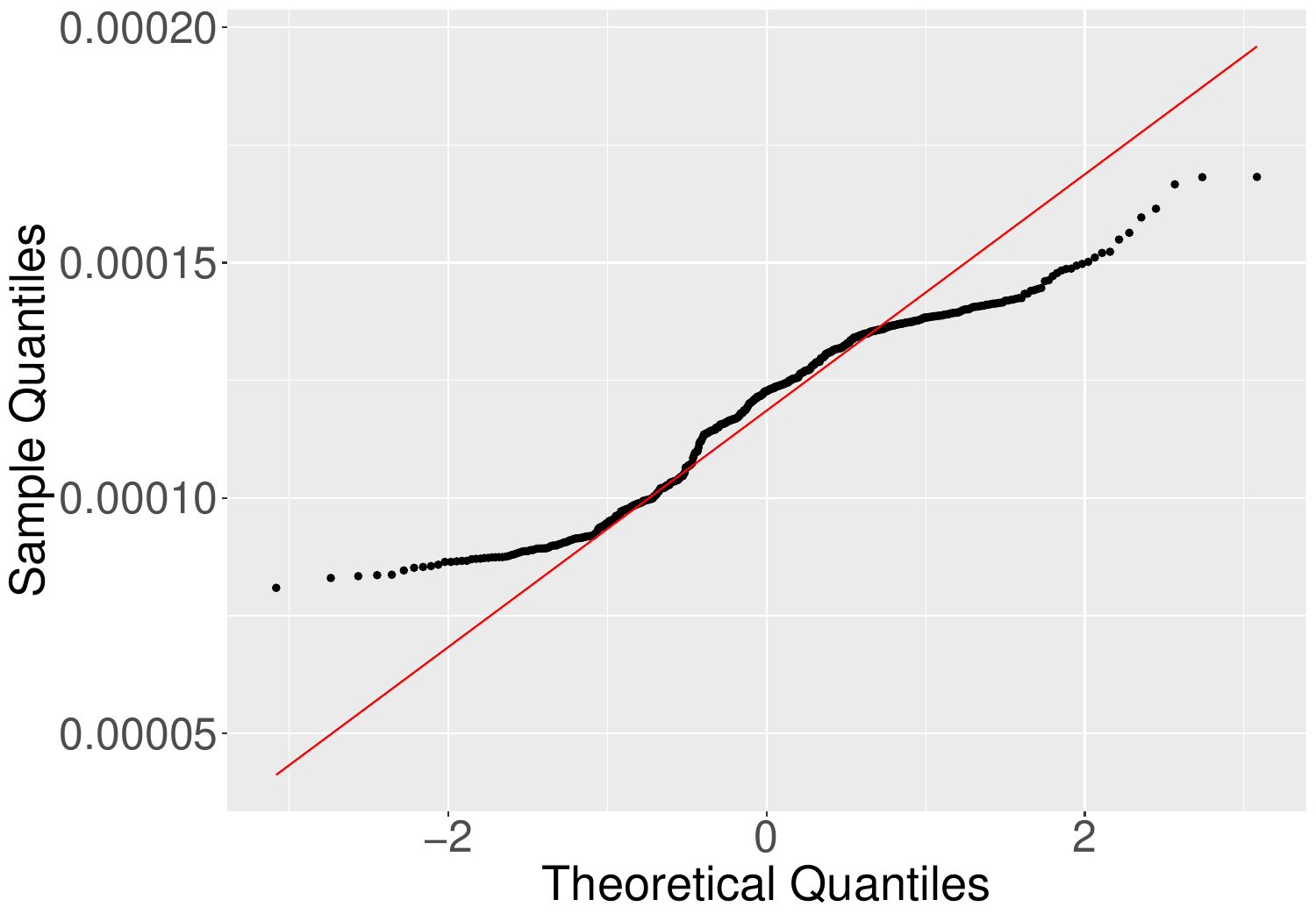}
         \caption{LCM\_SDXL}
     \end{subfigure}
     \begin{subfigure}[b]{0.23\textwidth}
         \centering
         \includegraphics[width=\textwidth]{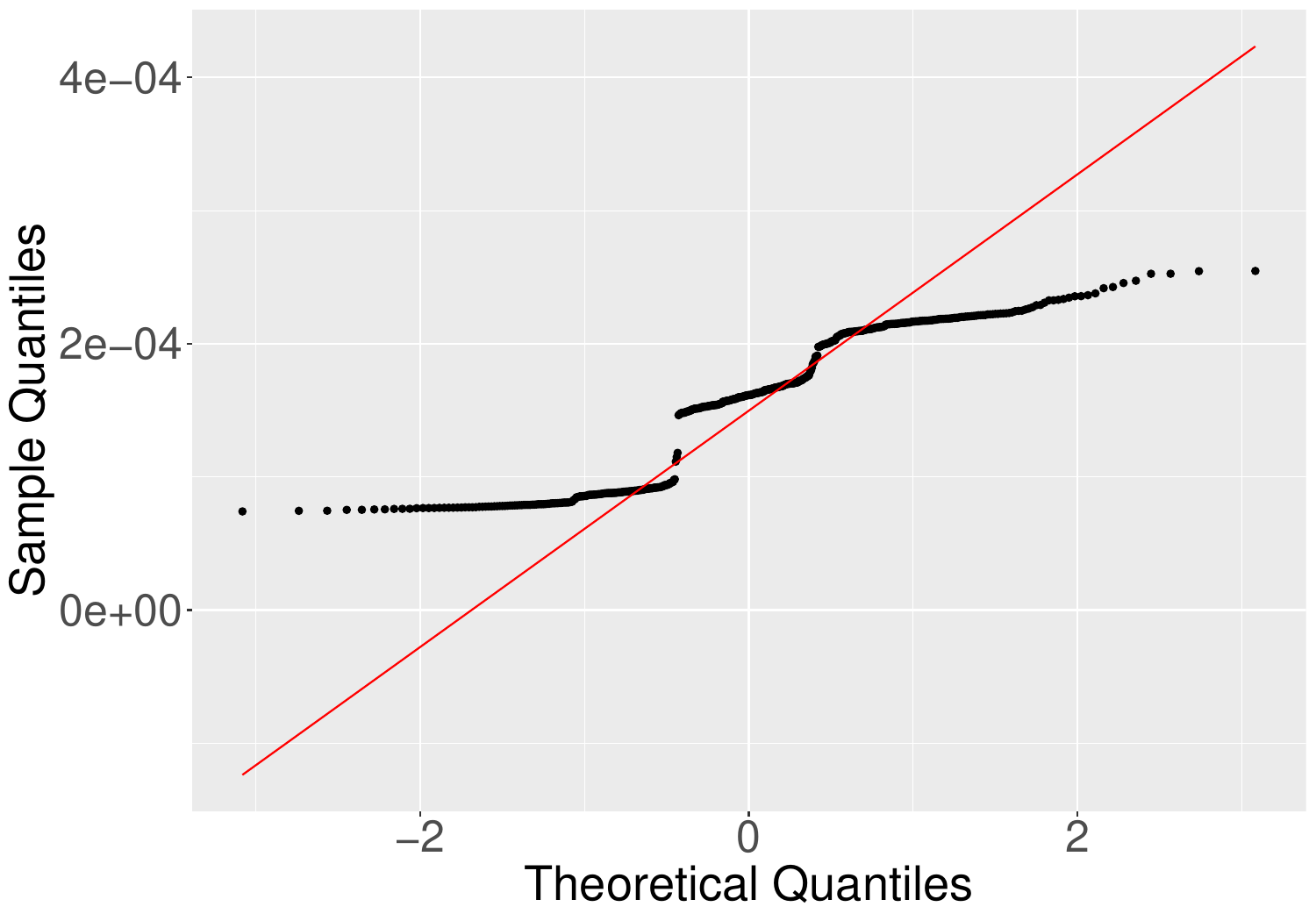}
         \caption{Flash\_SD}
     \end{subfigure}
     \hfill
     \begin{subfigure}[b]{0.23\textwidth}
         \centering
         \includegraphics[width=\textwidth]{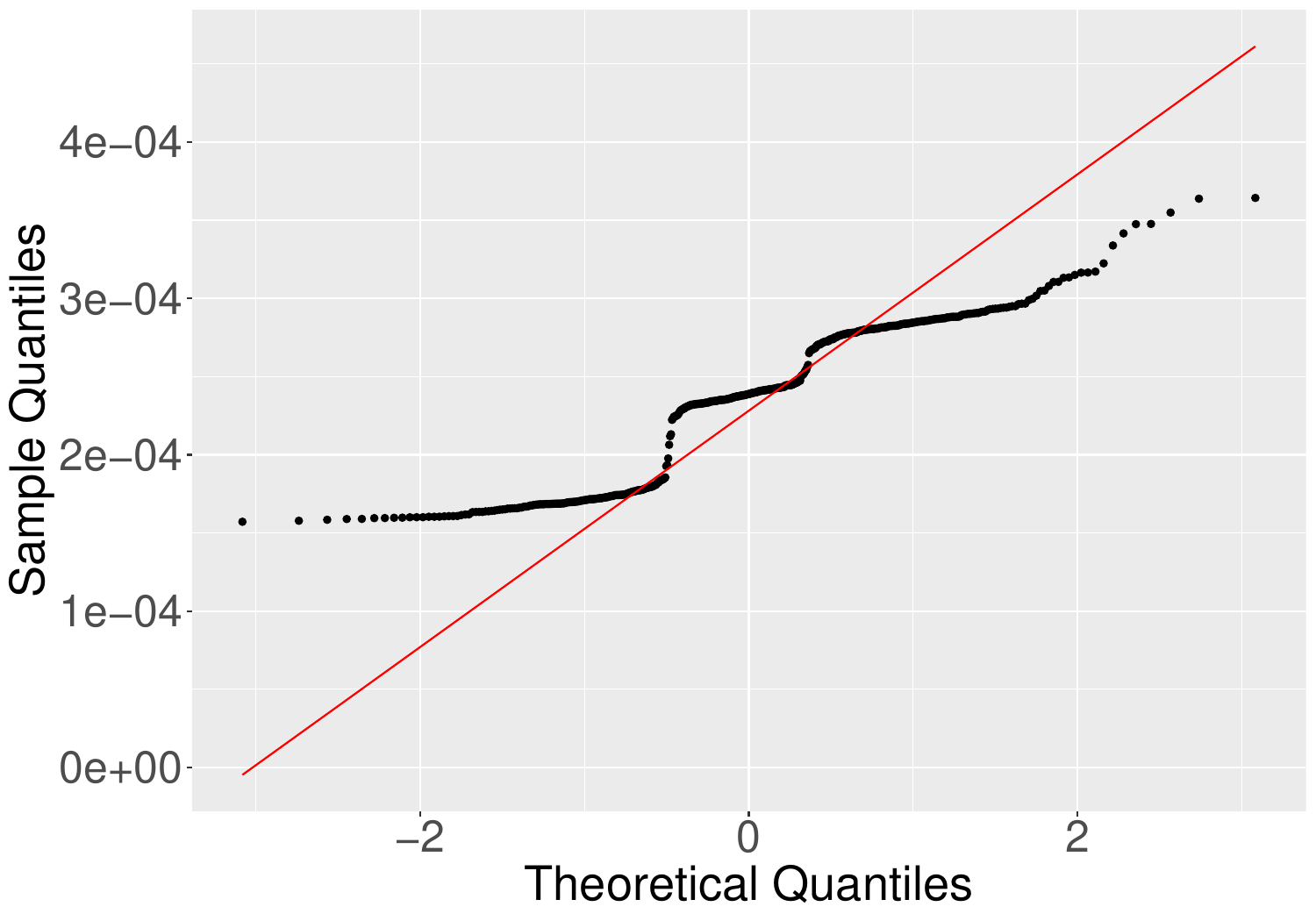}
         \caption{Flash\_SDXL}
     \end{subfigure}
     \hfill
     \begin{subfigure}[b]{0.23\textwidth}
         \centering
         \includegraphics[width=\textwidth]{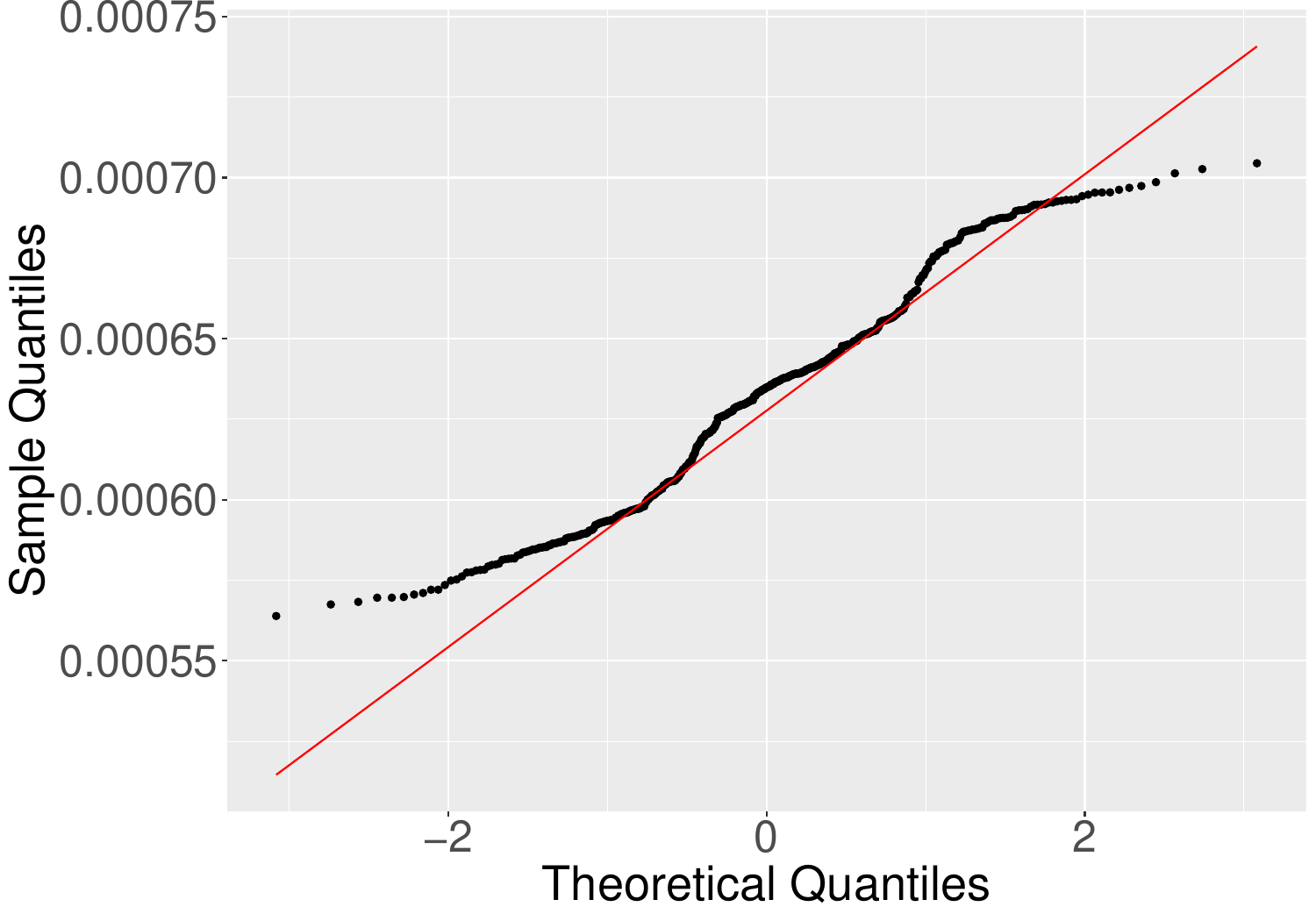}
         \caption{PixArt\_Alpha}
     \end{subfigure}
     \hfill
     \begin{subfigure}[b]{0.23\textwidth}
         \centering
         \includegraphics[width=\textwidth]{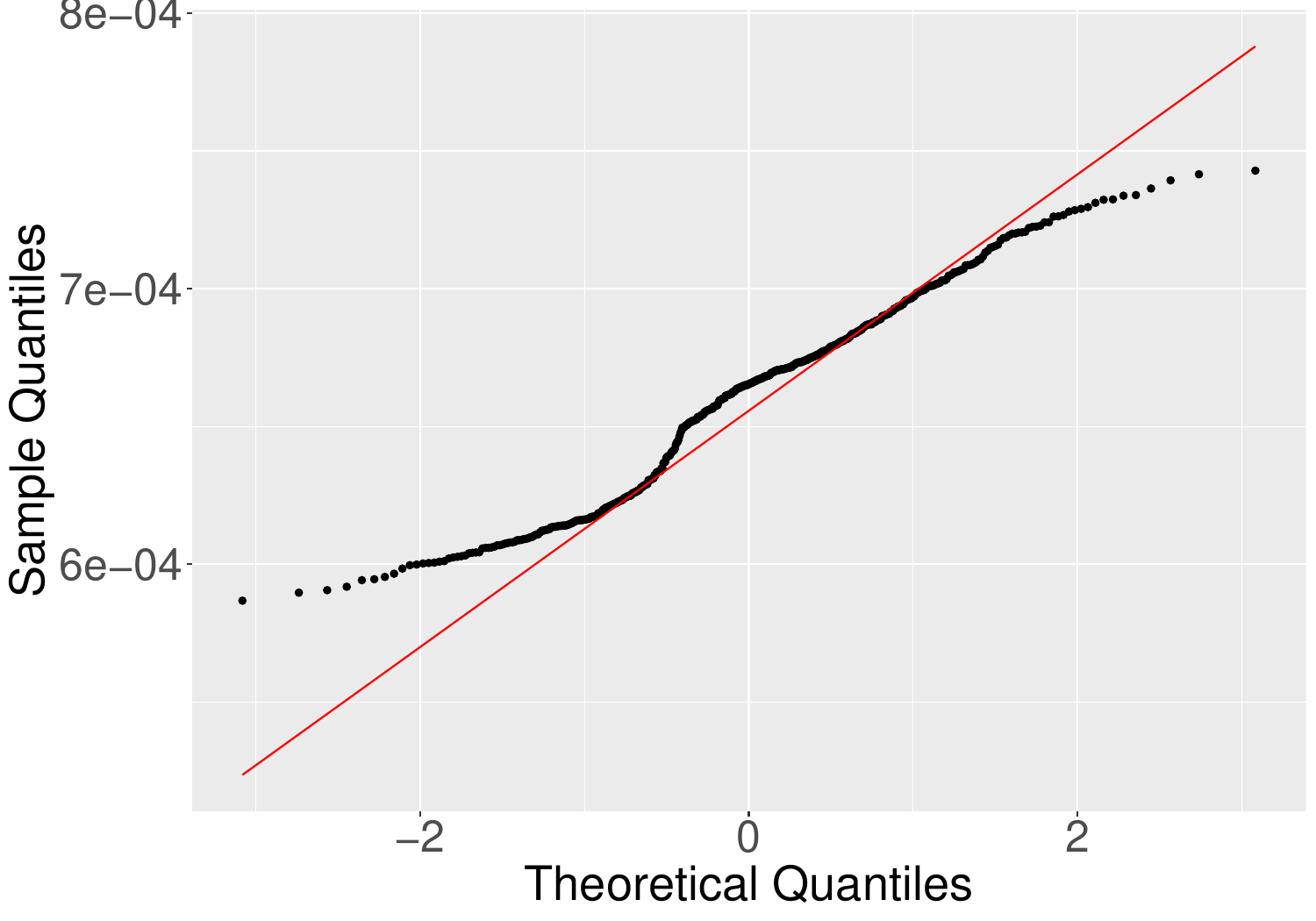}
         \caption{PixArt\_Sigma}
     \end{subfigure}
     \hfill
     \begin{subfigure}[b]{0.23\textwidth}
         \centering
         \includegraphics[width=\textwidth]{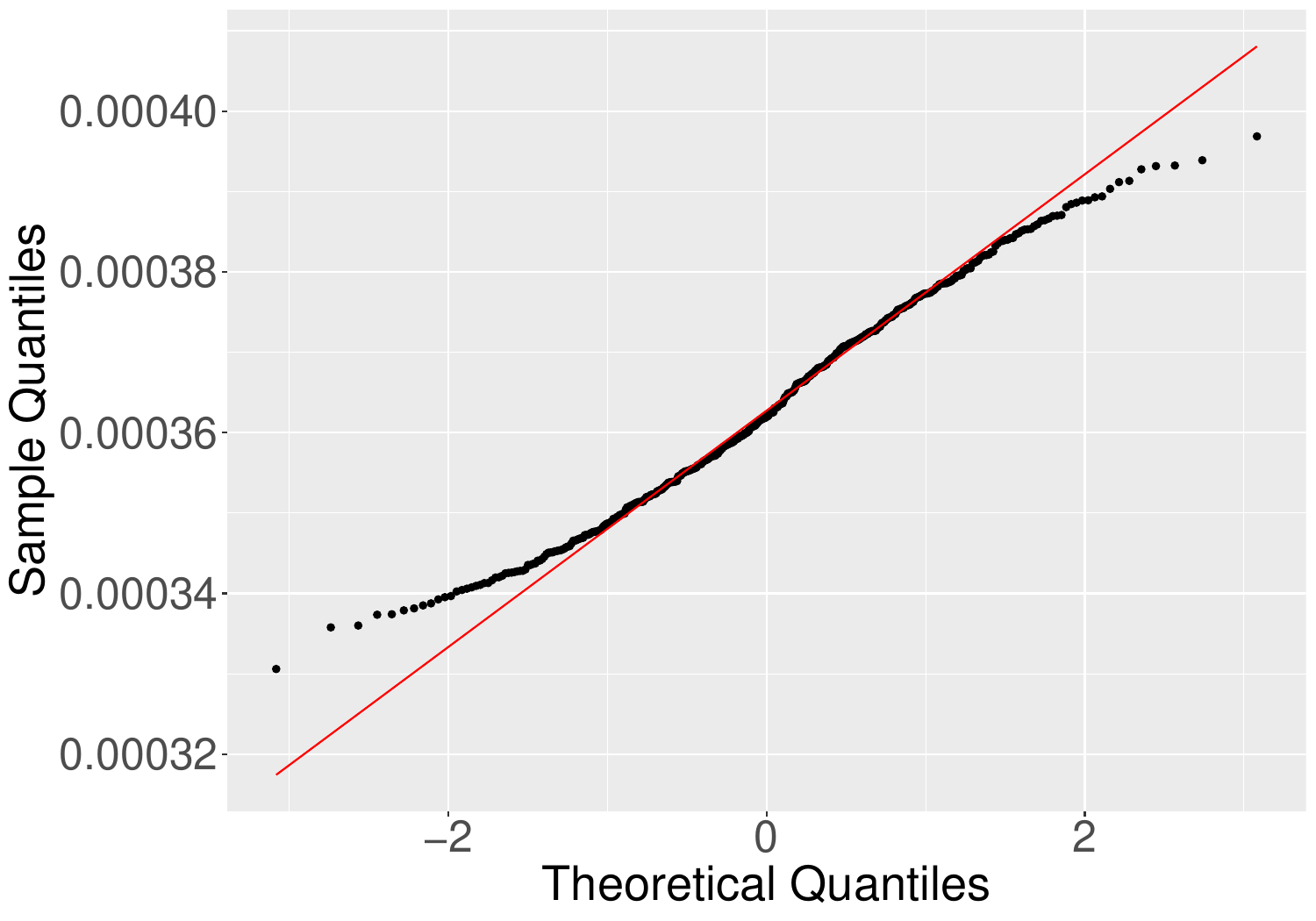}
         \caption{Flash\_PixArt}
     \end{subfigure}
     \hfill
     \begin{subfigure}[b]{0.23\textwidth}
         \centering
         \includegraphics[width=\textwidth]{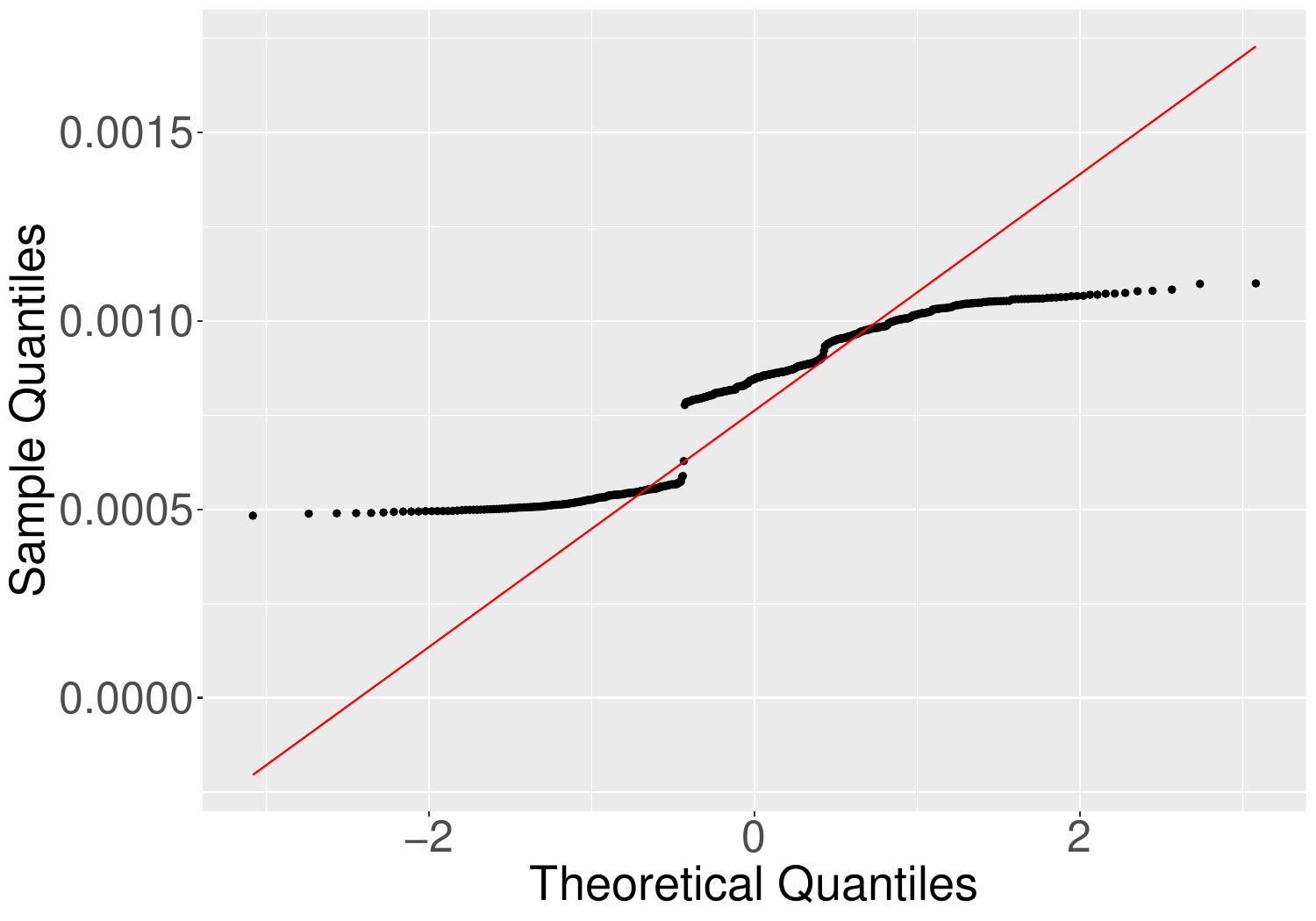}
         \caption{SD\_3}
     \end{subfigure}
     \hfill
     \begin{subfigure}[b]{0.23\textwidth}
         \centering
         \includegraphics[width=\textwidth]{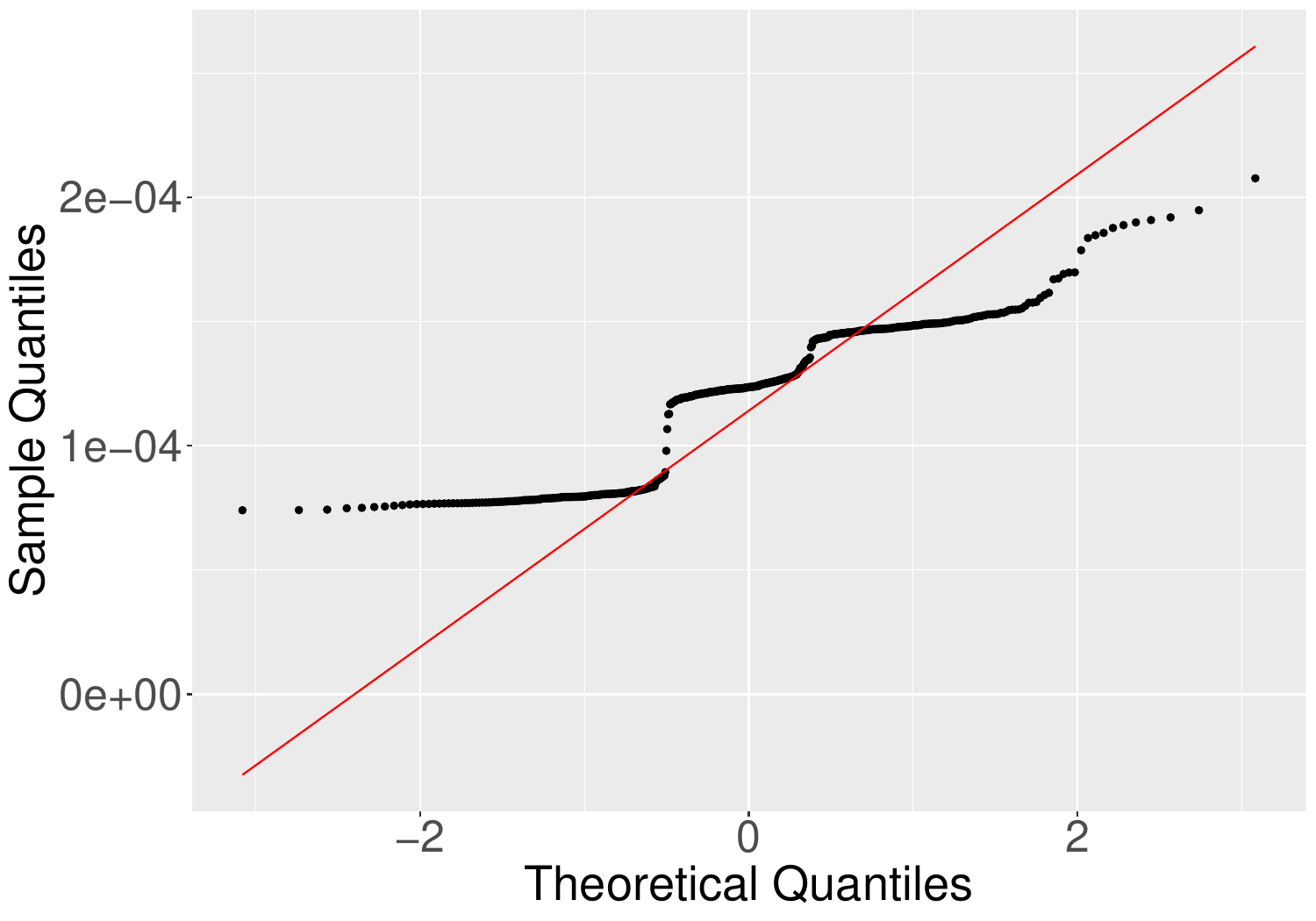}
         \caption{Flash\_SD3}
     \end{subfigure}
     \hfill
     \begin{subfigure}[b]{0.23\textwidth}
         \centering
         \includegraphics[width=\textwidth]{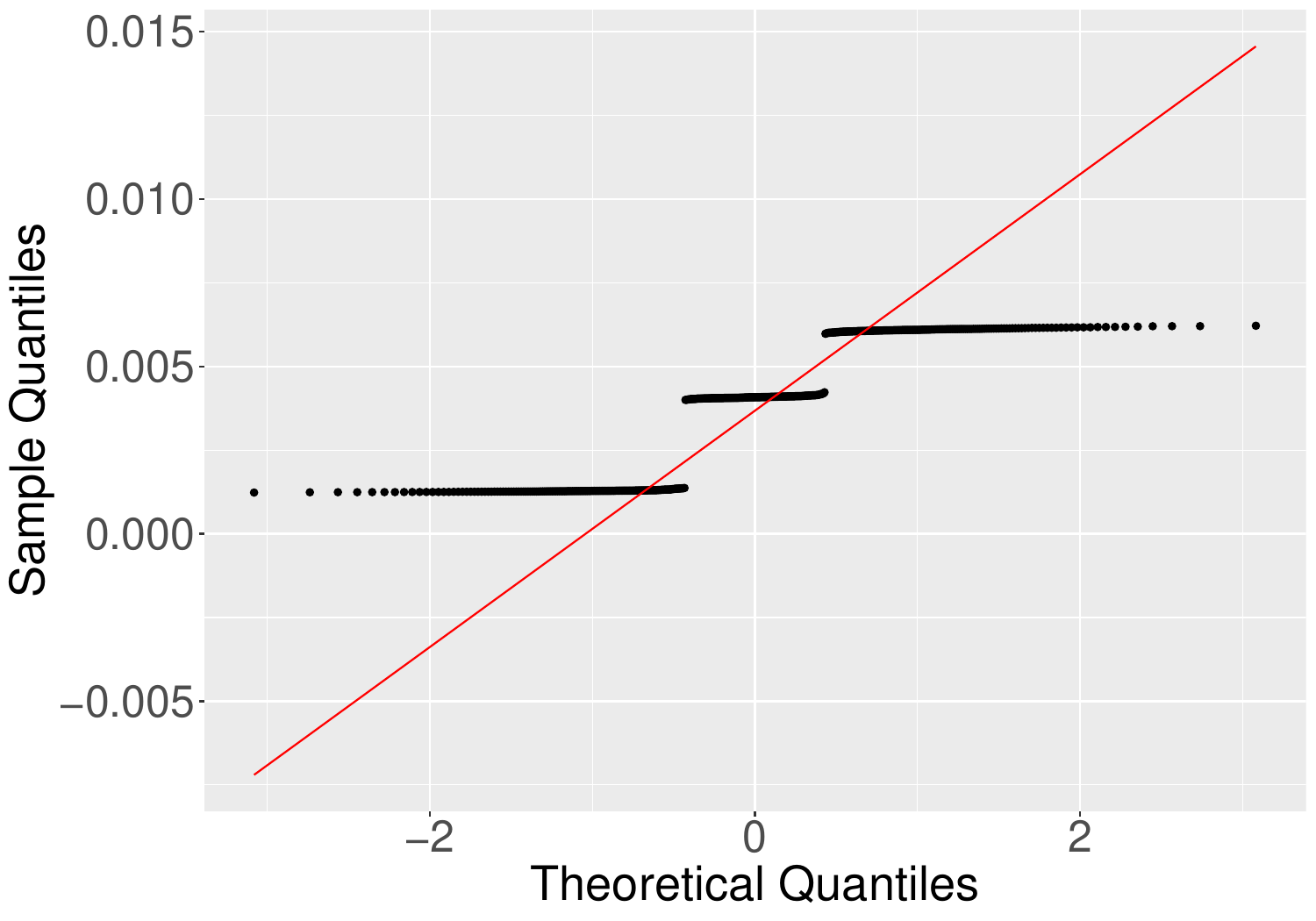}
         \caption{Lumina}
     \end{subfigure}
     \hfill
     \begin{subfigure}[b]{0.23\textwidth}
         \centering
         \includegraphics[width=\textwidth]{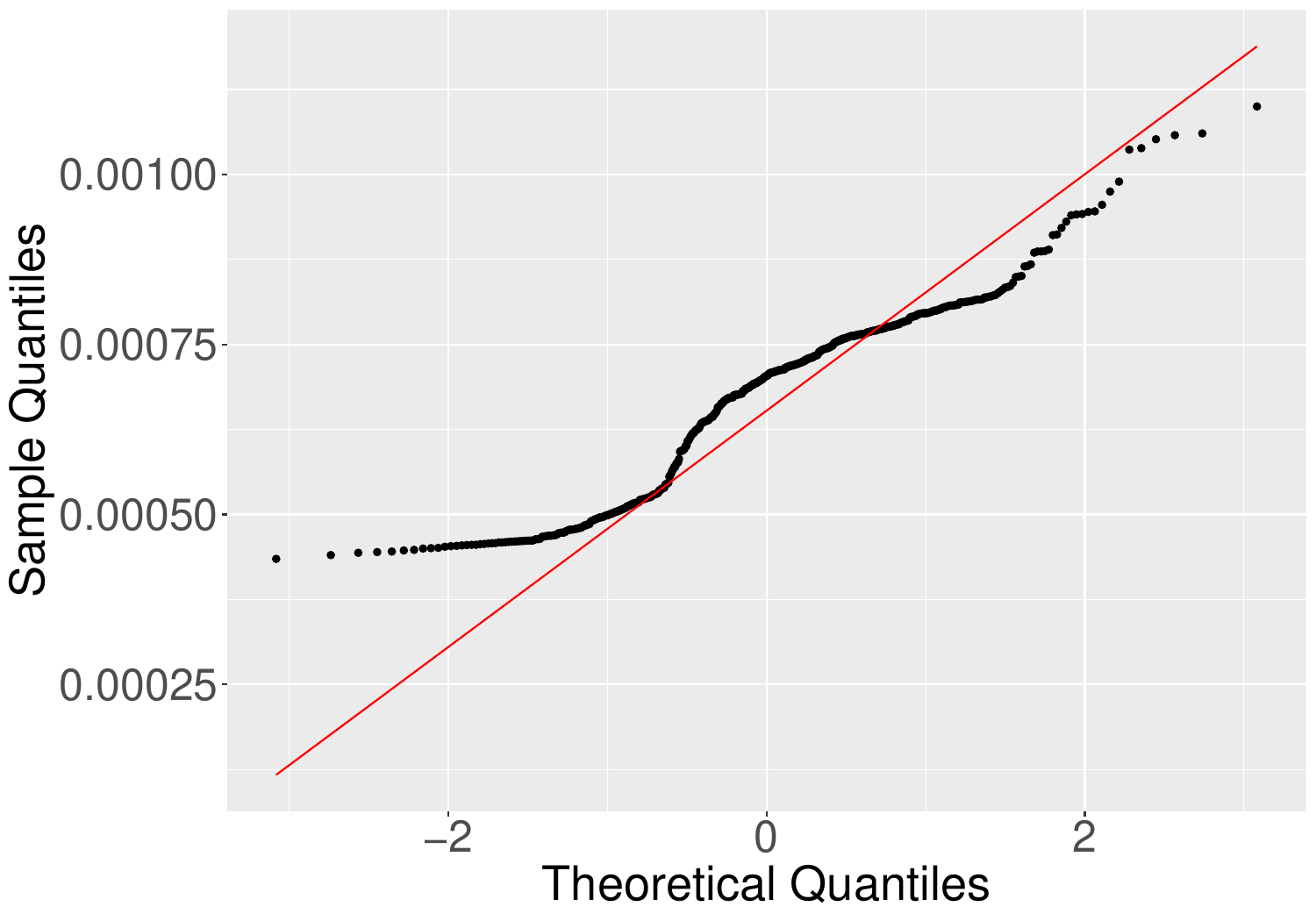}
         \caption{Flux\_1}
     \end{subfigure}
        \caption{Q-Q plots comparing the energy consumption of the analyzed models across various prompt length to the normal distribution. Each plot visualizes the distribution of energy consumption values for a specific model, assessing how closely they follow a normal distribution.}
        \label{fig:qqplot_length}
\end{figure}

Before performing the correlation analysis between the length of prompts and the energy consumed during the image generation process, we verified if the energy consumption for each model followed or not a normal distribution, by examining Quantile-Quantile (QQ) plots, which are illustrated in \Cref{fig:qqplot_length}. 
If the data closely follows the normal distribution, the points on the Q-Q plot will move on a diagonal line (depicted in red), while deviations from the reference line indicate departures from the expected distribution.
\Cref{fig:qqplot_length} shows that for all the analyzed models, data does not follow a normal distribution, as the points deviate significantly from the straight diagonal line.
This step ensured the validity of applying Kruskal-Wallis test, which is a non-parametric statistical test, which does not assume a normally distributed data.

\begin{figure}[h]
     \centering
     \begin{subfigure}[b]{0.23\textwidth}
         \centering
         \includegraphics[width=\textwidth]{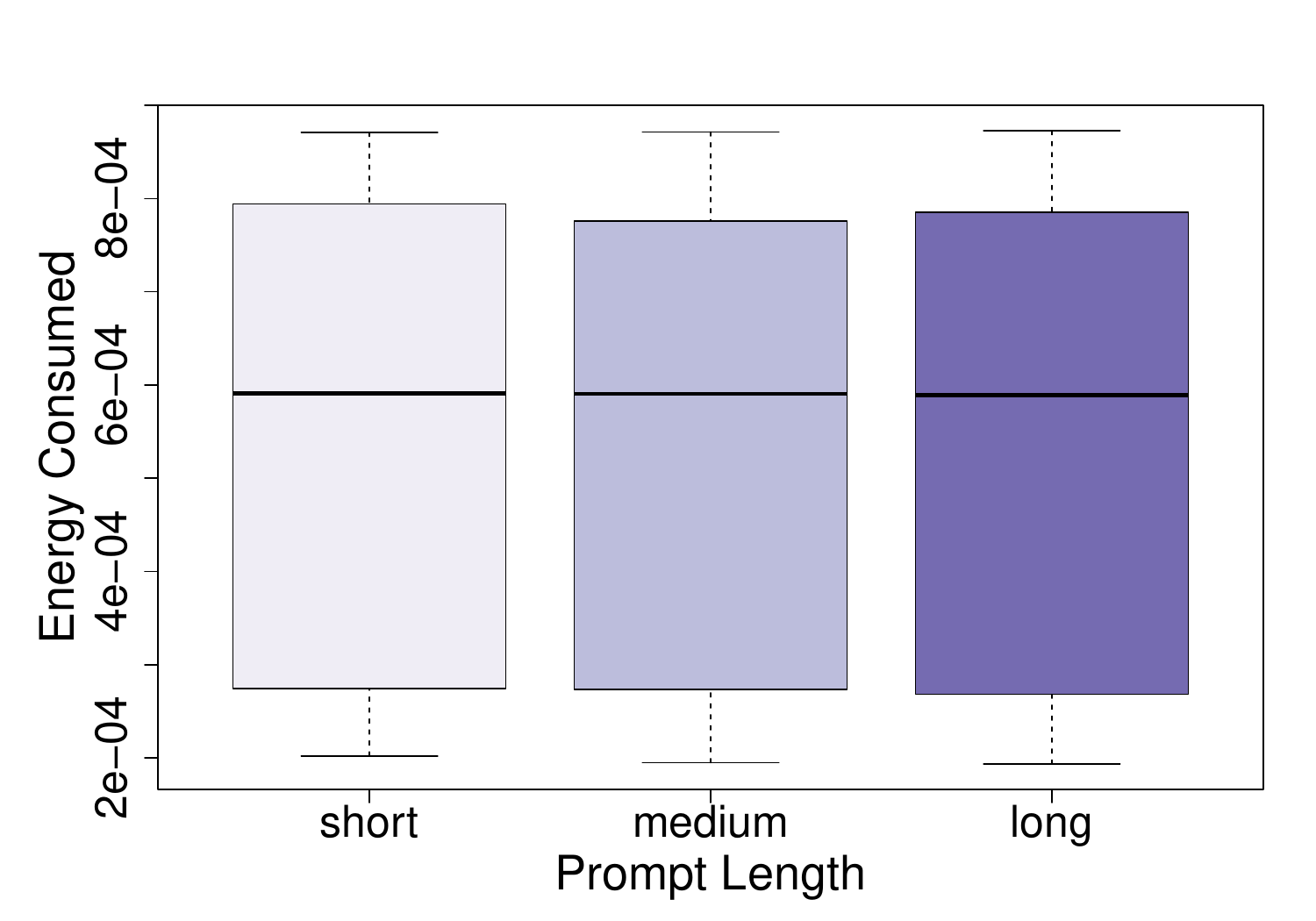}
         \caption{SD\_1.5}
     \end{subfigure}
     \hfill
     \begin{subfigure}[b]{0.23\textwidth}
         \centering
         \includegraphics[width=\textwidth]{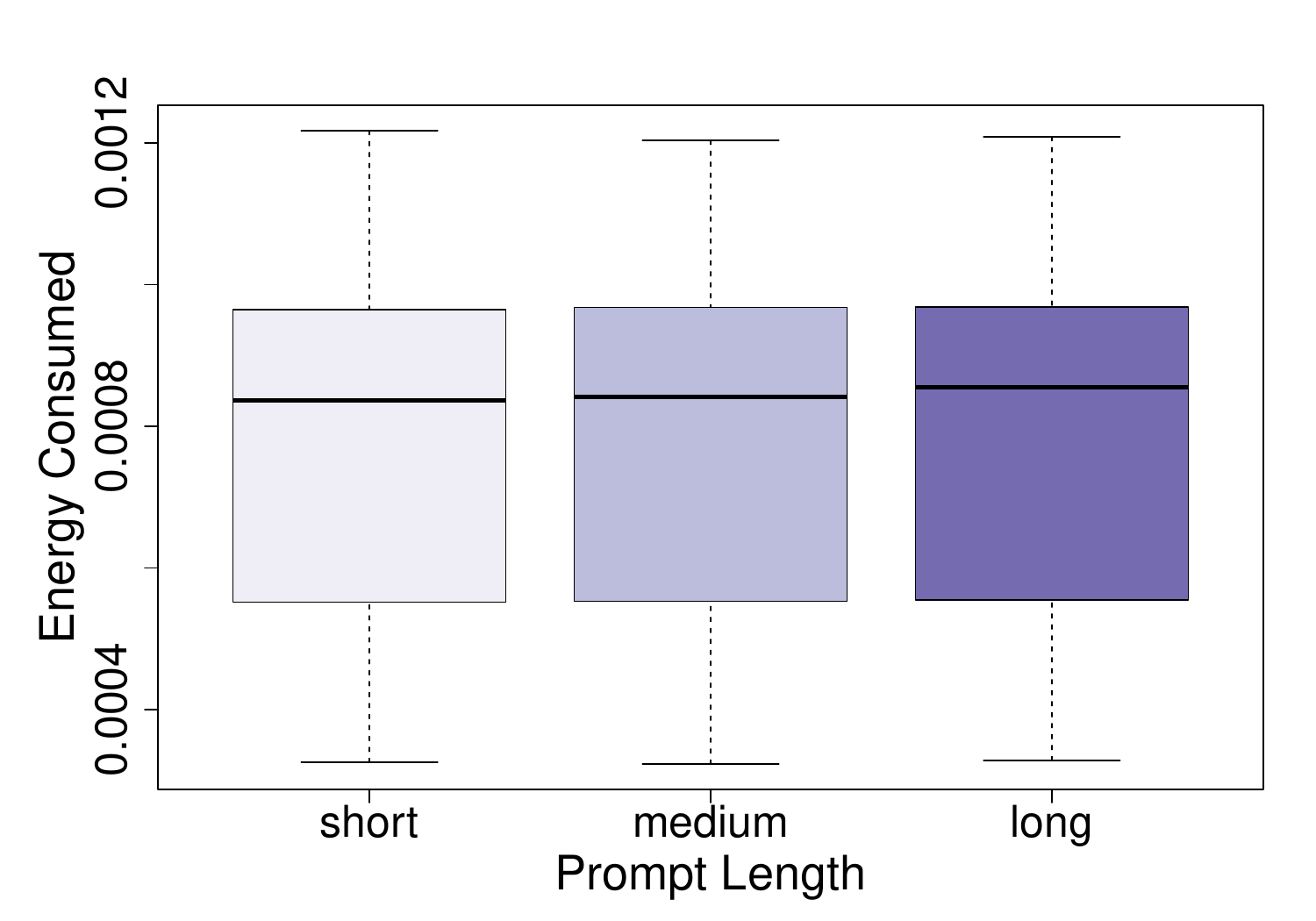}
         \caption{SDXL}
     \end{subfigure}
     \hfill
     \begin{subfigure}[b]{0.23\textwidth}
         \centering
         \includegraphics[width=\textwidth]{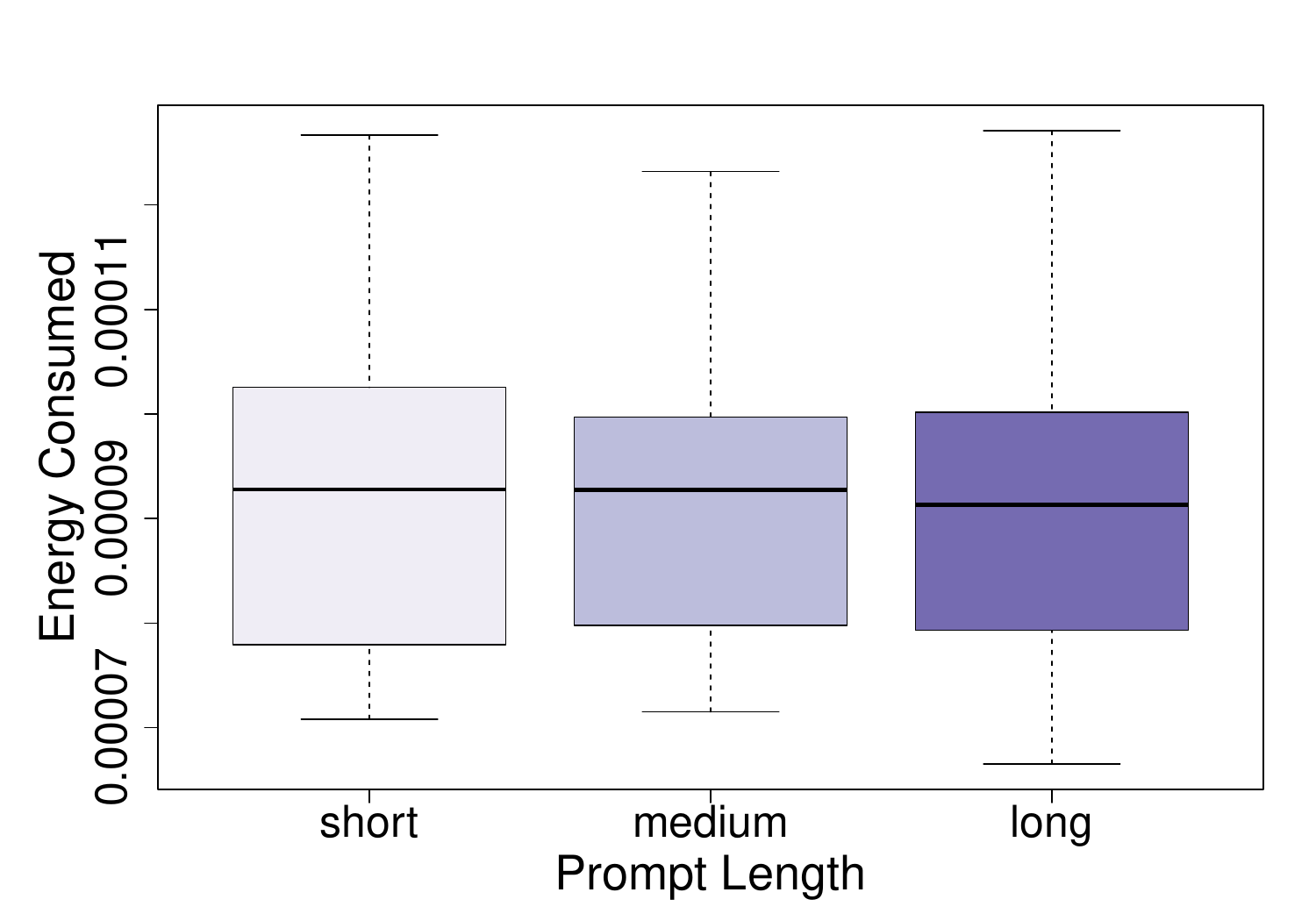}
         \caption{SDXL\_Turbo}
     \end{subfigure}
     \hfill
     \begin{subfigure}[b]{0.23\textwidth}
         \centering
         \includegraphics[width=\textwidth]{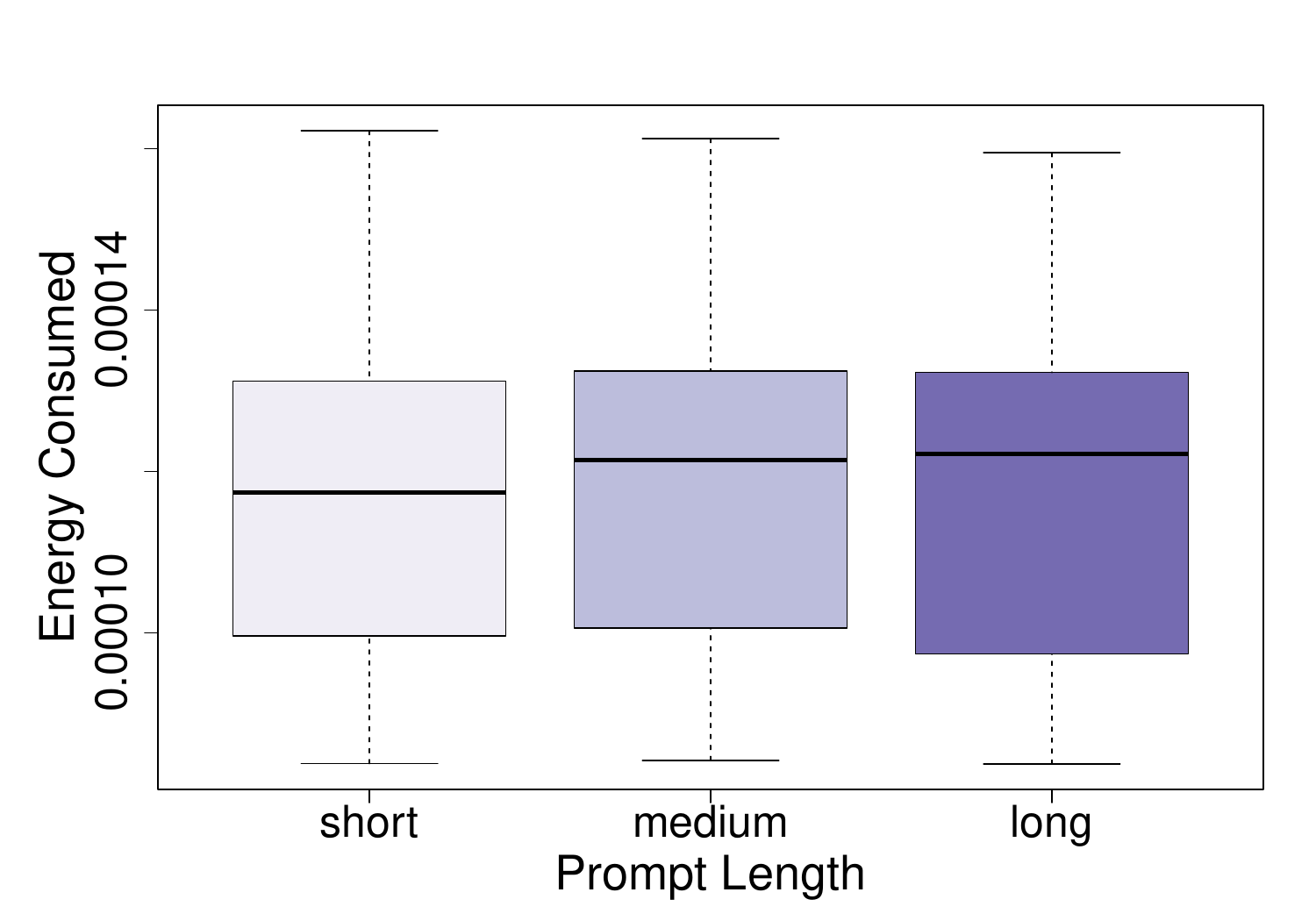}
         \caption{SDXL\_Lightning}
     \end{subfigure}
     \begin{subfigure}[b]{0.23\textwidth}
         \centering
         \includegraphics[width=\textwidth]{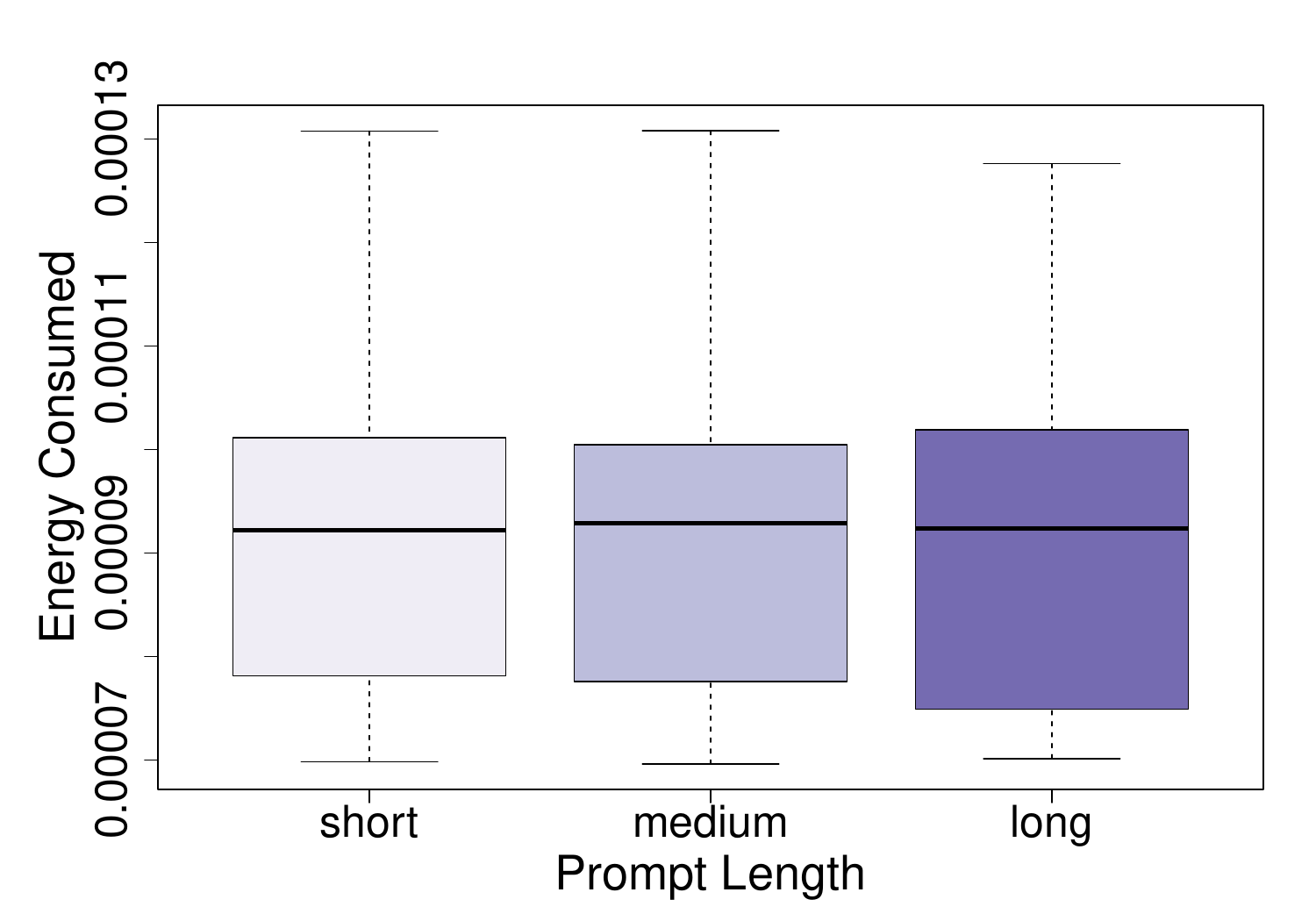}
         \caption{Hyper\_SD}
     \end{subfigure}
     \hfill
     \begin{subfigure}[b]{0.23\textwidth}
         \centering
         \includegraphics[width=\textwidth]{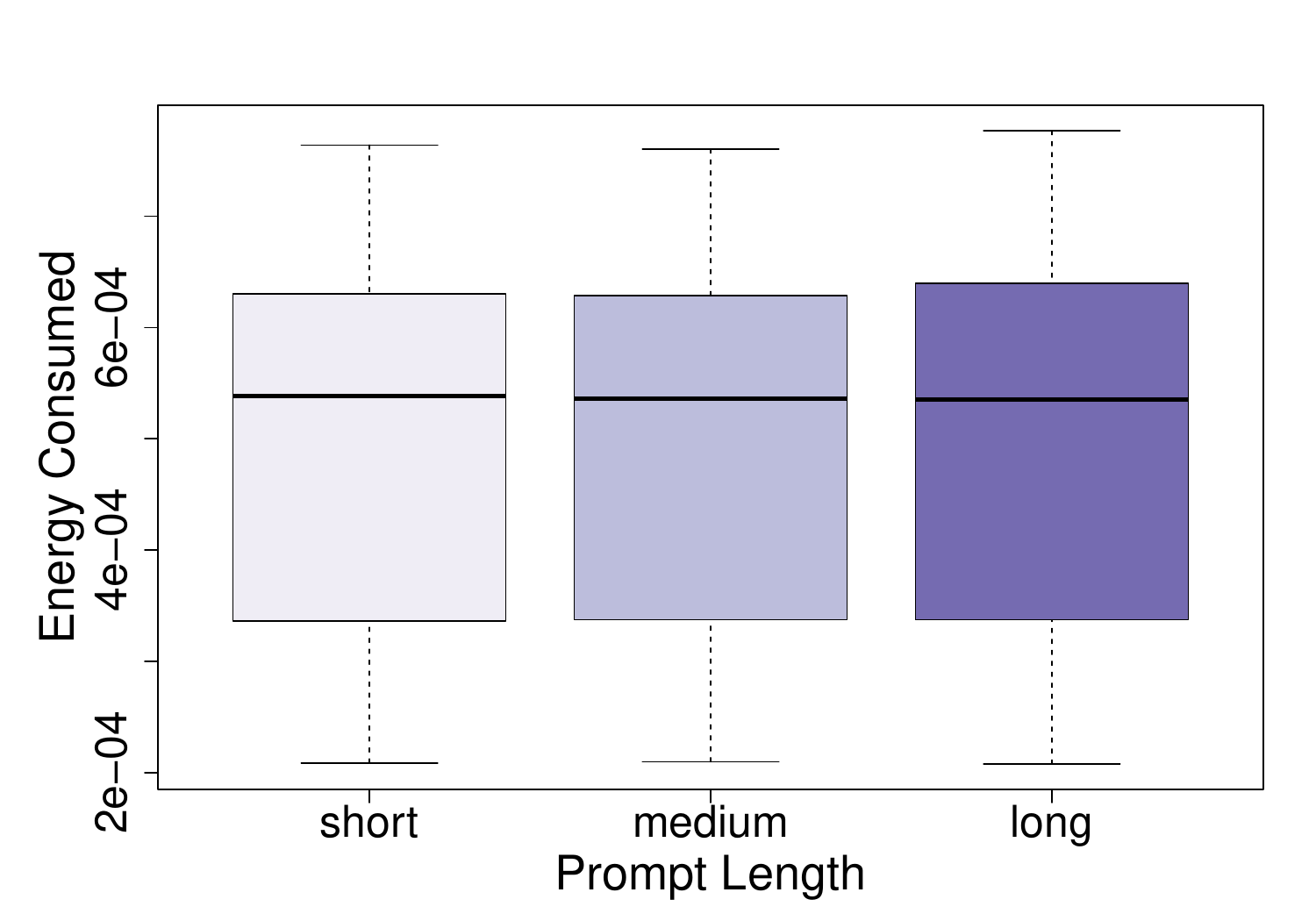}
         \caption{SSD\_1B}
     \end{subfigure}
     \hfill
     \begin{subfigure}[b]{0.23\textwidth}
         \centering
         \includegraphics[width=\textwidth]{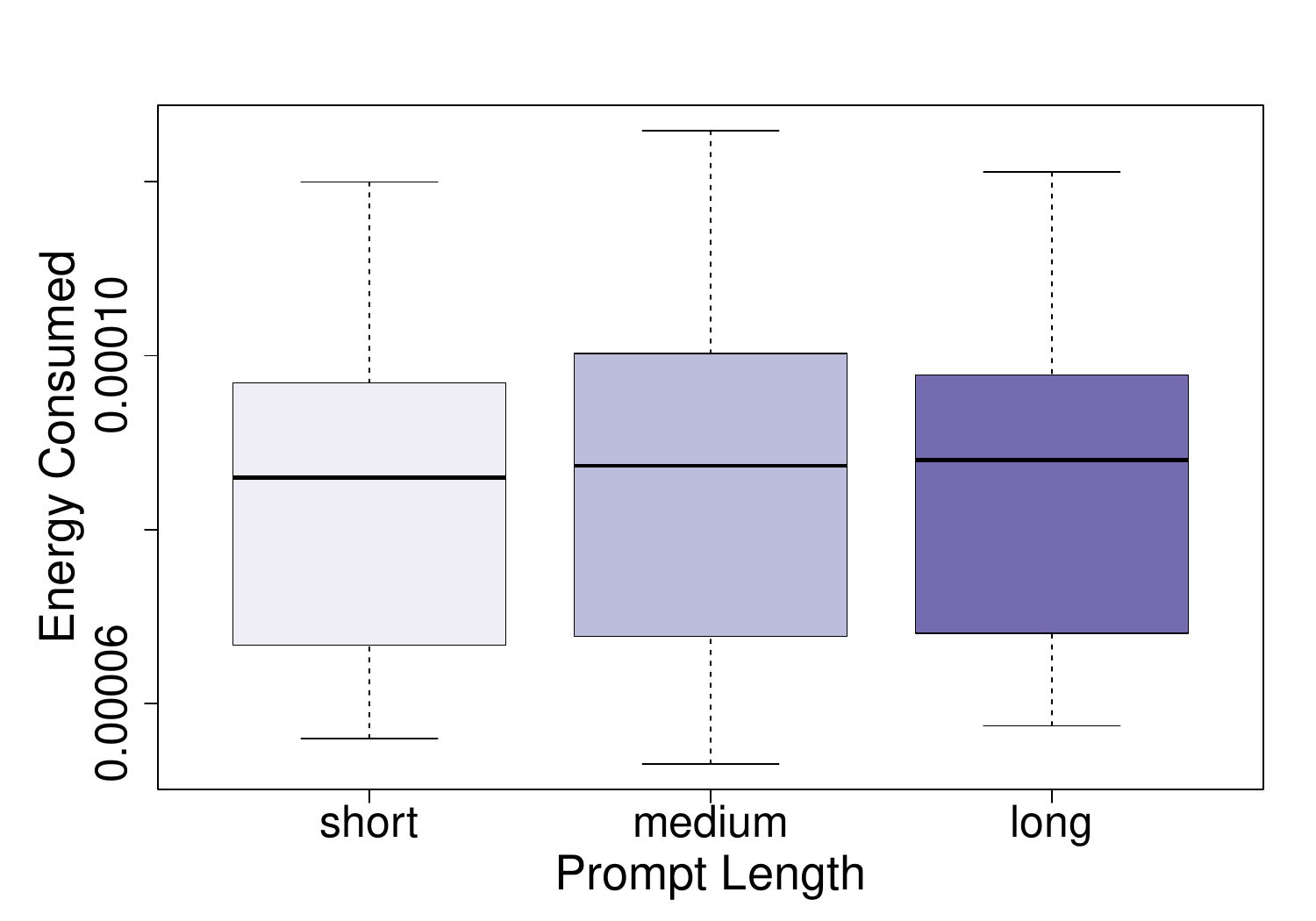}
         \caption{LCM\_SSD\_1B}
     \end{subfigure}
     \hfill
     \begin{subfigure}[b]{0.23\textwidth}
         \centering
         \includegraphics[width=\textwidth]{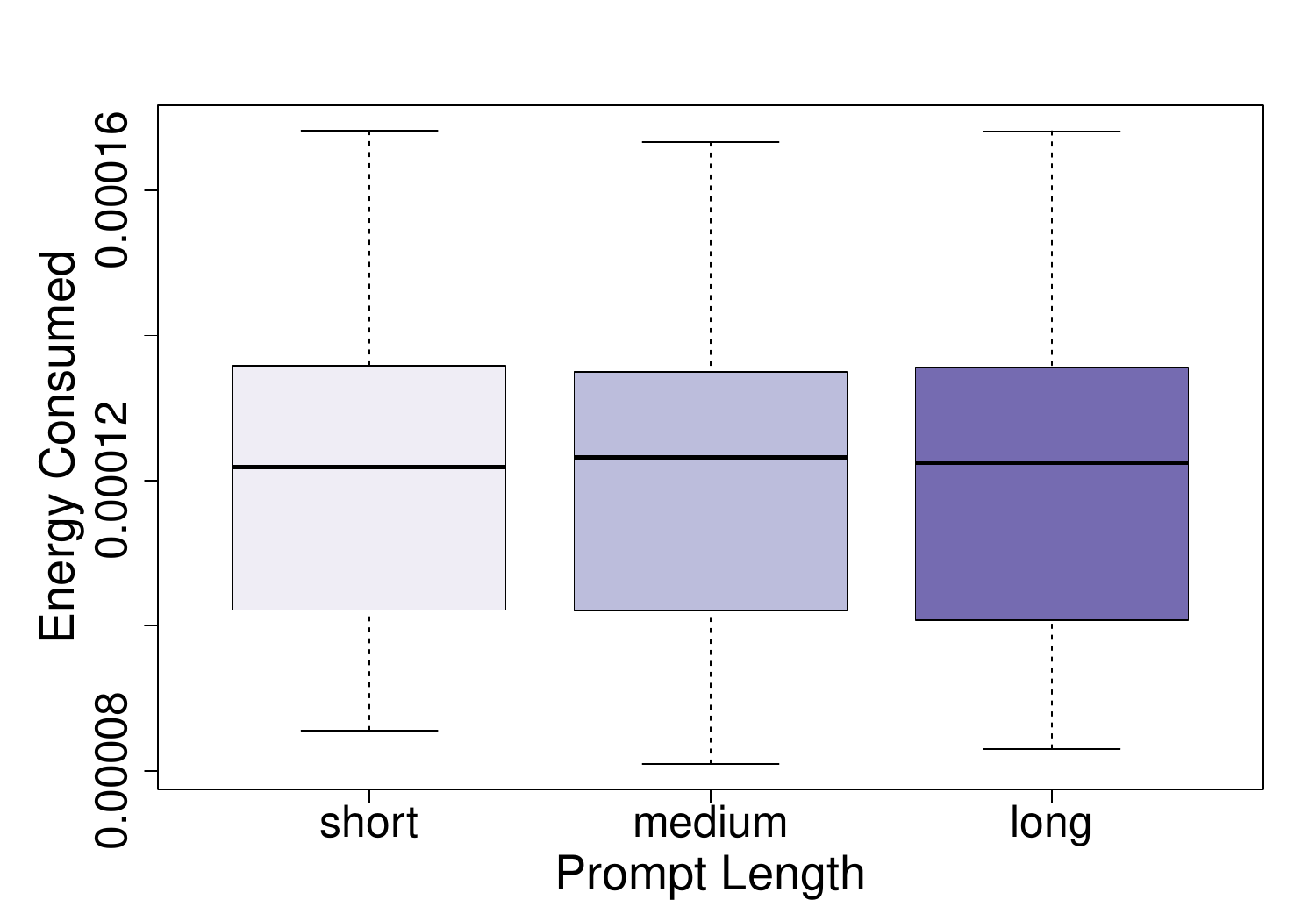}
         \caption{LCM\_SDXL}
     \end{subfigure}
     \begin{subfigure}[b]{0.23\textwidth}
         \centering
         \includegraphics[width=\textwidth]{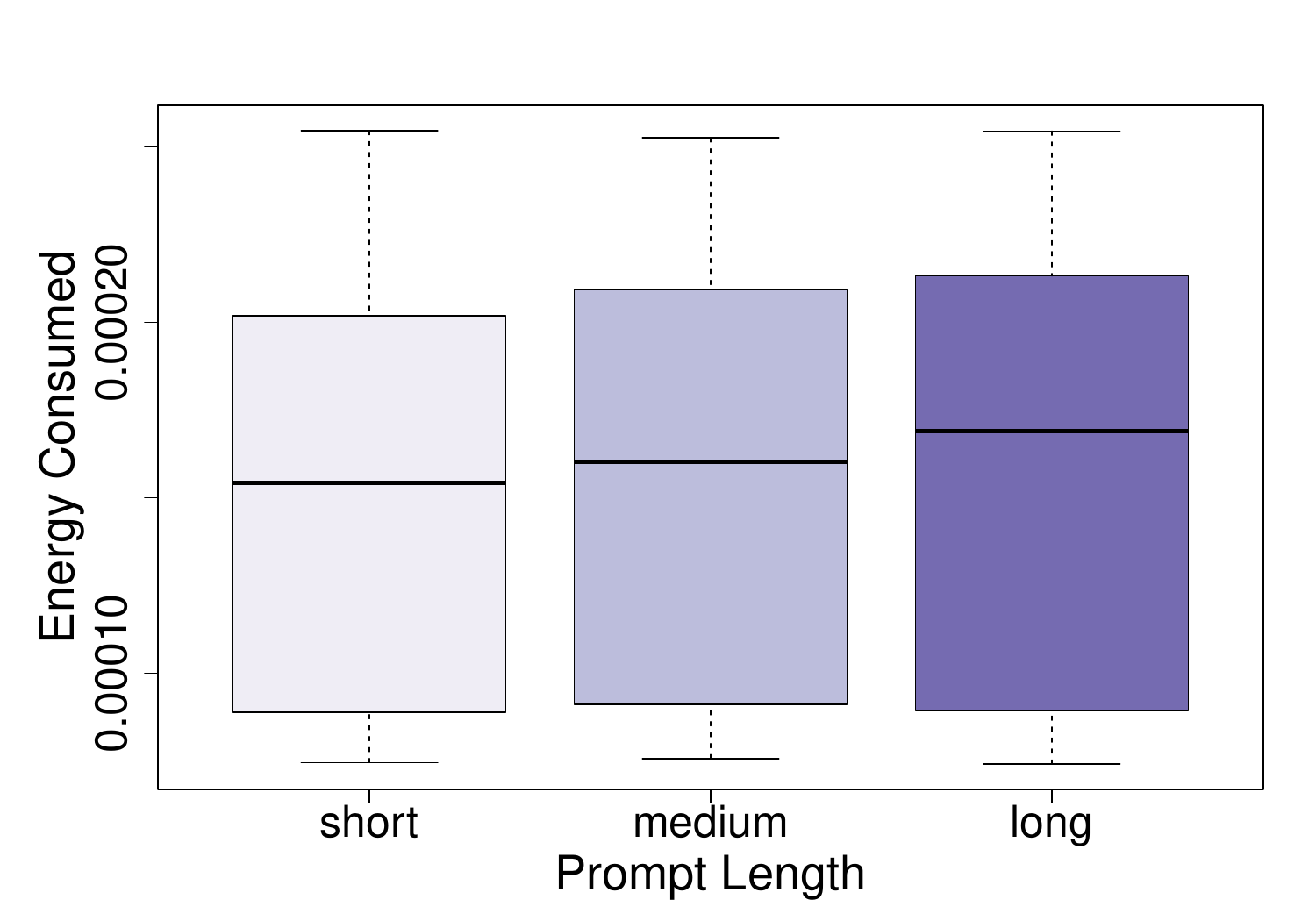}
         \caption{Flash\_SD}
     \end{subfigure}
     \hfill
     \begin{subfigure}[b]{0.23\textwidth}
         \centering
         \includegraphics[width=\textwidth]{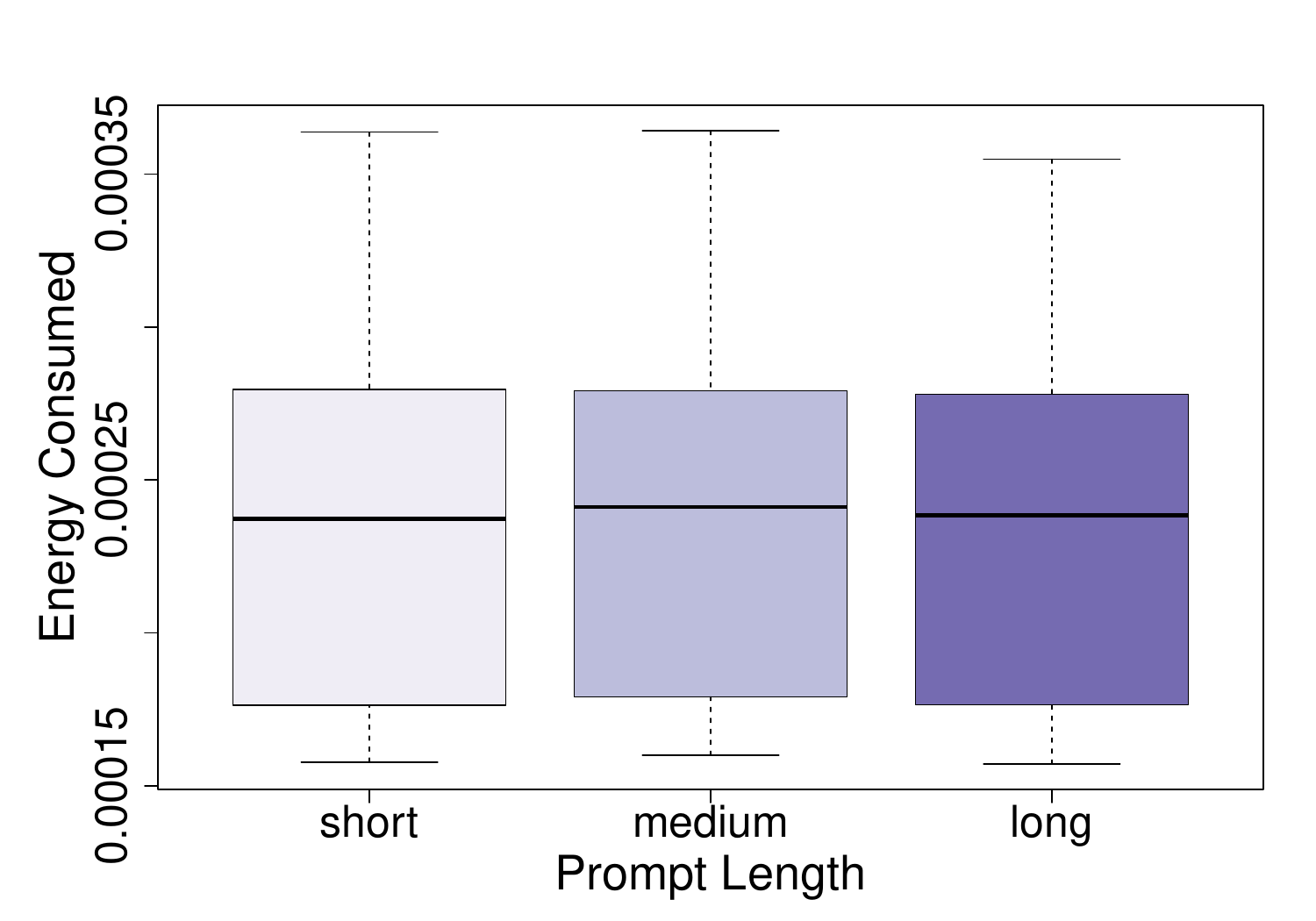}
         \caption{Flash\_SDXL}
     \end{subfigure}
     \hfill
     \begin{subfigure}[b]{0.23\textwidth}
         \centering
         \includegraphics[width=\textwidth]{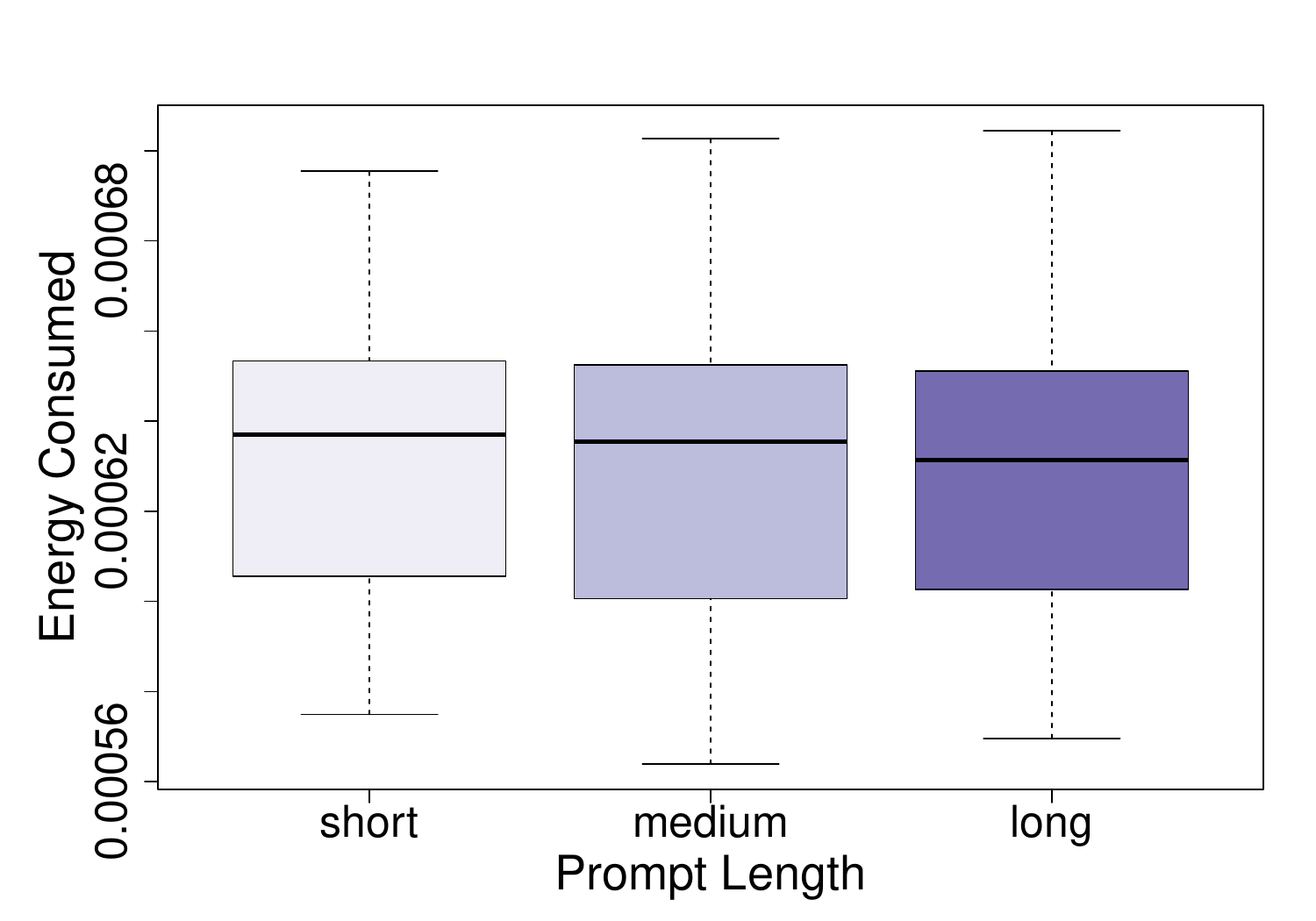}
         \caption{PixArt\_Alpha}
     \end{subfigure}
     \hfill
     \begin{subfigure}[b]{0.23\textwidth}
         \centering
         \includegraphics[width=\textwidth]{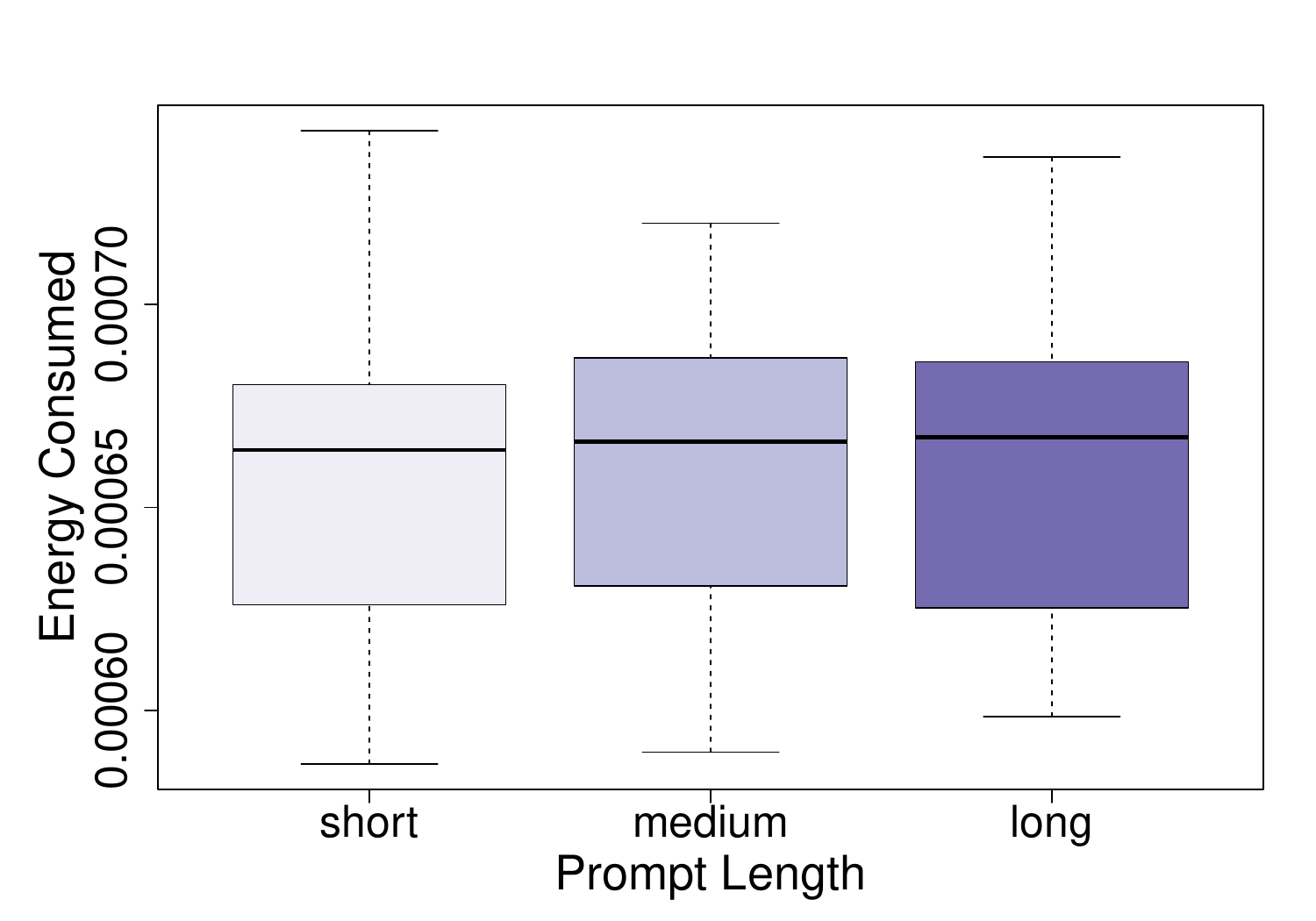}
         \caption{PixArt\_Sigma}
     \end{subfigure}
     \hfill
     \begin{subfigure}[b]{0.23\textwidth}
         \centering
         \includegraphics[width=\textwidth]{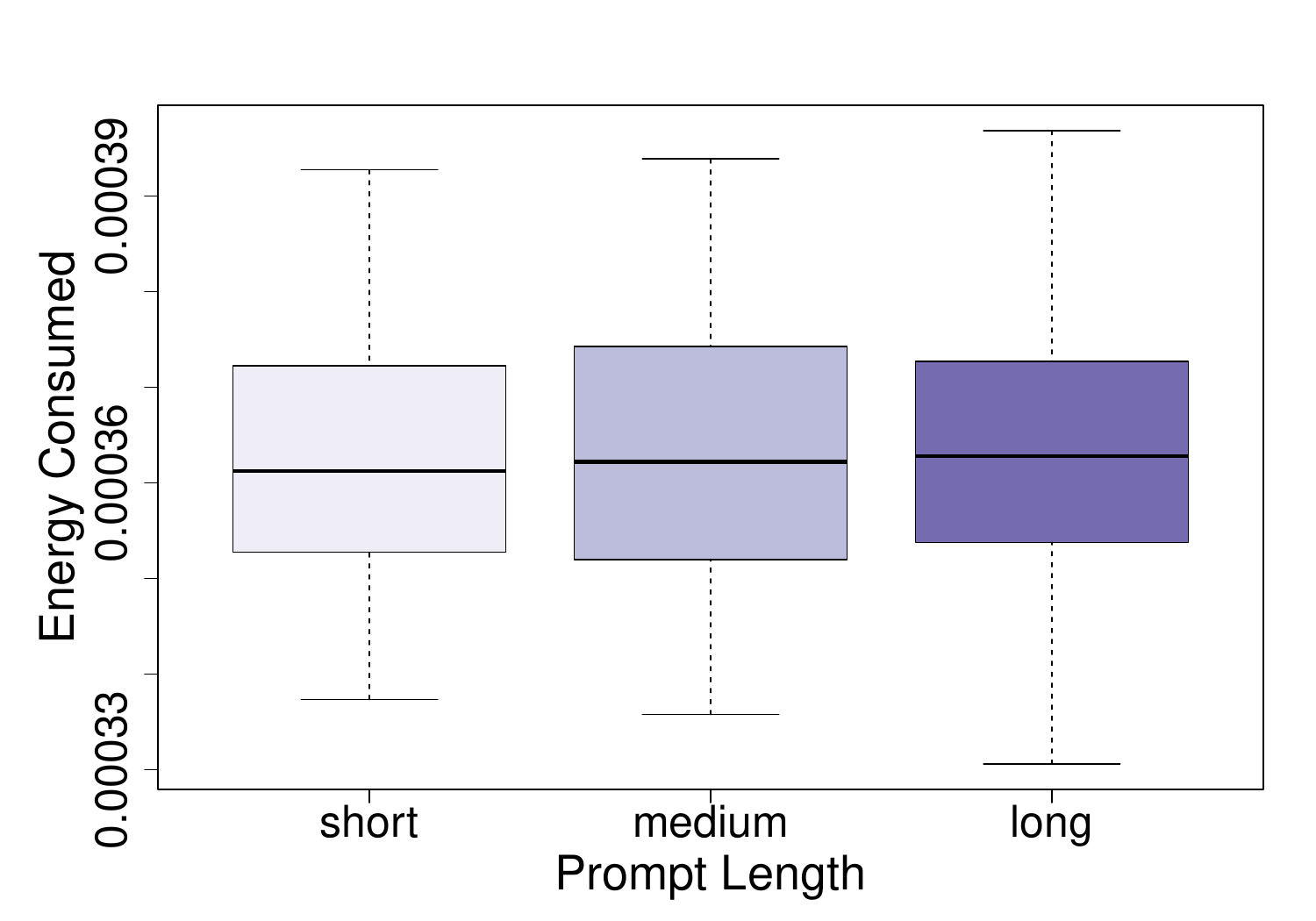}
         \caption{Flash\_PixArt}
     \end{subfigure}
     \hfill
     \begin{subfigure}[b]{0.23\textwidth}
         \centering
         \includegraphics[width=\textwidth]{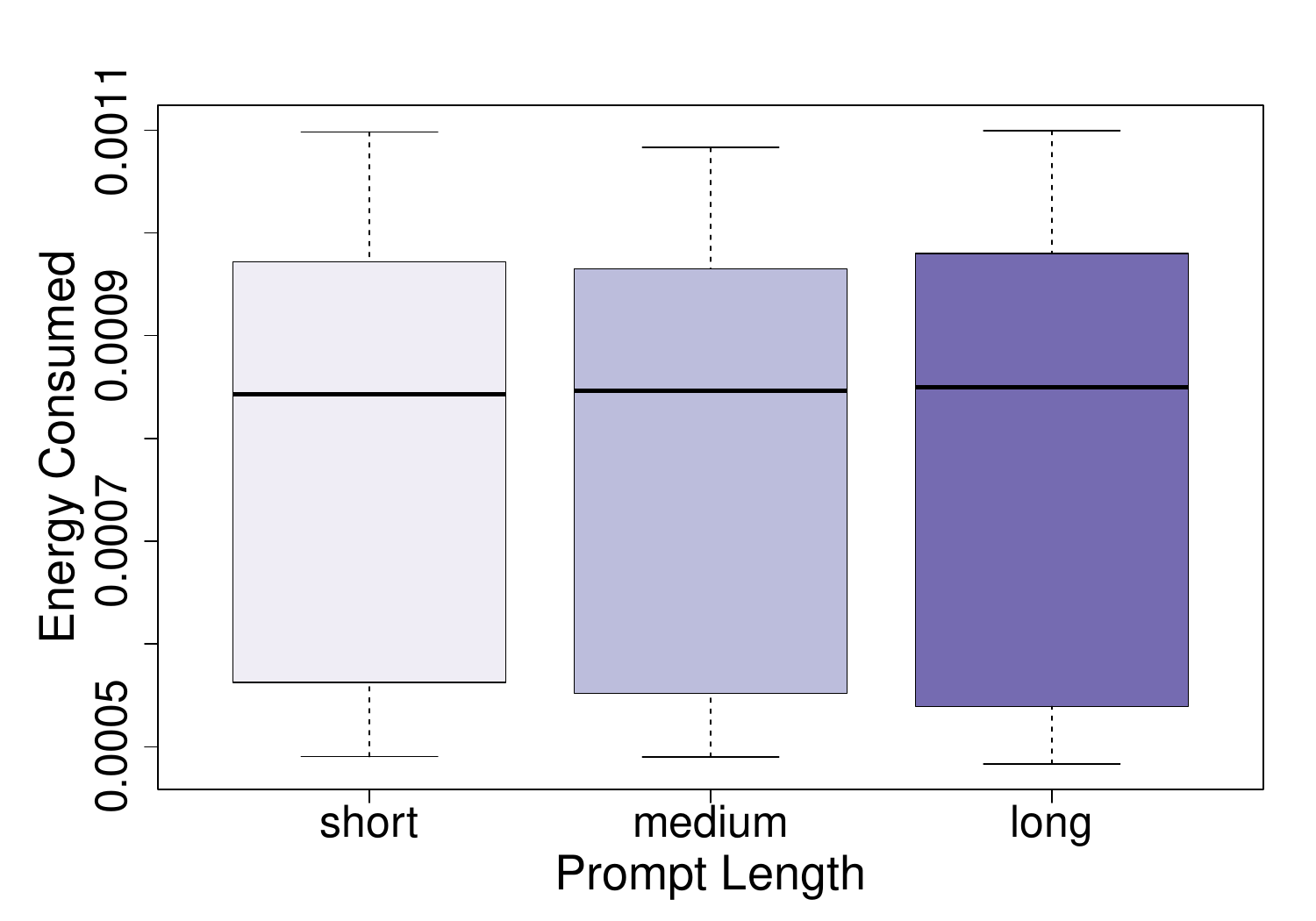}
         \caption{SD\_3}
     \end{subfigure}
     \hfill
     \begin{subfigure}[b]{0.23\textwidth}
         \centering
         \includegraphics[width=\textwidth]{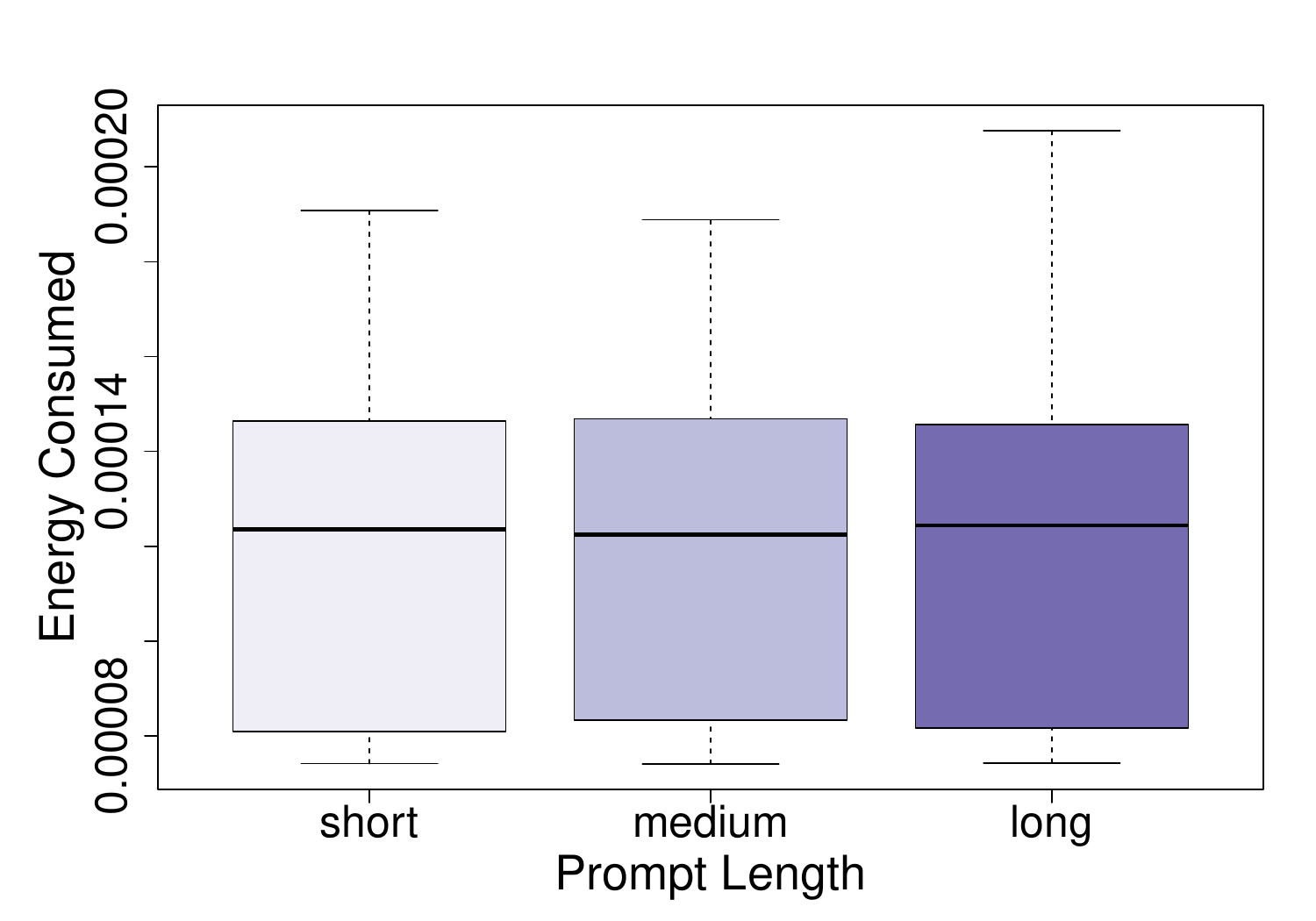}
         \caption{Flash\_SD3}
     \end{subfigure}
     \hfill
     \begin{subfigure}[b]{0.23\textwidth}
         \centering
         \includegraphics[width=\textwidth]{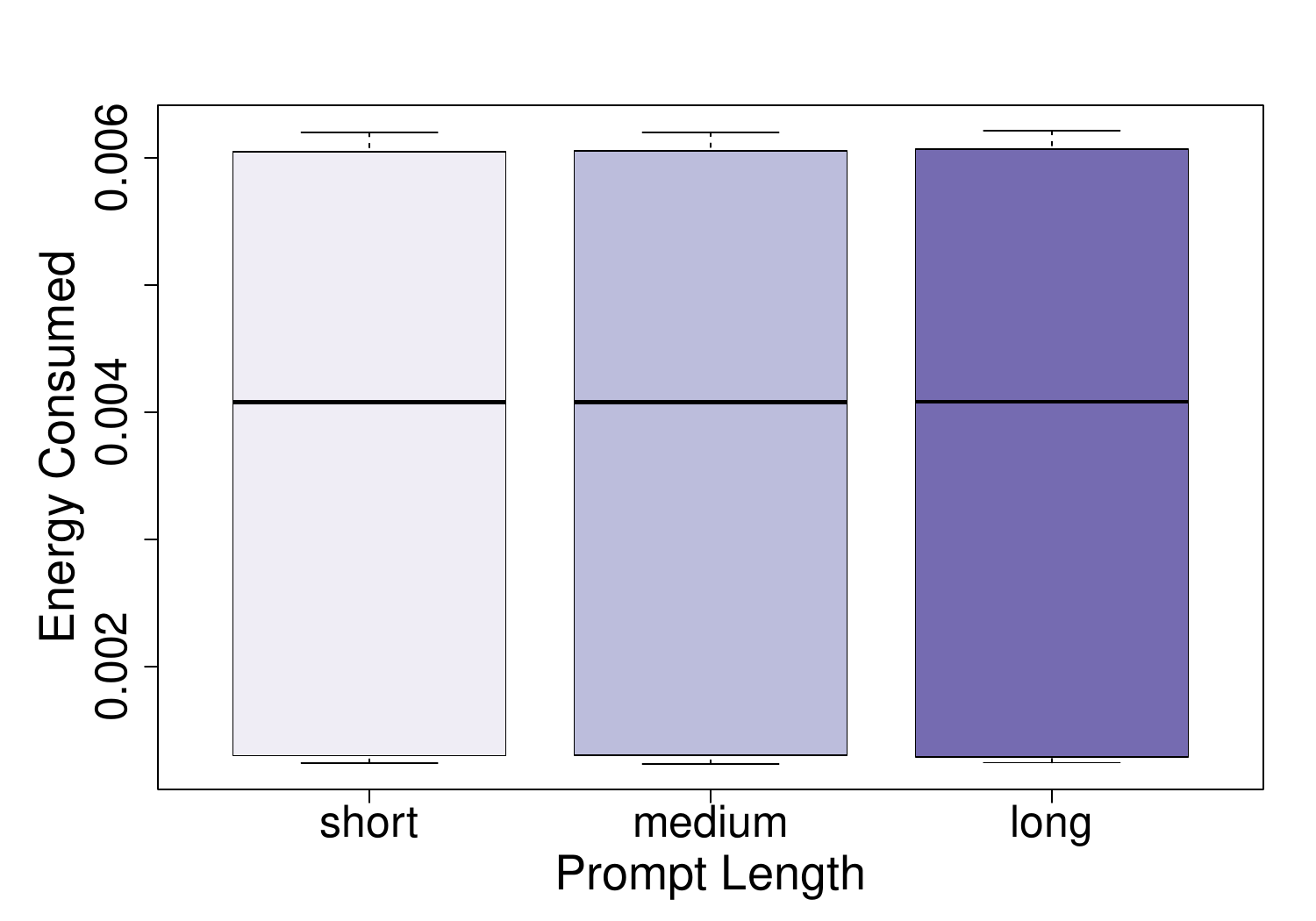}
         \caption{Lumina}
     \end{subfigure}
     \hfill
     \begin{subfigure}[b]{0.23\textwidth}
         \centering
         \includegraphics[width=\textwidth]{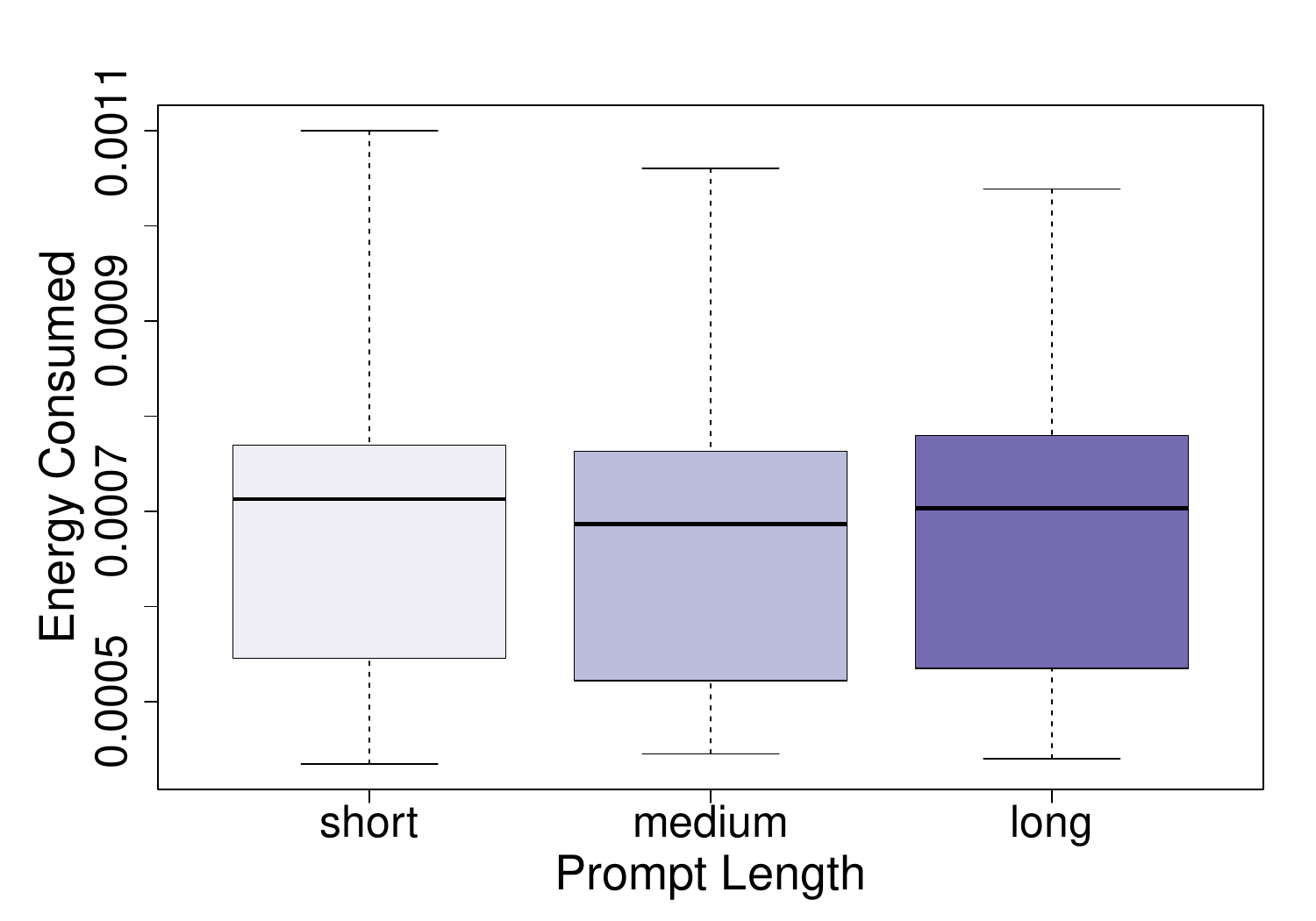}
         \caption{Flux\_1}
     \end{subfigure}
        \caption{Box plots of the energy consumption across various prompt length for the analyzed models.}
        \label{fig:boxplot_length}
\end{figure}

Moreover, \Cref{fig:boxplot_length} show the box plots which represent the distribution of energy consumption across different models for the three different prompt lengths. 
As we can notice, the median energy consumption remains stable across different prompt lengths for all models, with no clear increasing or decreasing trend.
This observation suggests that prompt length does not have a significant influence on energy consumption, as no evident pattern is observed across different models. 
This finding is also supported by the Kruskal-Wallis statistical test. 

\Cref{tab:prompt_length} presents the results of the Kruskal-Wallis test, which examines whether variations in prompt lengths lead to statistically significant differences in energy consumption.

\begin{table}[t]
\caption{Rank sum and p-values obtained from Kruskal-Wallis test to assess the relationship between the length of prompt and energy consumption during image generation process across different models.}\label{tab:prompt_length}
\centering
\adjustbox{width=0.45\textwidth}{
\begin{tabular}{l|c|c}
\toprule
\textbf{Model} & \textbf{Rank Sum} & \textbf{P-value} \\
\midrule
Flash\_PixArt & 0.77 & 0.68 \\
Flash\_SD & 3.66 & 0.16 \\
Flash\_SD3 & 0.08 & 0.96 \\
Flash\_SDXL & 0.95 & 0.62 \\
Flux\_1 & 2.92 & 0.23 \\
Hyper\_SD & 0.06 & 0.97 \\
LCM\_SDXL & 0.04 & 0.98 \\
LCM\_SSD\_1B & 0.25 & 0.88 \\
Lumina & 0.26 & 0.88 \\
PixArt\_Alpha & 0.75 & 0.69 \\
PixArt\_Sigma & 0.51 & 0.77 \\
SD\_1.5 & 0.56 & 0.76 \\
SD\_3 & 0.25 & 0.88 \\
SDXL & 0.14 & 0.93 \\
SDXL\_Lightning & 0.92 & 0.63 \\
SDXL\_Turbo & 2.99 & 0.23 \\
SSD\_1B & 0.20 & 0.91 \\
\bottomrule
\end{tabular}}
\end{table}

\section{Quality Assessment}\label{app:quality_assessment}

As previously stated in \Cref{sec:qass}, the validation set of ImageNet-1k was properly filtered to ensure that only the classes matching with the prompt labels were retained, with random uniform sampling in case of multiple matches. Unfortunately, for prompts \textit{apples}, \textit{girl}, \textit{poppies}, \textit{acquarium fish}, \textit{oak}, \textit{shrew}, \textit{ray} and \textit{tulips} were not possible to find a matching ImageNet-1k class. Consequently, we excluded all images associated with the aforementioned prompts from the calculation of the quality metrics.

\begin{figure}
    \centering
     \begin{subfigure}[b]{0.45\textwidth}
         \centering
         \includegraphics[width=\textwidth]{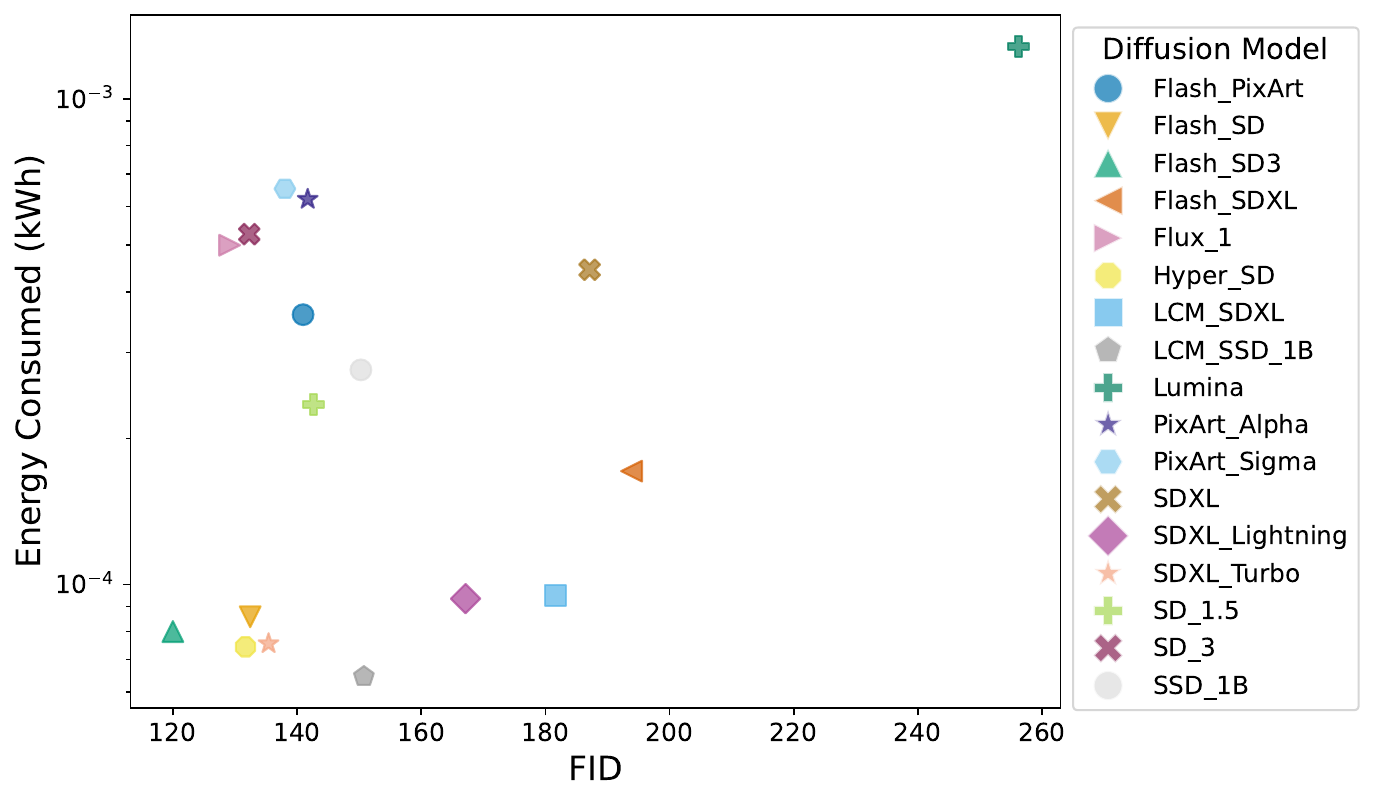}
         \caption{}
     \end{subfigure}
     \begin{subfigure}[b]{0.45\textwidth}
         \centering
         \includegraphics[width=\textwidth]{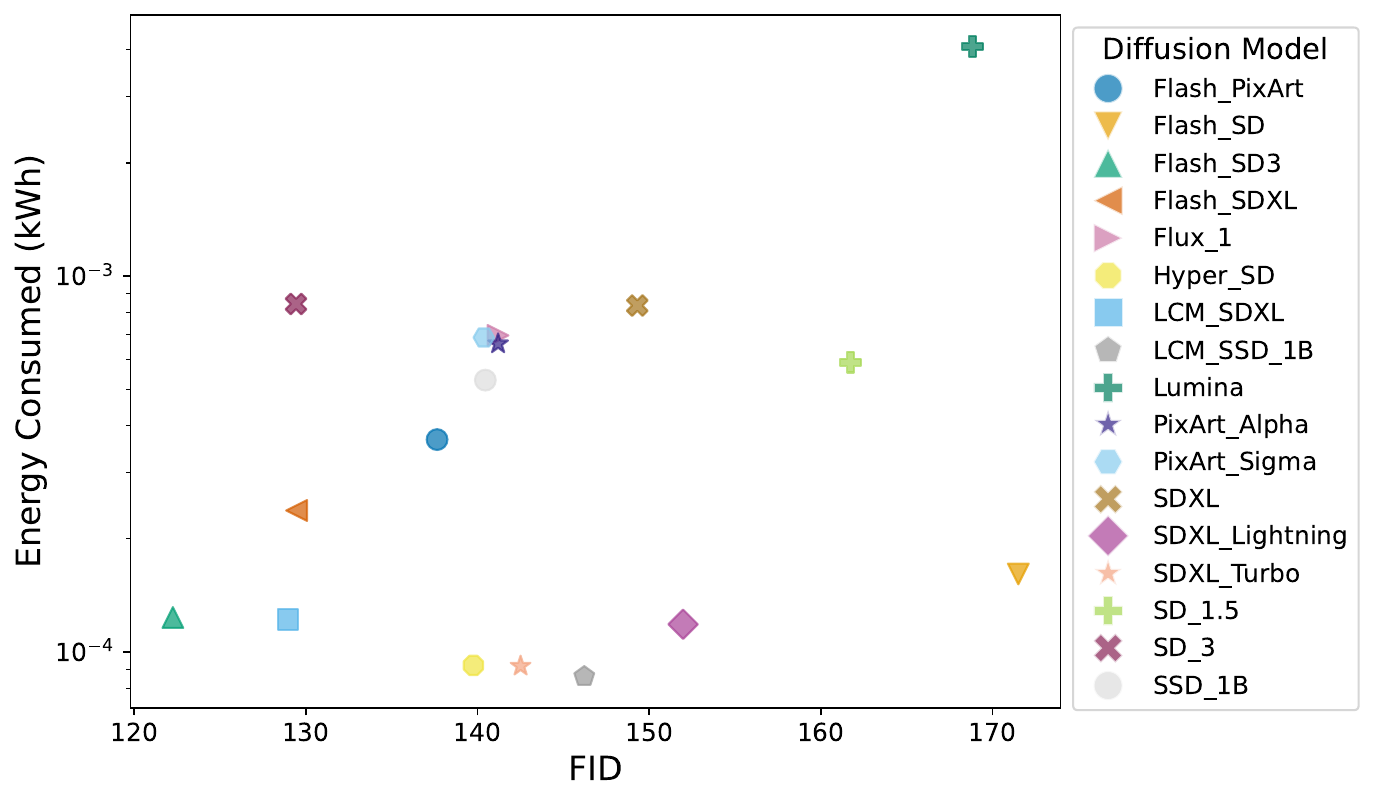}
         \caption{}
     \end{subfigure}
     \begin{subfigure}[b]{0.45\textwidth}
         \centering
         \includegraphics[width=\textwidth]{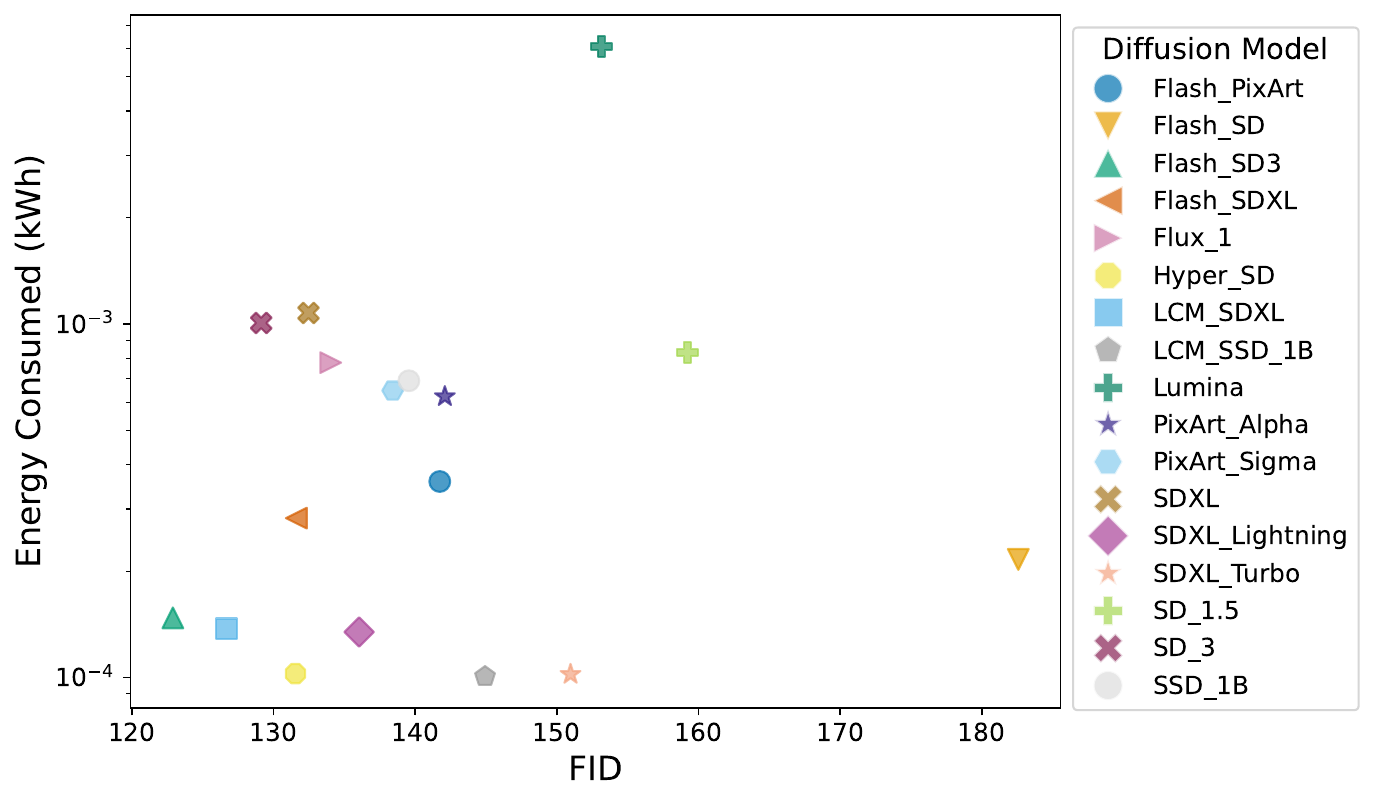}
         \caption{}
     \end{subfigure}
      \caption{Energy consumption (kWh) vs FID values for all the models at resolution (a) 512$\times$512 (b) 768$\times$1024 and (c) 1024$\times$1024.}
        \label{fig:fid_energy_diff_res}
\end{figure}

\begin{figure}
    \centering
     \begin{subfigure}[b]{0.45\textwidth}
         \centering
         \includegraphics[width=\textwidth]{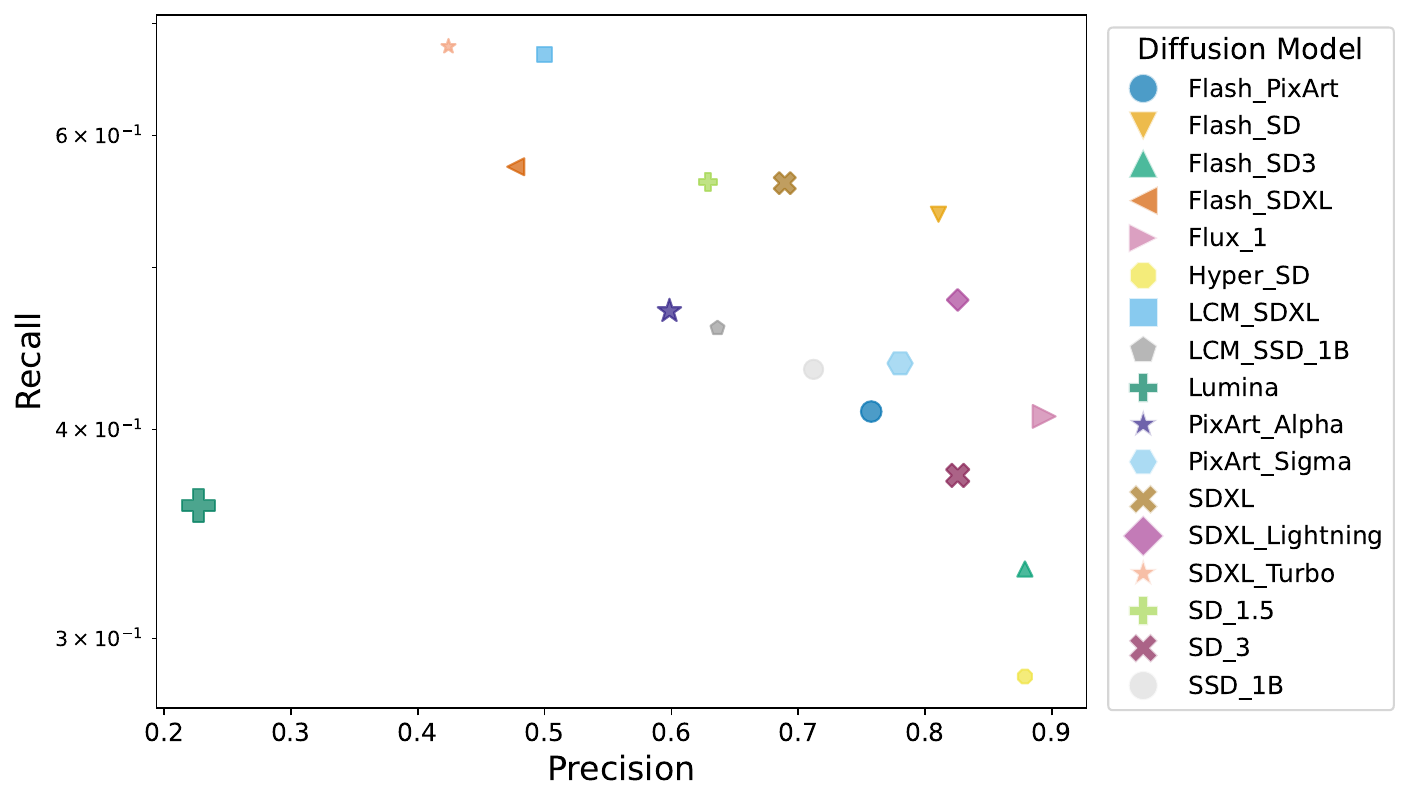}
         \caption{}
     \end{subfigure}
     \begin{subfigure}[b]{0.45\textwidth}
         \centering
         \includegraphics[width=\textwidth]{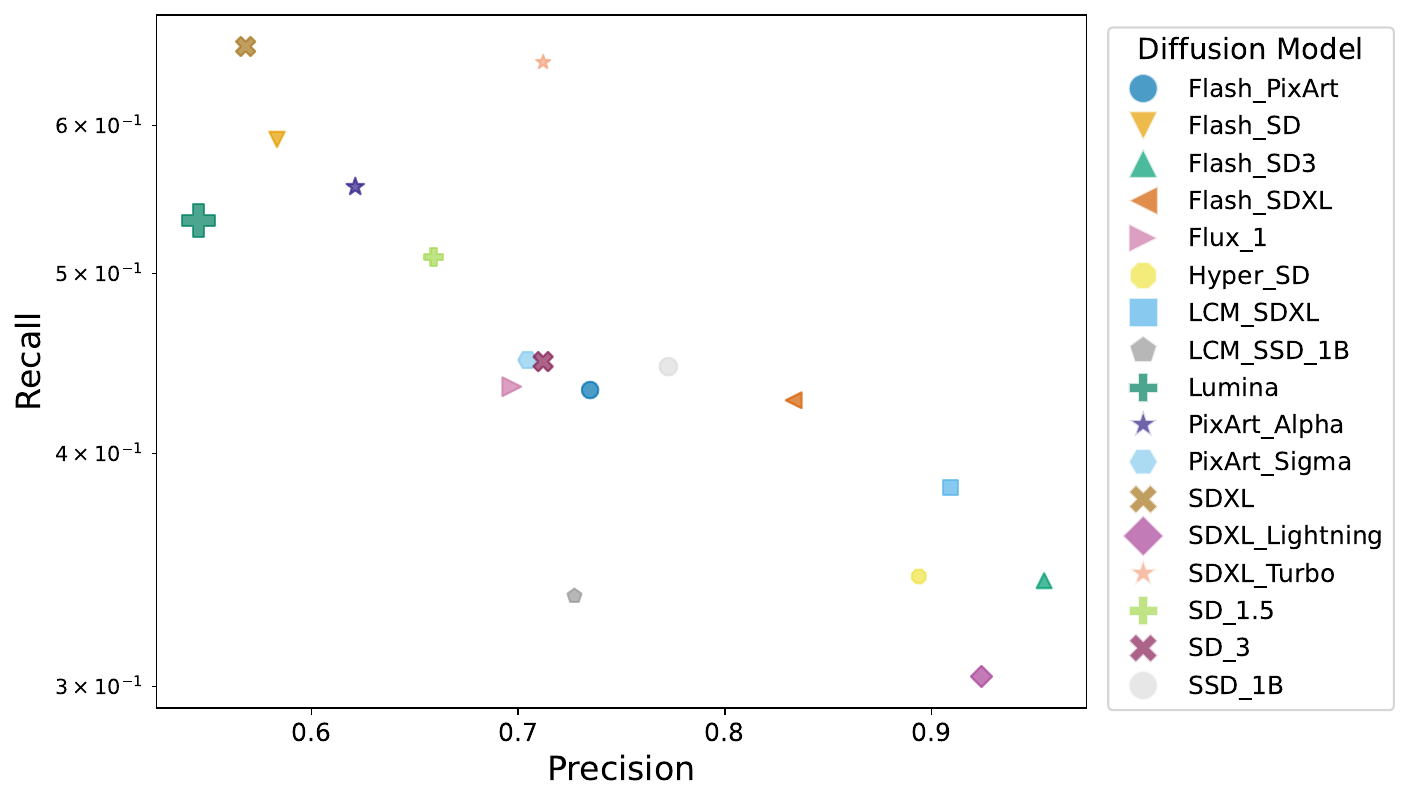}
         \caption{}
     \end{subfigure}
     \begin{subfigure}[b]{0.45\textwidth}
         \centering
         \includegraphics[width=\textwidth]{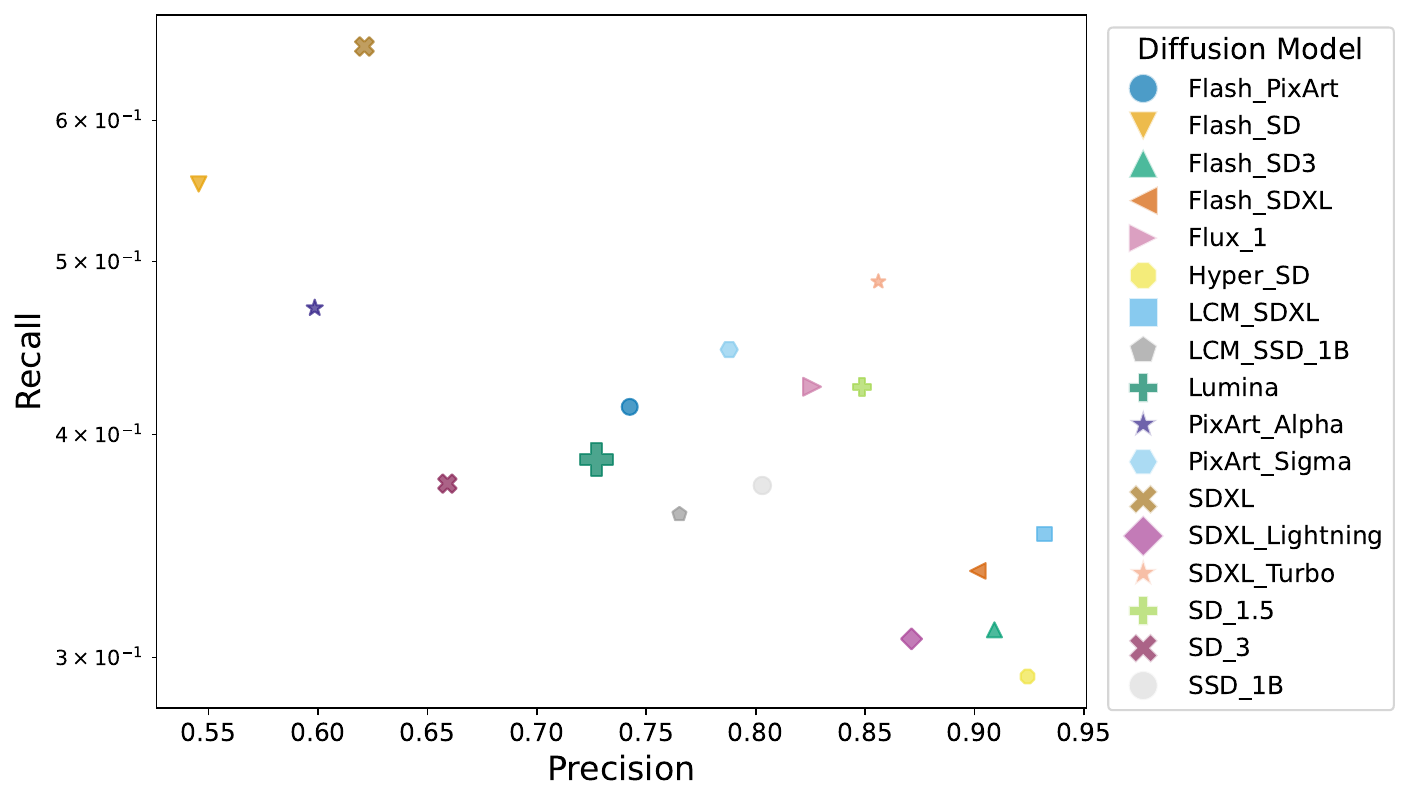}
         \caption{}
     \end{subfigure}
      \caption{Precision and Recall values for each model at resolution (a) 512$\times$512 (b) 768$\times$1024 and (c) 1024$\times$1024. Each symbol of the plot has different size based on the energy consumed.}
        \label{fig:pr_energy_diff_res}
\end{figure}

\begin{figure}
    \centering
    \includegraphics[width=\linewidth]{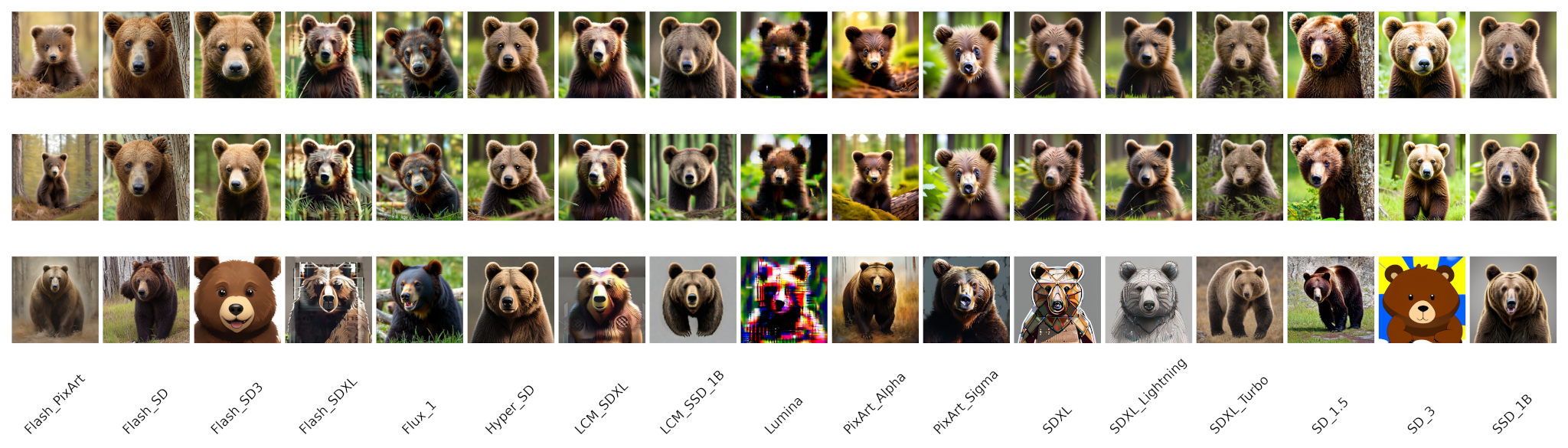}
    \caption{Example images generated by each model at resolution 512$\times$512, quantization int8 and prompt ``bear". Each row corresponds to a different prompt length: long (top), medium (middle), and short (bottom).}
    \label{fig:prompt_length_512x512}
\end{figure}

\begin{figure}
    \centering
    \includegraphics[width=\linewidth]{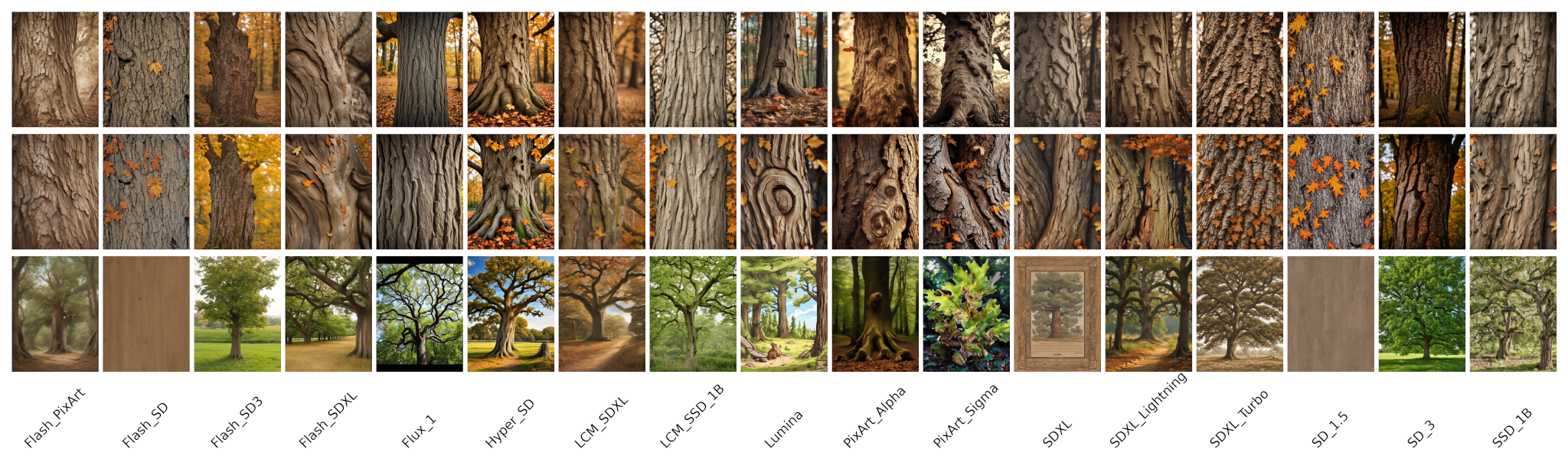}
    \caption{Example images generated by each model at resolution 768$\times$1024, quantization int8 and prompt ``tree". Each row corresponds to a different prompt length: long (top), medium (middle), and short (bottom).}
    \label{fig:prompt_length_768x1024}
\end{figure}

\begin{figure}
    \centering
    \includegraphics[width=\linewidth]{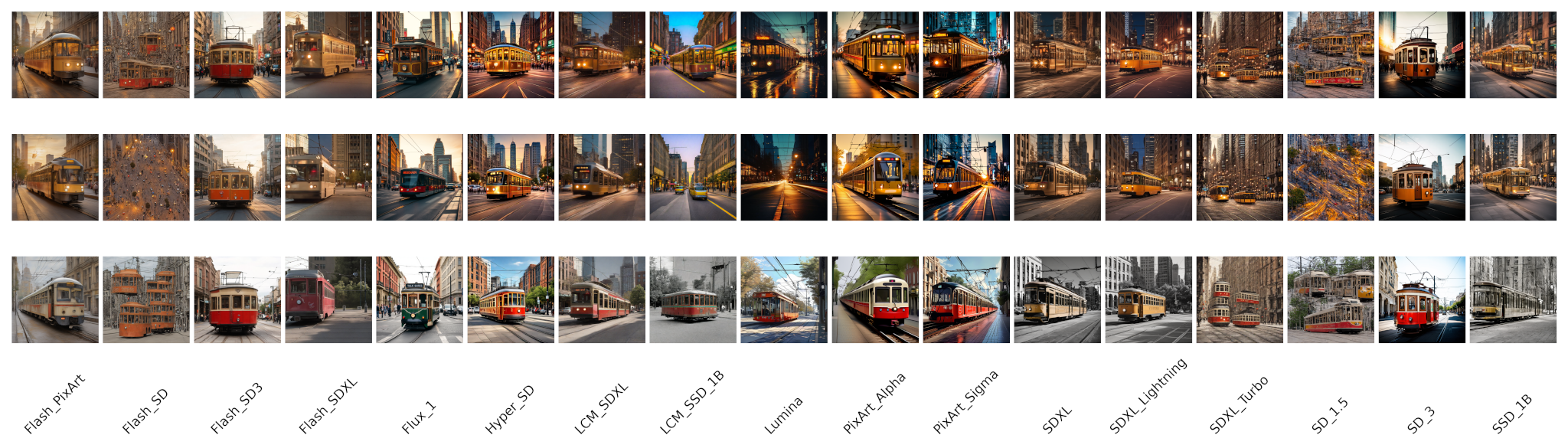}
    \caption{Example images generated by each model at resolution 1024$\times$1024, quantization int8 and prompt ``streetcar". Each row corresponds to a different prompt length: long (top), medium (middle), and short (bottom).}
    \label{fig:prompt_length_1024x1024}
\end{figure}

\Cref{fig:fid_energy_diff_res} and \Cref{fig:pr_energy_diff_res} illustrate how the quality-energy relationship varies when considering images at specific resolutions separately.
Flash\_SD3 maintains consistent FID and Precision-Recall performance across different resolutions, whereas many other models exhibit significant variations. Among them, Flash\_SD shoes low FID combined with high precision at 512$\times$512 but shifts to higher FID, lower precision and higher recall at 768$\times$1024 and 1024$\times$1024. These trends are visually confirmed by the generated images in \Cref{fig:prompt_length_512x512}, \Cref{fig:prompt_length_768x1024} and \Cref{fig:prompt_length_1024x1024}.  

\Cref{fig:prompt_length_1024x1024} presents the same subject (prompt number 0) across different diffusion models at 1024$\times$1024 resolution with int8 quantization, varying only the prompt length. 
As discussed in \Cref{app:prompt_length}, prompt length generally does not impact the energy consumption of the generation process. 
From a quality perspective, short-prompt images cannot be definitively classified as lower quality. However, it is interesting to note that the most significant differences in image content arise between short and long prompts. This aligns with the goal of prompt engineering for text-to-image generation, which is to produce more detailed and complex images.
Other examples are shown in \Cref{fig:prompt_length_512x512} and \Cref{fig:prompt_length_768x1024}.

\begin{figure}
    \centering
    \includegraphics[width=0.6\linewidth]{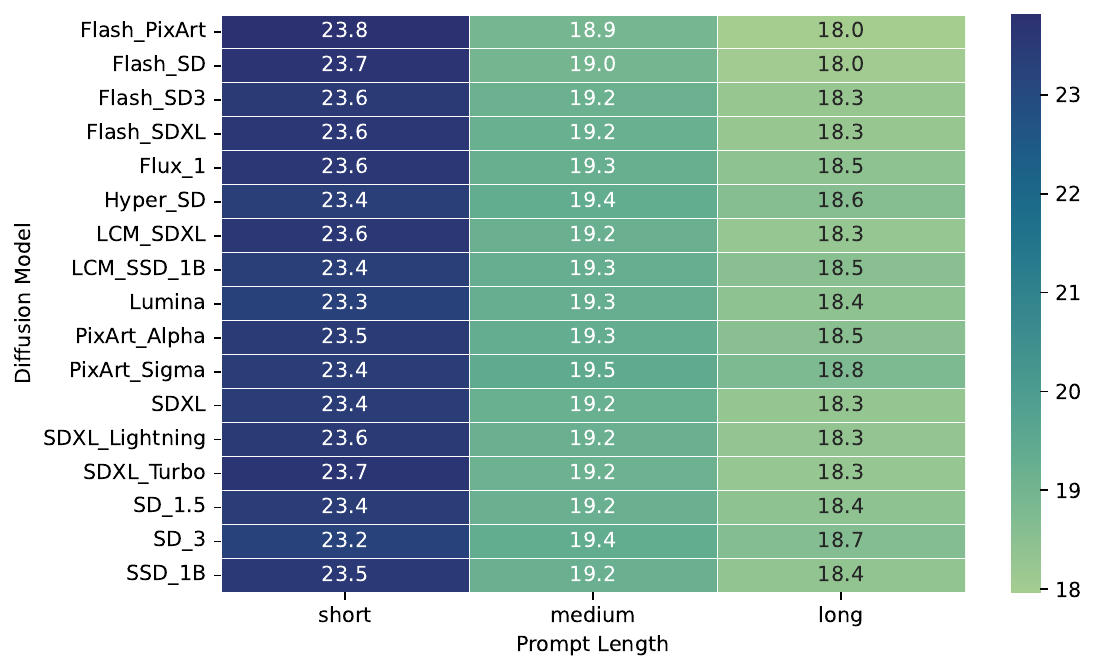}
    \caption{Average CLIPScore values for each diffusion model with respect to different prompt lengths.}
    \label{fig:clipscore}
\end{figure}

\Cref{fig:clipscore} presents the average CLIPScore values for all examined models across varying prompt lengths. The results indicate that shorter prompts consistently achieve the highest scores for every model, whereas longer prompts yield the lowest. This pattern suggests that when prompts contain only limited information, the generated images are less likely to deviate from the requested content, resulting in looser alignment with the given text. In contrast, longer prompts introduce more details, which the generation process may struggle to accurately capture, leading to greater misalignment, and consequently lower CLIPScore values.

\end{document}